\documentclass[]{jair}

\setcopyright{cc}
\copyrightyear{2025}
\acmDOI{10.1613/jair.1.xxxxx}

\JAIRAE{Insert JAIR AE Name}
\JAIRTrack{} 
\acmVolume{4}
\acmArticle{6}
\acmMonth{8}
\acmYear{2025}

\RequirePackage[
  datamodel=acmdatamodel,
  style=acmnumeric,
  backend=biber,
  giveninits=true
  ]{biblatex}

\usepackage{graphicx}
\usepackage{framed}
\usepackage{subcaption}
\usepackage{wrapfig}
\usepackage{url}
\usepackage{amsmath}
\usepackage{amsfonts}
\usepackage{amsthm}
\usepackage{cleveref}
\usepackage[utf8]{inputenc}
\usepackage{algorithm}
\usepackage[noend]{algpseudocode}

\usepackage{comment}
\usepackage{bbm}
\usepackage{bbding}
\usepackage{wrapfig,lipsum,booktabs}
\usepackage{float}
\usepackage{soul}
\usepackage{enumitem}

\newcommand{\etal}{\textit{et al}.}

\DeclareMathOperator*{\argmin}{arg\,min}

\newcommand{\rebuttal}[1]{#1}
\newcommand{\newtheory}[1]{#1}

\newtheorem{theorem}{Theorem}
\newtheorem{asm}{Assumption}
\newtheorem{remark}{Remark}
\newtheorem{lemma}{Lemma}
\newtheorem{proposition}{Proposition}

\begin{document}

\title[Implicit Safe Set Algorithm]{Implicit Safe Set Algorithm for Provably Safe Reinforcement Learning}

\author{Weiye Zhao}
\orcid{0000-0002-8426-5238}
\email{weiyezha@andrew.cmu.edu}
\affiliation{%
  \institution{Carnegie Mellon University}
  \city{Pittsburgh}
  \state{Pennsylvania}
  \country{United States}
}.

\author{Feihan Li}
\orcid{0000-0003-1770-4664}
\email{feihanl@andrew.cmu.edu}
\affiliation{%
  \institution{Carnegie Mellon University}
  \city{Pittsburgh}
  \state{Pennsylvania}
  \country{United States}
}

\author{Changliu Liu}
\authornote{Corresponding Author.}
\orcid{0000-0002-3767-5517}
\email{cliu6@andrew.cmu.edu}
\affiliation{%
  \institution{Carnegie Mellon University}
  \city{Pittsburgh}
  \state{Pennsylvania}
  \country{United States}
}

\renewcommand{\shortauthors}{Zhao, Li \& Liu}

\begin{abstract}
    Deep reinforcement learning (DRL) has demonstrated remarkable performance in many continuous control tasks. However, a significant obstacle to the real-world application of DRL is the lack of safety guarantees. Although DRL agents can satisfy system safety in expectation through reward shaping, designing agents to consistently meet hard constraints (e.g., safety specifications) at every time step remains a formidable challenge. In contrast, existing work in the field of safe control provides guarantees on persistent satisfaction of hard safety constraints. However, these methods require explicit analytical system dynamics models to synthesize safe control, which are typically inaccessible in DRL settings. In this paper, we present a model-free safe control algorithm, the implicit safe set algorithm, for synthesizing safeguards for DRL agents that ensure provable safety throughout training. The proposed algorithm synthesizes a safety index (barrier certificate) and a subsequent safe control law solely by querying a black-box dynamic function (e.g., a digital twin simulator). Moreover, we theoretically prove that the implicit safe set algorithm guarantees \textit{finite time convergence} to the safe set and \textit{forward invariance} for both continuous-time and discrete-time systems. We validate the proposed algorithm on the state-of-the-art Safety Gym benchmark, where it achieves zero safety violations while gaining $95\% \pm 9\%$ cumulative reward compared to state-of-the-art safe DRL methods. Furthermore, the resulting algorithm scales well to high-dimensional systems with parallel computing.
\end{abstract}



\received{10 June 2025}
\received[accepted]{10 June 2025}

\maketitle

\section{Introduction}
\label{intro}
Deep reinforcement learning (DRL) has achieved impressive results in continuous control tasks~\cite{fujimoto2025generalpurposemodelfreereinforcementlearning,10994242,zhao2024absolute,schulman2017proximalpolicyoptimizationalgorithms}, but its real-world deployment is hindered by the absence of hard safety guarantees. Ensuring safety in terms of persistently satisfying hard state constraints has long been recognized as a critical requirement in robotics~\cite{zhao2024absolutestatewiseconstrainedpolicy,zhao2023state}. For example, vehicles must avoid colliding with pedestrians; robotic arms should not strike walls; and aerial drones must steer clear of buildings.
When robots learn skills via reinforcement learning, they need to sufficiently explore the environment to discover optimal policies. However, such exploration is not inherently safe. A common strategy to encourage safe behavior is to introduce penalties for visiting unsafe states. While these \textit{posterior} measures are widely adopted~\cite{papini2019smoothing,pirotta2013safe,zhao2023state}, they are often insufficient to ensure robust safety. Instead, \textit{prior} measures are required to proactively prevent robots from entering unsafe states in the first place.


Constraining robot motion to persistently satisfy hard safety specifications in uncertain environments is an active area of research in the safe control community. Among the most widely used approaches are \textit{energy function-based methods}. These methods~\cite{khatib1986real,ames2014control,liu2014control,gracia2013reactive} define an energy function that assigns low energy to safe states and construct a control law that dissipates this energy. When the energy function and the control law are properly designed, the system remains within the safe set (i.e., exhibits \textit{forward invariance}) and may also converge to safe states in finite time if it starts from an unsafe condition (i.e., demonstrates \textit{finite-time convergence}).
Initially developed for deterministic systems, these methods have been extended to handle stochastic and uncertain systems~\cite{pandya2024multimodalsafecontrolhumanrobot,cosner2023robustsafetystochasticuncertainty,cheng2019end,taylor2020adaptive}, where uncertainties may arise from state measurements, the future evolution of the ego robot, or the behavior of obstacles. Furthermore, they have been integrated with robot learning frameworks to enhance safety assurance~\cite{zhang2025passivitycentricsafereinforcementlearning,du2023reinforcementlearningsaferobot,fisac2018general}. 
However, a key limitation of energy function-based methods is their reliance on \textit{white-box analytical models} of the system dynamics (e.g., the Kinematic Bicycle Model~\cite{kong2015kinematic}) for both offline construction of the energy function and online computation of the safe control signal.

This paper extends the class of \textit{energy function-based methods} by proposing a model-free safe control approach that provides safety guarantees without requiring an analytical dynamics model. The proposed method is capable of safeguarding any robot learning algorithm by integrating safety constraints directly into the control synthesis.
The key insight underlying our approach is that a safe control law can be synthesized without access to a white-box analytical dynamics model, provided that a black-box dynamics function is available—that is, a function that maps the current state and control input to the next state. Importantly, such black-box models are often accessible in real-world applications, for example, through high-fidelity digital twin simulators~\cite{abouchakra2025realissimbridgingsimtorealgap,das2022edgeassistedcollaborativedigitaltwin,liu2021twin} or deep neural network dynamics models trained in a data-driven manner~\cite{li2025continuallearningliftingkoopman,zhang2021fidelity}.
While this paper does not explicitly analyze formal guarantees on the accuracy of the black-box dynamics, it focuses on developing a reliable approach to synthesize the safe control strategy using only the black-box dynamics models, which is essential when designing safety-critical learning-based control systems in practice.

The key questions that we want to answer are: 1) 
how to synthesize the energy function (in our case called the safety index) with black-box dynamics, so that there always exists a feasible control to dissipate the system energy at all potentially unsafe states; 2) how to efficiently synthesize the optimal safe control with black-box dynamics, such that the control input can both dissipate the system energy and achieve good task efficiency; and 3) how to formally prove that the synthesized energy function and the generated safe control can guarantee \textit{forward invariance} and \textit{finite time convergence} to the safe set.
We propose a safety index design rule to address the first question, and a sample-efficient algorithm to perform black-box constrained optimization to address the second question. As for the third question, we show that under certain assumptions, the proposed safety index design rule together with the proposed black-box optimization algorithm can guarantee \textit{forward invariance} and \textit{finite time convergence} to a subset within the safe set.
By combining these approaches with the (model-based) safe set algorithm~\cite{liu2014control}, we propose the (model-free) implicit safe set algorithm (ISSA). We then use ISSA to safeguard deep reinforcement learning (DRL) agents. 

A short version of this paper has been presented earlier~\cite{zhao2021model}, where ISSA is proposed and evaluated with the assumption that the sampling time is almost zero (i.e., the system evolves almost like a continuous-time system). In reality, most control and simulation systems are implemented in a discrete-time manner with non-negligible sampling time. As a result, the controller may not be able to respond immediately to potential violations of safety. 
Therefore, assuring provable safety for discrete-time systems with non-negligible sampling time is important. In this paper, we introduce a general version of the implicit safe set algorithm which explicitly takes the sampling time into consideration. In particular, a novel convergence trigger (CTrigger) algorithm is introduced. The key contributions of this paper are summarized below:

\begin{itemize}
\item We propose two techniques to enable safe control with black-box models: (offline) continuous-time and discrete-time safety index design rules for mobile robots in a 2D plane, and a (online) sample-efficient black-box optimization algorithm using adaptive momentum boundary approximation (AdamBA). 
\item We propose the implicit safe set algorithm (ISSA) using these two techniques,  which guarantees to generate safe controls for all system states without knowing the explicit system dynamics. We show that ISSA can safeguard DRL agents to ensure zero safety violation during training in Safety Gym. Our code is available on Github.\footnote{\url{https://github.com/intelligent-control-lab/Implicit_Safe_Set_Algorithm}}
\item We provide the theoretical guarantees that 1) ISSA ensures the \textit{forward-invariance} safety for both continuous-time and discrete-time systems; and 2) \textit{finite time convergence} safety when ISSA is combined with CTrigger for both continuous-time and discrete-time systems (i.e., with any sampling time step).
\end{itemize}

In the remainder of the paper, we first formulate the mathematical problem in \Cref{subsec: probform}. We then discuss related work about safe reinforcement and safe control in \Cref{sec:related}. In \Cref{sec:method} and \Cref{sec:theory_contiuous}, we first introduce the implicit safe set algorithm, and provide theoretical results in continuous-time systems. We further extend the implicit safe set algorithm to the discrete-time systems and provide theoretical guarantees of \textit{forward invariance} and \textit{finite time convergence} in \Cref{sec:discrete method} and \Cref{sec:theory_discrete}, respectively. Finally, we validate our proposed method in Safety Gym environments in \Cref{sec:experiment}.

\section{Problem Formulation}
\label{subsec: probform}
\paragraph*{Dynamics}
 Let $x_t \in \mathcal{X}\subset \mathbb{R}^{n_x}$ be the robot state at time step $t$, where $n_x$ is the dimension of the state space $\mathcal{X}$; $u_t \in \mathcal{U}\subset \mathbb{R}^{n_u}$ be the control input to the robot at time step $t$, where $n_u$ is the dimension of the control space $\mathcal{U}$. Denote the sampling time as $dt$. The system dynamics are defined as: 
\begin{equation}\label{eq:dynamics fn}
\begin{split}
    &x_{t+1} = f(x_t,u_t), \\
\end{split}
\end{equation}
where $f: \mathcal{X} \times \mathcal{U} \rightarrow \mathcal{X}$ is a function that maps the current robot state and control to the next robot state. For simplicity, this paper considers deterministic dynamics. The proposed method can be easily extended to the stochastic case through robust safe control~\cite{emam2022safereinforcementlearningusing}, which will be left for future work. Moreover, it is assumed that we can only access an implicit black-box form of $f$, e.g., as an implicit digital twin simulator or a deep neural network model. \rebuttal{Note that the word \textit{implicit} refers to that we can only do point-wise evaluation of $f(x,u)$ without any explicit knowledge or analytical form of $f(x,u)$.}  

\begin{wrapfigure}{r}{0.24\textwidth}
    \centering
    \includegraphics[width=0.22\textwidth]{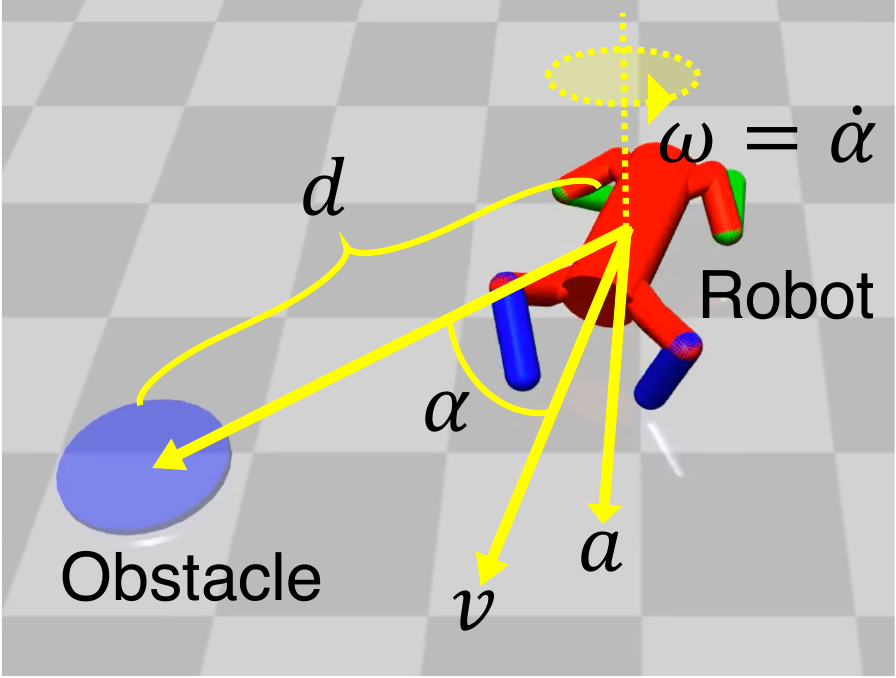}
    \caption{Visualization of robot notations.}
    \label{fig:doggo_rule}
\end{wrapfigure}

\paragraph*{Safety Specification}
The safety specification requires that the system state should be constrained in a regular and closed subset in the state space, called the safe set $\mathcal{X}_S$. The safe set can be represented by the zero-sublevel set of a continuous and piecewise smooth function $\phi_0:\mathbb{R}^{n_x} \rightarrow \mathbb{R}$, i.e., $\mathcal{X}_S = \{x\mid \phi_0(x)\leq 0\}$. $\mathcal{X}_S$ and $\phi_0$ are directly specified by users. A state $x(t_0)$ is considered \textit{safe} if it belongs to the safe set $\mathcal{X}_S$ at time $t_0$, notwithstanding the possibility that the trajectory $x(t)$ may exit $\mathcal{X}_S$ at some later time $t_1 > t_0$. The notion for the state, conditioned on the control policy, to result in a trajectory that always lies in the safe set is called \textit{invariantly safe}, which is later captured by the notion of \textit{forward invariance}. Hence the safe set $\mathcal{X}_S$ captures static safety and \textit{forward invariance}, which usually happens within a subset of $\mathcal{X}_S$, captures dynamic safety. \rebuttal{Nevertheless, the design of $\phi_0$ is straightforward in most scenarios. For example, for collision avoidance, $\phi_0$ can be designed as the negative closest distance between the robot and environmental obstacles.}

\paragraph*{Reward and Nominal Control}
A robot learning controller generates the nominal control which is subject to modification by the safeguard. The learning controller aims to maximize rewards in 
an infinite-horizon deterministic Markov decision process
(MDP). An MDP is specified by a tuple $(\mathcal{X}, \mathcal{U}, \gamma, r, f)$, where $r: \mathcal{X} \times \mathcal{U} \rightarrow \mathbb{R}$ is the reward function, $ 0 \leq \gamma < 1$ is the discount factor, and $f$ is the deterministic system dynamics defined in \eqref{eq:dynamics fn}. 

\paragraph*{Safeguard Synthesis}
Building on top of the nominal learning agent, the core problem of this paper is to synthesize a safeguard for the learning agent, which monitors and modifies the nominal control to ensure \textbf{forward invariance} in a subset of the safe set $\mathcal{X}_S$ and \textbf{finite time convergence} to that subset. 
\textit{Forward invariance} of a set means that the robot state will never leave the set if it starts from the set. \textit{Finite time convergence} to a set means that the robot state will return to the set in a finite amount of time if it is not initially in the set.
The reason why we need to find a subset instead of directly enforcing \textit{forward invariance} of $\mathcal{X}_S$ is that $\mathcal{X}_S$ 
may contain states that will inevitably go to the unsafe set no matter what the control input is. 
These states need to be penalized when we synthesize the energy function. For example, when a vehicle is moving toward an obstacle at high speed, it would be too late to stop. Even if the vehicle is safe now (if $\mathcal{X}_S$ only constrains the position), it will eventually collide with the obstacle (unsafe). Then we need to assign high energy values to these inevitably-unsafe states.

\paragraph*{Collision Avoidance for Mobile Robots}
In this paper, we  are focused on collision avoidance for mobile robots in a 2D plane. The robot and the obstacles are all geometrically simplified as point-mass circles with bounded collision radius. The safety specification is defined as $\phi_0 = \max_i {\phi_0}_i$, where ${\phi_0}_i = d_{min} - d_i$, and $d_i$ denotes the distance between the center point of the robot and the center point of the $i$-th obstacle (static or non-static), and the radius of both the obstacle and the robot are considered in $d_{min}$ such that $d_{min} \geq \text{obstacle radius} + \text{robot radius}$.
Denote $v$, $a$ and $w$ as the relative velocity, relative acceleration and relative angular velocity of the robot
with respect to the obstacle, respectively, as shown in \cref{fig:doggo_rule}.\footnote{Note that in our definition, $w = \dot \alpha$, where $\alpha$ is the angle between the robot's heading vector and the vector from the robot to the obstacle. The definition of $w$ is different from the robot angular velocity in the world frame, where 
$w$ is influenced by the 1) the robot's motion around the obstacle, which changes the vector from the robot to the obstacle and 2) the robot's self-rotation in world frame, which changes the robot's heading. Since the obstacle is treated as a circle, we do not consider its self-rotation in the world frame. In real world systems, a robot's self-rotation is bounded, and its motion around the obstacle is also bounded. Therefore, $w$ is bounded.} The remaining discussions are based on the following assumption. 

\begin{asm}[Actuation Reachability]
1) The state space is bounded ($ v \in [0, v_{max}]$), and the relative acceleration and angular velocity are bounded and both can achieve zeros, i.e., $w \in [w_{min}, w_{max}]$ for $w_{min} \leq 0 \leq w_{max}$ and $a \in [a_{min}, a_{max}]$ for $a_{min} \leq 0\leq a_{max}$; 
2) For all possible values of $a$ and $w$, there always exists an input $u$ such that the closed-loop low-level servo tracks that pair within the sampling period $dt$. 
\label{asm:ruleassumption}
\end{asm}

These assumptions are easy to meet in practice. The bounds in the first assumption will be directly used to synthesize the safety index $\phi$ (will be introduced in \Cref{sec:related}). The second assumption enables us to turn the question on whether there exists a feasible safe control in $\mathcal{U}$ to the question on whether there exists $a$ and $w$ to influence $\phi$. The requirement is satisfied by most wheeled or tracked mobile bases whose wheel/track torques can be commanded independently. It does not imply that the robot should be able to perform holonomic motion in Cartesian space; translational motion remains subject to the usual non-holonomic constraint. Our experiments use MuJoCo’s unicycle and quadruped models whose low-level velocity controllers meet this reachability property.

\section{Related Work}
\label{sec:related}
\paragraph*{Safe Reinforcement Learning}
Safe RL methods enforce safety through either soft safety constraints or hard safety constraints. For example, as for the safety specification of mobile robots, soft safety contraints limit a number of crashes per trajectory, whereas hard safety constraints require mobile robots should never crash into pedestrians.
Typical safe RL methods for soft safety constraints include risk-sensitive safe RL~\cite{11039715,garcia2015comprehensive}, Lagrangian methods~\cite{ray2019benchmarking}, trust-region-based constrained policy optimization ~\cite{achiam2017constrained,zhao2024statewiseconstrainedpolicyoptimization,zhao2024absolutestatewiseconstrainedpolicy}. These methods are able to find policies that satisfy 
the safety constraint
in expectation, but cannot ensure all visited states are safe. The methods that are more closely related to ours are safe RL methods with hard safety constraints. 
Safe RL methods with hard constraints can be divided into two categories: 1) safeguarding learning policies using control theories; and 2) safeguarding learning policies through a safety layer to encourage safe actions. 

For the first category, safeguard methods are proposed based on 1) Control Barrier Function (CBF), 2) Hamilton-Jacobi reachability, and 3) Lyapunov method. 
Richard \etal~\cite{cheng2019end} propose a general safe RL framework, which combines a CBF-based safeguard with RL algorithms to guide the learning process by constraining the exploratory actions to the ones that will lead to safe future states. However, this method strongly relies on the control-affine structure of the dynamics system, which restricts its applicability to general non-control-affine unknown dynamics systems. 
Ferlez \etal~\cite{ferlez2020shieldnn} also propose a ShieldNN leveraging Control Barrier Function to design a safety filter neural network with safety guarantees. However, ShieldNN is specially designed for an environment with the kinematic bicycle model (KBM) \cite{kong2015kinematic} as the dynamical model, which cannot be directly applied to general problems.
Besides CBF, reachability analysis is also adopted for the safeguard synthesis. Fisac \etal~\cite{fisac2018general} propose a general safety framework based on Hamilton-Jacobi reachability methods that works in conjunction with an arbitrary learning algorithm. However, these methods~\cite{fisac2018general,pham2018optlayer} still rely heavily on the explicit analytic form of system dynamics to guarantee constraint satisfaction.
In addition to CBF and reachability, Lyapunov methods are used to verify stability of a known system in control, which can be used to determine whether states and actions are safe or unsafe.
Berkenkamp \etal~\cite{berkenkamp2017safe} combine RL with Lyapunov analysis to ensure safe operation in discretized systems. Though provable safe control can be guaranteed under some Lipschitz continuity conditions, this method still requires explicit knowledge of the system dynamics (analytical form).
In summary, control theory based safeguard methods  provide good safety guarantees. However, those methods rely on the assumptions of 1) explicit knowledge of dynamics model or 2) control-affine dynamics model, which are hard to be satisfied in real world robotics applications. 

The second category to make RL meet hard safety constraints is shield synthesis–style approaches, where a safety layer monitors the action proposed by the RL policy and corrects it, if necessary, to satisfy safety constraints. Such methods can guarantee safety online by projecting unsafe actions into a certified safe set before execution.
Dalal \etal~\cite{dalal2018safe} propose to learn the system dynamics directly from offline collected data, and then add a safety layer that analytically solves a control correction formulation at each state to ensure every state is safe. However, the closed-form solution relies on the assumption of the linearized dynamics model and cost function, which is not true for most dynamics systems. They also assume the set of safe control can be represented by a linearized half-space for all states, which does not hold for most of the discrete-time system (i.e., safe control may not exist for some states). Yinlam \etal~\cite{chow2019lyapunov} propose to project either the policy parameters or the action to the set induced by linearized Lyapunov constraints, which still suffer from the same linear approximation error and non-control-affine systems as in~\cite{dalal2018safe} and is not able to guarantee zero-violation.
Bejarano \etal~\cite{Bejarano_2025} integrate safety filters into the RL training process, allowing the controller to adapt to the presence of the filter and thereby improving the overall safety–performance trade-off and sample efficiency. However, their approach still fundamentally relies on an analytical dynamics model, which limits applicability in scenarios where such a model is unavailable or inaccurate.
These works all fall under the general paradigm of shield synthesis in RL, where a pre-specified or learned model is used online to intercept and correct unsafe actions before execution.
In contrast, our proposed method does not require the explicit form or control-affine form of system dynamics, and our method can be guaranteed to generate safe control 
as long as the safe set is regular (i.e., the set equal to the closure of its interior; for example, it is not a collection of singleton points) and contains forward invariant subsets.

\paragraph*{Safe Control}

Representative \textit{energy function-based methods} for safe control include potential field methods~\cite{khatib1986real}, control barrier functions (CBF)~\cite{ames2014control}, safe set algorithms (SSA)~\cite{liu2014control}, sliding mode algorithms~\cite{gracia2013reactive}, and a wide variety of bio-inspired algorithms~\cite{zhang2017bio}. The next step towards safe controller synthesis is to design a desired energy function offline, ensuring that 1) the low energy states are safe and 2) there always exists a feasible control input to dissipate the energy. SSA has introduced a rule-based approach~\cite{li2023learning} to synthesize the energy function as a continuous, piece-wise smooth scalar function on the system state space $\phi:\mathbb{R}^{n_x} \rightarrow \mathbb{R}$. And the energy function $\phi(x)$ is called a safety index in this approach. \rebuttal{The general form of the safety index was proposed as $\phi = \phi_0^* + k_1\dot\phi_0 + \cdots + k_n\phi_0^{(n)}$ where 1) the roots of $1+k_1s+\ldots + k_n s^n=0$ are all negative real (to ensure zero-overshooting of the original safety constraints); 2) the relative degree from $\phi_0^{(n)}$ to $u$ is one (to avoid singularity); and 3) $\phi_0^*$ defines the same set as $\phi_0$ (to nonlinear shape the gradient of $\phi$ at the boundary of the safe set). 
It is shown in \cite{liu2014control} that if the control input is unbounded ($\mathcal{U} =\mathbb{R}^{n_u}$), then there always exist a control input that satisfies the constraint $\dot\phi \leq 0 \text{ when } \phi= 0$; and if the control input always satisfies that constraint, then the set $\{x\mid\phi(x)\leq 0\}\cap \{x\mid \phi_0(x)\leq 0\}$ is forward invariant.}
In practice, the actual control signal is computed through a quadratic projection of the nominal control $u^r$ to the control constraint:
\begin{align}\label{eq: quadratic program for control}
    u =& \argmin_{u\in\mathcal{U}} \|u-u^r\|^2\text{ s.t. } \dot\phi \leq -\eta(\phi),
\end{align}
where $\dot\phi \leq -\eta(\phi)$ is a general form of the constraint; $\eta:\mathbb{R}\rightarrow\mathbb{R}$ is a non-decreasing function that $\eta(0)\geq 0$. For example, in CBF, $\eta(\phi)$ is designed to be $\lambda\phi$ for some positive scalar $\lambda$. In SSA, $\eta(\phi)$ is designed to be a positive constant when $\phi\geq0$ and $-\infty$ when $\phi<0$. 
Note there are two major differences between this paper and the existing results presented in \cite{liu2014control}. First, this paper considers constrained control space, which then requires careful selection of the parameters in $\phi$ to make sure the control constraint $\mathcal{U}_S(x):=\{u\in\mathcal{U}\mid \dot\phi \leq -\eta(\phi)\}$ is nonempty for states that $\phi\geq 0$. Secondly and most importantly, this paper considers general black-box dynamics, while the existing work considers analytical control-affine dynamics. For analytical control-affine dynamic models, $\mathcal{U}_S(x)$ is essentially a half-space, and the projection \eqref{eq: quadratic program for control} can be efficiently computed by calling a quadratic programming solver. However, for black-box dynamics, this constraint is challenging to quantify. 

Since real-world robot systems are always controlled in discrete-time with dynamics $x_{t+1} = f(x_t, u_t)$, we consider the discrete-time version of the set of safe control $\mathcal{U}_S^D(x):=\{u\in \mathcal{U}\mid \phi(f(x, u)) \leq \max\{\phi(x)-\eta, 0\} \}$ in the remaining discussions. By defining $\mathcal{X}_S^D:= \{x|\phi(x) \leq 0\}$ , our ultimate goal is to ensure forward invariance and finite-time convergence with respect to the set$\mathcal{S}:=\mathcal{X}_S\cap\mathcal{X}_S^D$.
Notice that the continuous-time system is a special case of a discrete-time system where the simulation time step (sampling time) is negligible, i.e. $dt \rightarrow 0$. Additionally, $\mathcal{U}_S(x)$ and $\mathcal{U}_S^D(x)$ pose similar requirements for safe control, which require $\phi$ to decrease at the next time step when $\phi \geq 0$ at the current time step.

\section{Implicit Safe Set Algorithm for Systems with Negligible Sampling Time}
\label{sec:method}

This section introduces the implicit safe set algorithm (ISSA), which is able to leverage \textit{energy function-based methods} (SSA in particular) with black-box dynamics, and be used to safeguard any nominal policy. ISSA contains two parts: 1) a safety index synthesis rule to make sure $\mathcal{U}_S^D(x)$ is nonempty for all $x$, and 2) a sample-efficient black-box optimization method to solve the projection of the nominal control to $\mathcal{U}_S^D(x)$.
With these two components, the overall pipeline of the implicit safe set algorithm is summarized as follows: 
\begin{itemize}
    \item \textbf{Offline:} Design the safety index $\phi(x)$ according to the safety index design rule.
    \item \textbf{Online:} Project nominal control into $\mathcal{U}_S^D(x)$ during online robot maneuvers.
\end{itemize}




\subsection{Synthesize Safety Index for Continuous-time System}
\label{sec:synthesis}
The safety index for collision avoidance in 2D plane will be synthesized without referring to the specific dynamic model, but under the following assumptions.

\begin{asm}[2D Collision Avoidance for continuous-time system]
1) The system time step $dt \to 0$.
2) At any given time, there can at most be one obstacle becoming safety-critical, such that $\phi_i \geq 0$ (Sparse Obstacle Environment).
\label{asm:continuous assumption}
\end{asm}

These assumptions are easy to meet in practice. The first assumption ensures that the discrete-time approximation error is negligible, i.e., the system can essentially be treated as a continuous-time system. The second assumption makes the safety index design rule applicable with multiple moving obstacles.

\begin{figure}[htbp]
    \centering
    \begin{subfigure}[t]{0.32\textwidth}
        \centering
        \includegraphics[width=\linewidth]{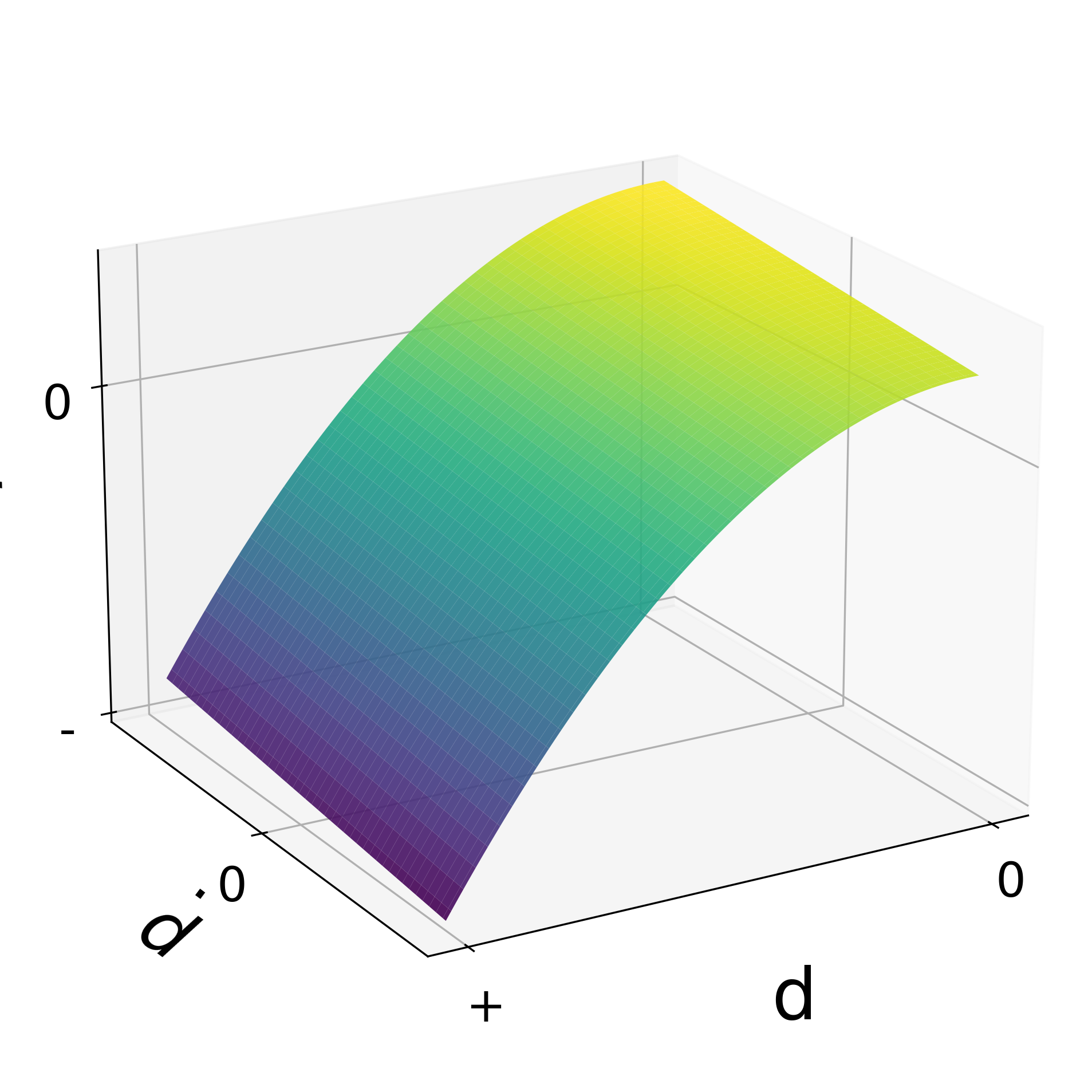}
        \caption{$k = 1, \ n = 2$}
    \end{subfigure}
    \begin{subfigure}[t]{0.32\textwidth}
        \centering
        \includegraphics[width=\linewidth]{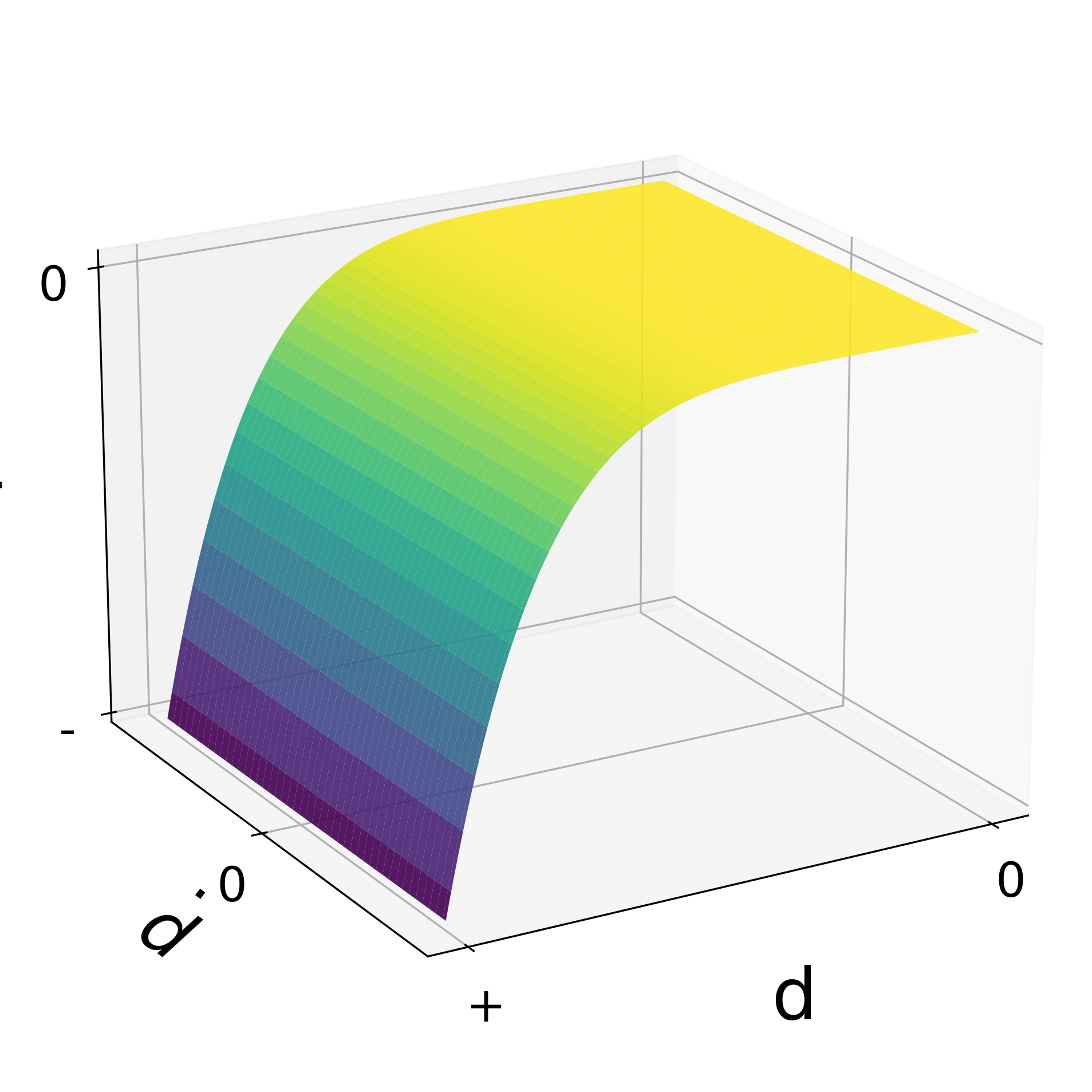}
        \caption{$k = 1, \ n = 6$}
    \end{subfigure}
    \begin{subfigure}[t]{0.32\textwidth}
        \centering
        \includegraphics[width=\linewidth]{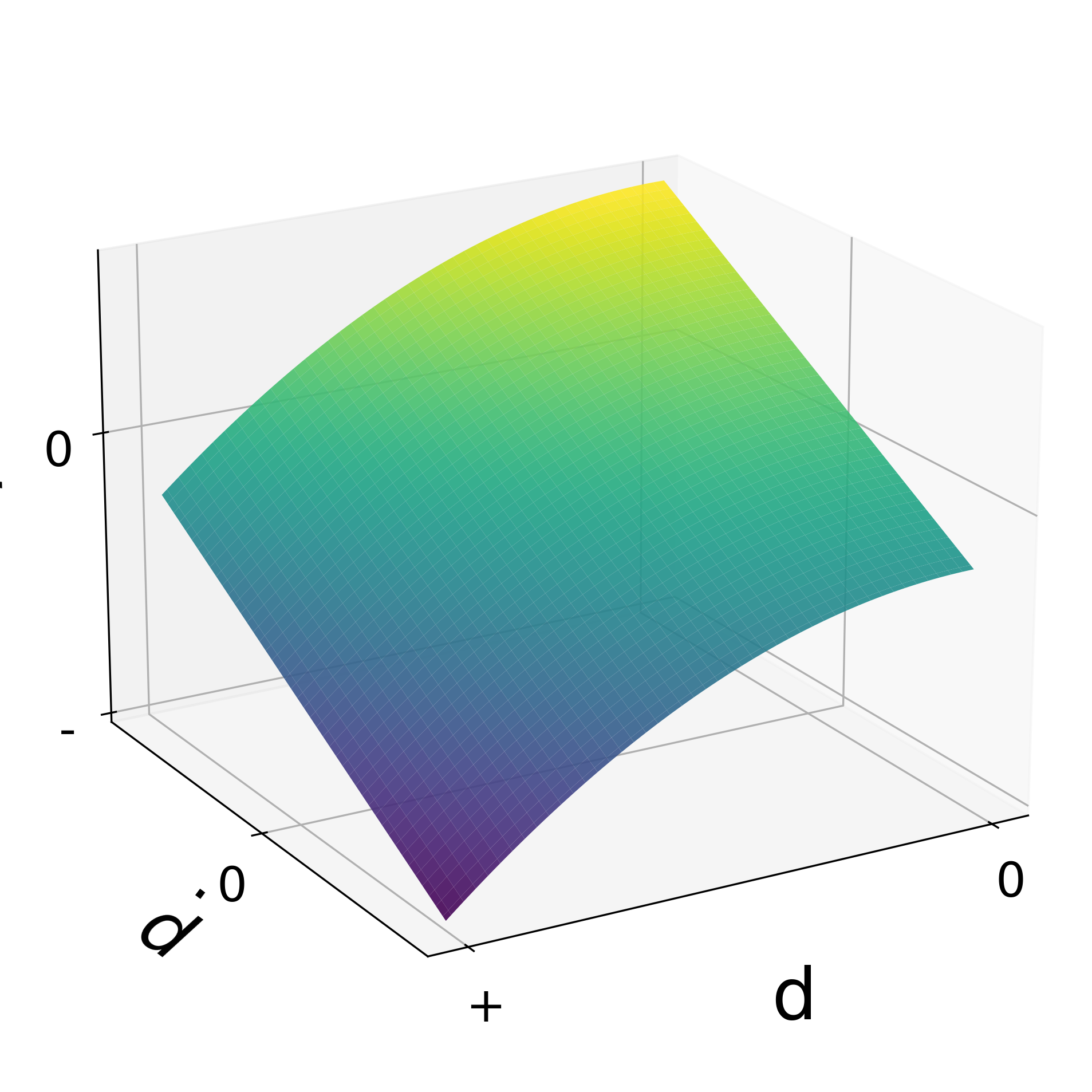}
        \caption{$k = 10, \ n = 2$}
    \end{subfigure}
    
    \caption{3D heat map illustration of the behavior of the safety index $\phi$ with respect to $d$ and $\dot{d}$ under varying $k$ and $n$, where z axis denotes the value of $\phi$. The brighter color denotes higher $\phi$ value. In the figure axes, “+” denotes positive values and “--” denotes negative values.}

    \label{fig:k_n_ablation}
\end{figure}

Following the rules in \cite{liu2014control}, we parameterize the safety index as $\phi = \max_i\phi_i$, 
\begin{align}\label{eq: safety index}
\phi_i = \sigma + d_{min}^n - d_i^n - k\dot d_i,
\end{align}
where all $\phi_i$ share the same set of tunable parameters $\sigma,n,k,\eta \in \mathbb{R}^+$. 
We illustrate the behavior of the safety index $\phi$ with respect to $d$ and $\dot{d}$ under varying $k$ and $n$ in \Cref{fig:k_n_ablation}.
Our goal is to choose these parameters such that $\mathcal{U}_S^D(x)$ is always nonempty. 
Under the above assumptions, the safety index design rule is constructed below. The main idea of designing the safety index is to derive the sufficient condition of parameters of $\phi$ to ensure there exist safe controls for all states.

\textbf{Safety Index Design Rule for Continuous-Time System:} By setting $\eta = 0$, the parameters $k, n$, and  $\sigma$ should be chosen such that
\begin{align}
\label{eq:continuous_rule}
    \frac{n (\sigma + d_{min}^n + k v_{max})^{\frac{n-1}{n}}}{k} \leq \frac{-a_{min}}{v_{max}},
\end{align}
where $v_{max}$ is the maximum relative velocity that the robot can achieve in the obstacle frame. This design rule, together with the conditions in \Cref{asm:ruleassumption} and \Cref{asm:continuous assumption}, guarantees that the safe control set $\mathcal{U}_S^D(x)$ is non-empty at any arbitrary state, which is the key to establishing \Cref{thoem:main}. The proof of this non-emptiness is provided in \Cref{prop:1}. 

\subsection{Understanding of Safety Index Design Rule}

\noindent\textit{Design philosophy.} We treat $n$ as a user-chosen shaping parameter for the safety index, while $k$ is then selected to satisfy \eqref{eq:continuous_rule} so that a safe control set $\mathcal{U}_S^D(x)$ is non-empty for all states. We begin with the role of $k$.

\paragraph{Effect of $k$.}
With safety index parameterization form in \eqref{eq: safety index}, the coefficient $k$ scales the index's sensitivity to the relative speed $\dot d$. Increasing $k$ steepens the slope of $\phi$ along the $\dot d$ axis, so $\phi$ rises more aggressively when the robot is closing in (i.e., $\dot d<0$). Visually, compare Fig.~\ref{fig:k_n_ablation}(a) ($k{=}1,n{=}2$) and Fig.~\ref{fig:k_n_ablation}(c) ($k{=}10,n{=}2$): the surface tilts more strongly with $\dot d$, indicating earlier and stronger reactions to robot-obstacle approach speed ($\dot d$). Equation~\eqref{eq:continuous_rule} explains why $k$ cannot be arbitrarily small: to keep the safe control set nonempty, $k$ must grow when the maximum relative velocity $v_{\max}$ is larger or when the available maximum deceleration $|a_{\min}|$ is smaller. 
Intuitively, faster environments or 
weaker braking demand earlier reactions, i.e., the larger the $k$ is, the more sensitive the safety mechanism is to $\dot d$, working like a time-to-collision style constraint.

\paragraph{Effect of $n$ (distance--velocity trade-off).}
The distance term $d_{\min}^n - d^n$ in \eqref{eq: safety index} sharpens as $n$ increases, redistributing attention toward \emph{short distances}. Comparing Fig.~\ref{fig:k_n_ablation}(a) ($k{=}1,n{=}2$) and Fig.~\ref{fig:k_n_ablation}(b) ($k{=}1,n{=}6$), the surface rises much more quickly as $d$ shrinks. Thus, for the same $(d,\dot d)$, a larger $n$ yields a larger $\phi$ when the robot is close, i.e., less tolerance to running toward an obstacle at short range, yet allows more tolerance to aggressive motion when the robot is far away. In other words, increasing $n$ changes the \emph{preference} of the safety index: strict when close, permissive when far.

\paragraph{Why $n$ and $k$ should change together.}
Once $n$ is chosen to set that preference, $k$ must be tuned to preserve feasibility under \eqref{eq:continuous_rule}. A larger $n$ makes $\phi$ grow rapidly at small $d$; with bounded deceleration $|a_{\min}|$, there is only so much (and so fast) the controller can slow down. To ensure nonempty set of safe control that prevents $\phi$ from diverging upward, the design must respond more sensitively when $\dot d$ decreases, i.e., increase $k$. This coupling is explicit in \eqref{eq:continuous_rule}, where $n$ and $k$ appear together (through a factor of the form $\tfrac{n}{k}\bigl(\sigma + d_{\min}^n + k\,v_{\max}\bigr)^{(n-1)/n}$): raising $n$ typically necessitates raising $k$ to keep the inequality satisfied. Practically: choose $n$ to sculpt proximity sensitivity; then increase $k$ as needed to guarantee a nonempty safe control set by considering speed/braking limits of the environment.

\subsection{Sample-Efficient Black-Box Constrained Optimization}
The nominal control $u_t^r$ needs to be projected to $\mathcal{U}_S^D(x)$ by solving the following optimization:
\begin{equation}\label{eq:adamba_discrete}
\begin{split}
    &\min_{u_t\in\mathcal{U}} \| u_t - u^r_t\|^2\\
    & \text{s.t. } \phi(f(x_t, u_t)) \leq \max\{\phi(x_t)-\eta, 0\}
\end{split}
\end{equation}

A key insight we have is that: since the objective of \eqref{eq:adamba_discrete} is convex, its optimal solution will always lie on the boundary of $\mathcal{U}_S^D(x)$ if $u^r \notin \mathcal{U}_S^D(x)$. Therefore, it is desired to have an efficient algorithm to find the safe controls on the boundary of $\mathcal{U}_S^D(x)$. To efficiently perform this black-box optimization, we propose a sample-efficient boundary approximation algorithm called Adaptive Momentum Boundary Approximation Algorithm (AdamBA), which is summarized in \Cref{alg:adamba}. AdamBA leverages the idea of adaptive line search with exponential outreach/decay to efficiently search for the boundary points.
The inputs for AdamBA are the approximation error bound ($\epsilon$), learning rate ($\beta$), reference control ($u^r$), gradient vector covariance ($\Sigma$), gradient vector number ($n$), reference gradient vector ($\vec{v}^r$), safety status of reference control ($S$), and the desired safety status of control solution ($S_{goal}$).

\begin{figure*}
\captionsetup[subfigure]{}
     \centering
    \begin{subfigure}[t]{0.23\textwidth}
        \raisebox{-\height}{\includegraphics[width=\textwidth]{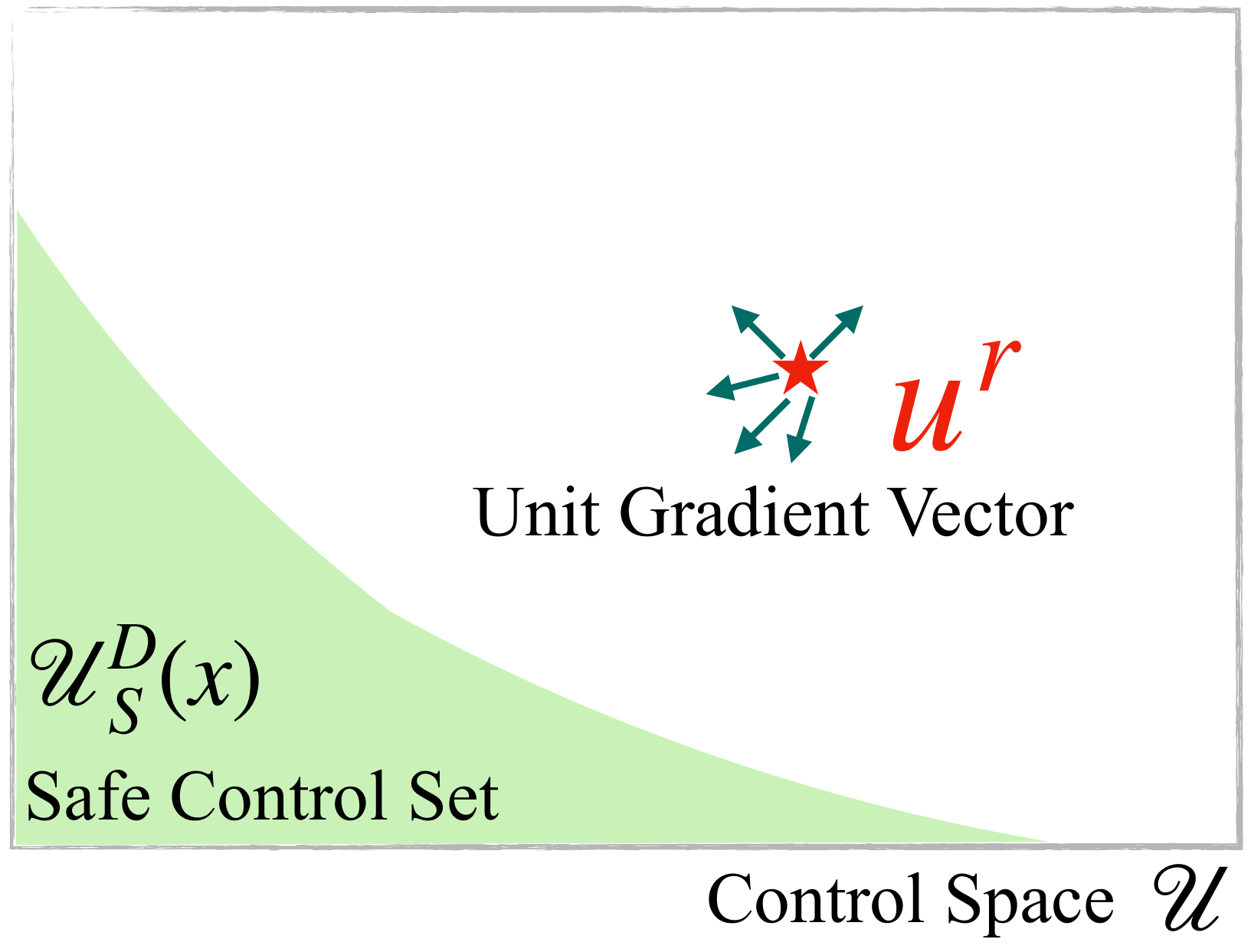}}
        \caption{Initial gradient vectors}
        \label{fig:adamba1}
    \end{subfigure}
    \hfill
    \begin{subfigure}[t]{0.23\textwidth}
        \raisebox{-\height}{\includegraphics[width=\textwidth]{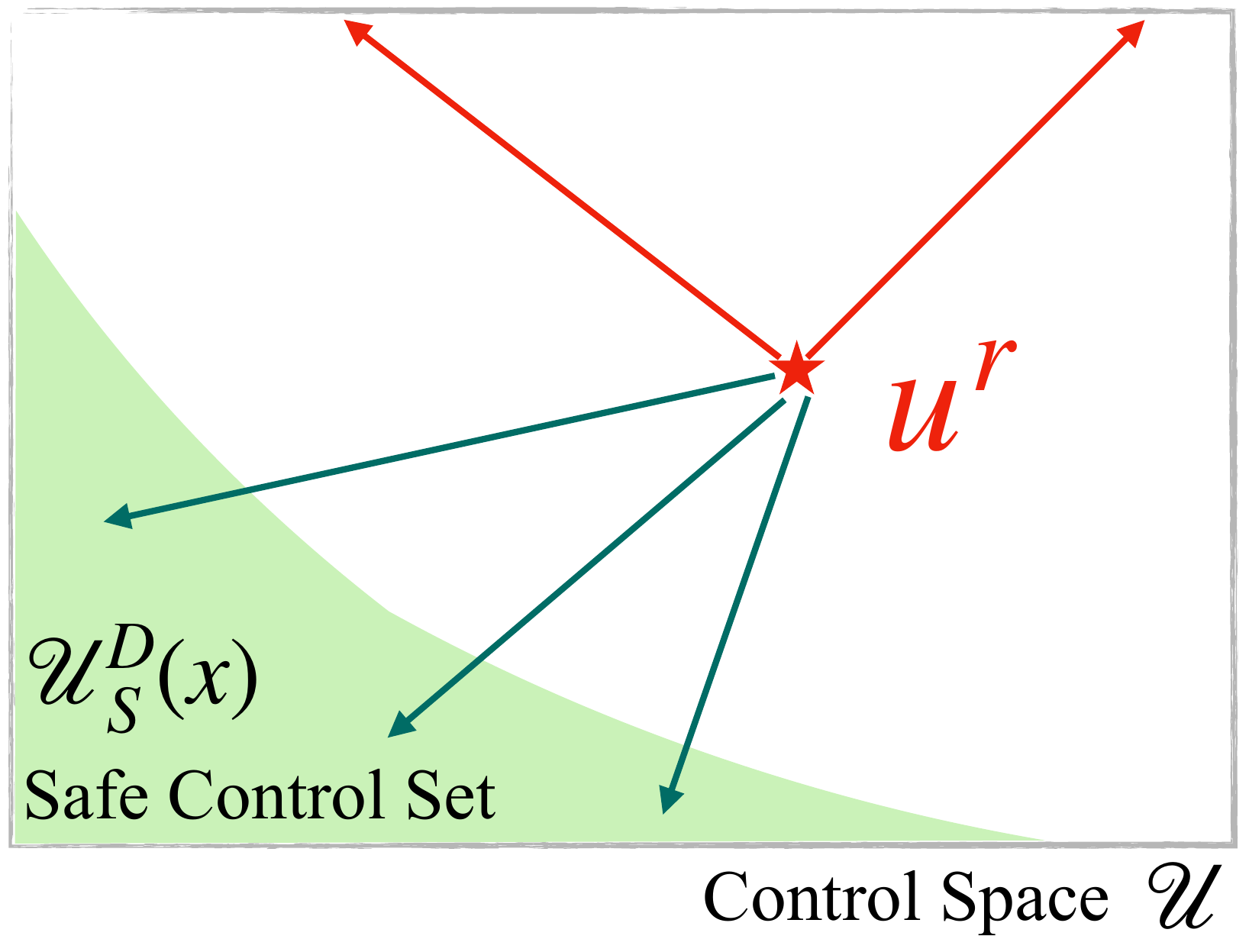}}
        \caption{Exponential outreach.}
        \label{fig:adamba2}
    \end{subfigure}
    \hfill
    \begin{subfigure}[t]{0.23\textwidth}
        \raisebox{-\height}{\includegraphics[width=\textwidth]{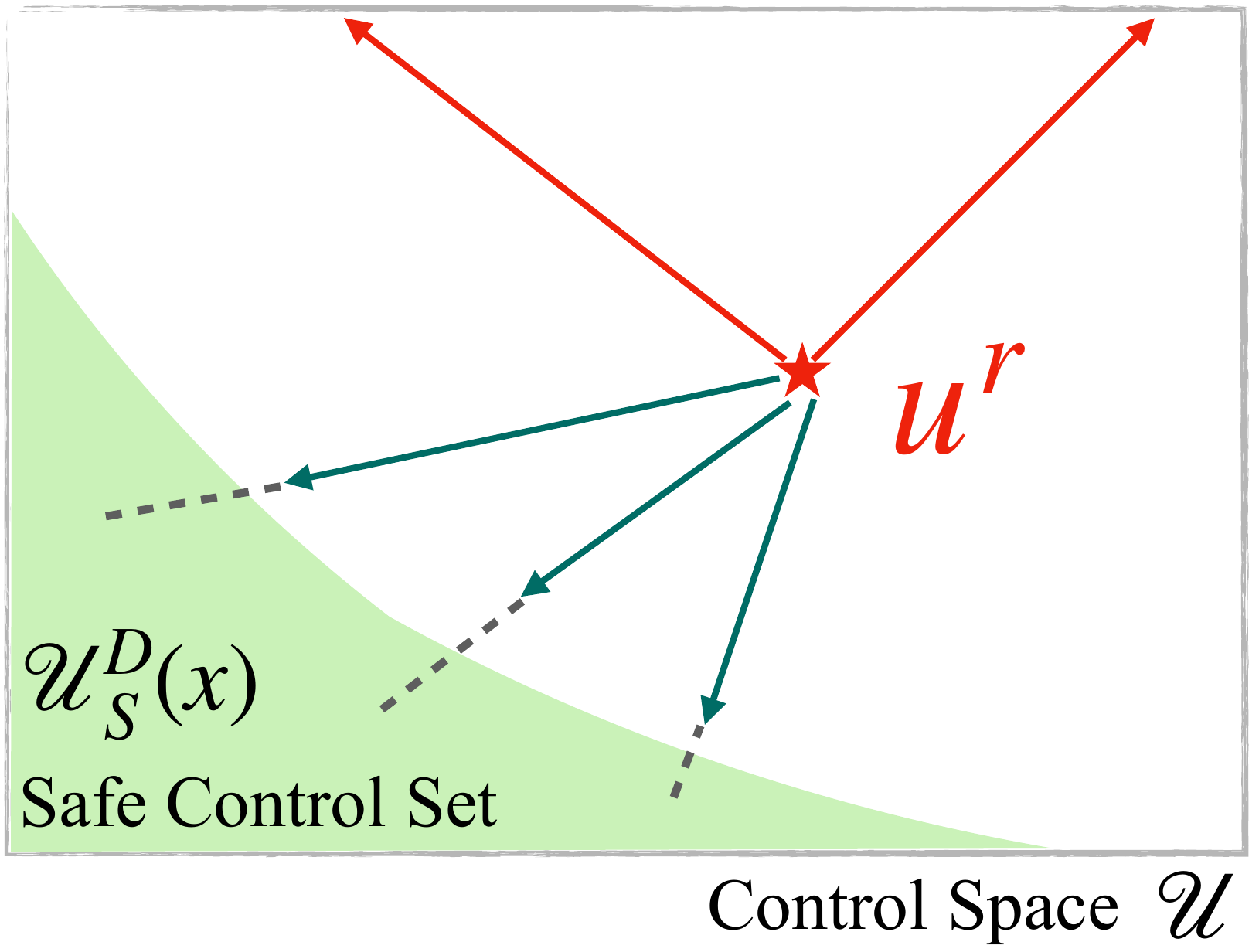}}
    \caption{Exponential decay.} 
    \label{fig:adamba3}
    \end{subfigure}
    \hfill
    \begin{subfigure}[t]{0.23\textwidth}
        \raisebox{-\height}{\includegraphics[width=\textwidth]{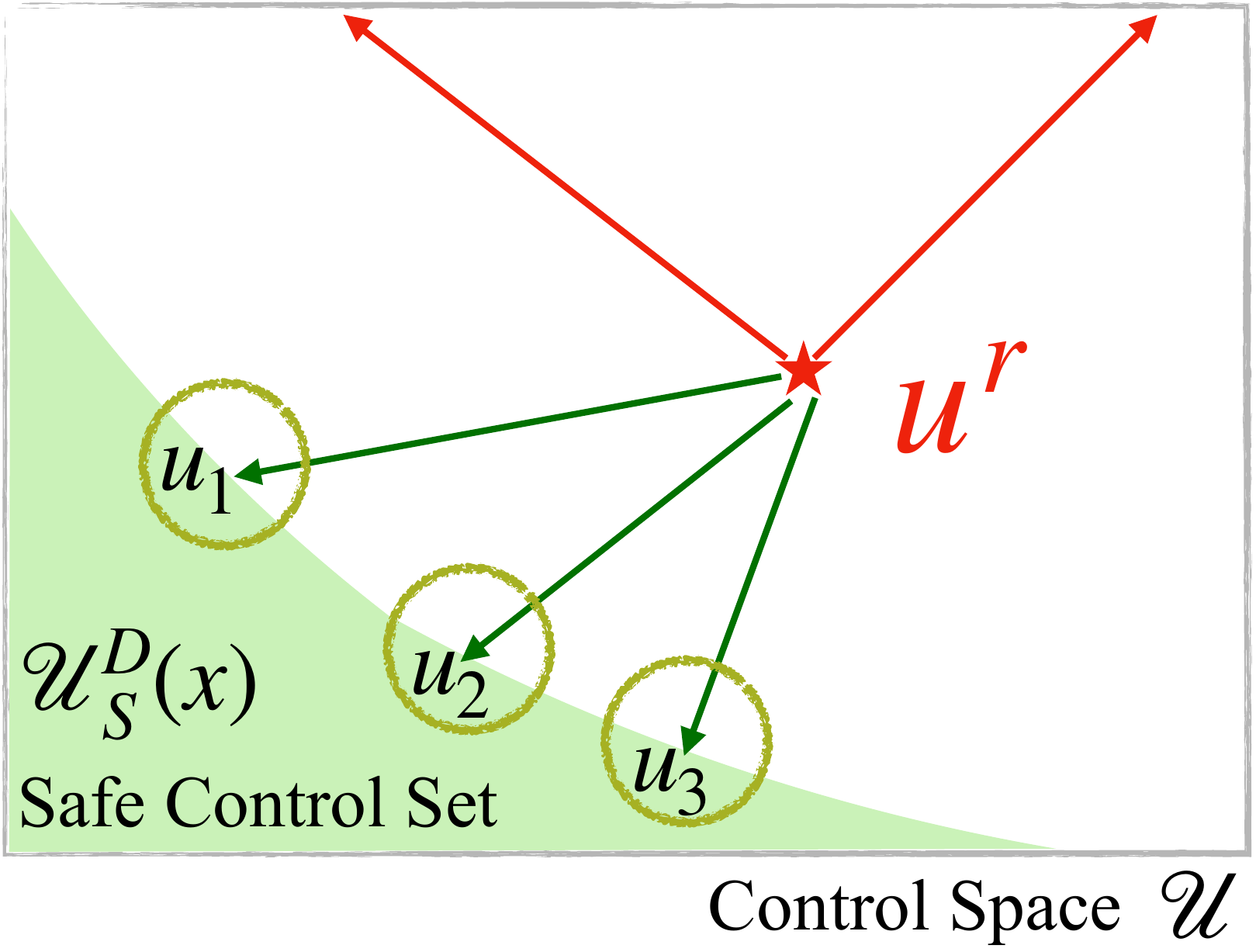}}
    \caption{Boundary points found.} 
    \label{fig:adamba4}
    \end{subfigure}
    \hfill
     \caption{Illustration of the procedure of the
     AdamBA algorithm. The algorithm is divided into four stages.}
     \label{fig:adamba_process}
\end{figure*}

We illustrate the main boundary approximation procedures of AdamBA in \Cref{fig:adamba_process}, where AdamBA is supposed to find the boundary points of $\mathcal{U}_S^D(x)$ (green area) with respect to the reference control $u^r \notin \mathcal{U}_S^D(x)$ (red star). The core idea of the AdamBA algorithm follows the adaptive line search~\cite{armijo1966minimization}, where three main procedures are included. I. AdamBA first initializes several unit gradient vectors (green vectors) to be the sampling directions, as shown in \Cref{fig:adamba1}. II. AdamBA enters the \textit{exponential outreach} phase by exponentially increasing the gradient vector length until they reach $\mathcal{U}_S^D(x)$ as shown in \Cref{fig:adamba2}. Note that we discard those gradient vectors that go out of control space (red vectors). III. Next, AdamBA enters the \textit{exponential decay} phase by iteratively applying Bisection search to find boundary points as shown in  \Cref{fig:adamba3}. Finally, a set of boundary points will be returned after AdamBA converges as shown in \Cref{fig:adamba4}. 
Note that AdamBA and the line search methods are fundamentally similar to each other, while the purpose of AdamBA is to find the boundary of safe/unsafe action, while the line search methods are to find the minimum of a function.
It can be shown that the boundary approximation error can be upper bounded within an arbitrary resolution. 

\begin{algorithm}
\caption{Adaptive Momentum Boundary Approximation}
\label{alg:adamba}
\begin{algorithmic}[1]
\Procedure{AdamBA}{$\epsilon, \beta, \Sigma, n, u^r, \vec{v}^r, S, S_{goal}$} 
\State \textbf{Initialize:}
\If{$\vec{v}^r$ is empty}
\State Generate $n$ Gaussian distributed unit gradient vectors $\vec{v}_i \sim \mathcal{N}(0,\Sigma)$, $i = 1,2,\ldots,n$
\Else 
    \State Initialize one unit gradient vector $\vec{v}_1 = \frac{\vec{v}^r}{\| \vec{v}^r\|}$
\EndIf

\State \textbf{Approximation:}
\For{$i = 1,2,\cdots,n$}
    \State Initialize the approximated boundary point $P_i = u^r$, and \textit{stage} = \textit{exponential outreach}. 
        \While{\textit{stage} = \textit{exponential outreach}}
            \State Set $P_S \leftarrow P_i$ and $P_{NS} \leftarrow P_i$
            \State $P_i = P_i + \vec{v}_i\beta$
            
            \If{$P_i$ is out of the control set}
                \State \textbf{break}
            \EndIf 
            
            \If{$P_i$ safety status $\neq S$ }
                \State Set $P_{NS} \leftarrow P_i$ \label{line:PNS set},  \textit{stage} $\leftarrow$ \textit{exponential decay}  \label{line:16}
                \State \textbf{break}
            \EndIf 
            \State $\beta = 2\beta$
        \EndWhile

    \If{\textit{stage} = \textit{exponential decay}}
        \State Apply Bisection method to locate boundary point until $\|P_{NS} - P_S\| < \epsilon$
        \State Set $P_i \leftarrow P_S$ if $S_{goal} = S$, $P_i \leftarrow P_{NS}$ otherwise \label{line output}
    \EndIf
\EndFor
\State Return Approximated Boundary Set $P$
\EndProcedure
\end{algorithmic}
\end{algorithm}

\paragraph{Relation to momentum methods.}
The proposed AdamBA algorithm employs a momentum-inspired strategy in step-size control. During the exponential outreach phase, the step length $\beta$ is geometrically amplified ($\beta \leftarrow 2\beta$) as long as the sampled point remains within the admissible region and retains its nominal safety label. This behavior emulates the inertial buildup of momentum and allows for rapid traversal of benign interior space. Upon detecting a change in the safety status, the algorithm transitions to a bisection-based exponential decay stage, refining the bracketing pair until a prescribed tolerance $\varepsilon$ is satisfied. This mirrors the damping phase observed in line-search corrections.

The interplay between these phases underscores the importance of hyperparameter selection. An initial $\beta$ that is too small results in wasted outreach doublings, while an overly aggressive choice may cause immediate overshoot and necessitate excessive corrective iterations. Similarly, the tolerance $\varepsilon$ governs the trade-off between computational cost and boundary fidelity. Empirically, setting $\beta$ to 5--10\% of the state-space scale, using a growth factor of 2, and selecting $\varepsilon$ commensurate with actuator resolution have been found to minimize the total number of  evaluations. Additionally, periodic resampling of probe directions helps mitigate directional bias in non-stationary environments. Collectively, these design principles enable AdamBA to harness momentum-like dynamics for fast yet precise boundary localization, making it a practical tool for safety-critical control applications.

\subsection{Implicit Safe Set Algorithm for Continuous-Time System}
\label{sec: issa}

\begin{algorithm}[h]
\caption{Implicit Safe Set Algorithm (ISSA)}\label{alg:adamsc}
\begin{algorithmic}[1]
\Procedure{ISSA}{$\epsilon, \beta, \Sigma, n, u^r$} 
\State \textbf{Phase 1:} \Comment{Phase 1}
\State Use AdamBA($\epsilon, \beta, \Sigma, n, u^r, \emptyset, \textit{UNSAFE}, \textit{SAFE}$) to sample a collection $\mathbb{S}$ of safe control on the boundary of $\mathcal{U}_S^D$.
\If{$\mathbb{S} = \emptyset$}
    \State Enter \textbf{Phase 2}
\Else
    \State For each primitive action $u_i \in \mathbb{S}$, compute the deviation $d_i = \|u_i - u^r \|^2$
    \State return $\argmin_{u_i} d_i$
\EndIf
\State 
\State \textbf{Phase 2:} \Comment{Phase 2}
\State Use grid sampling by iteratively increasing sampling resolution to find an anchor safe control $u^a$, s.t. safety status of $u^a$ is \textit{SAFE}. \label{line grid sampling}
\State Use AdamBA($\epsilon, \frac{\|u^r - u^a\|}{4}, \Sigma, 1, u^r, \frac{u^a - u^r}{\|u^a - u^r\|}, \textit{UNSAFE}, \textit{SAFE}$) to search for boundary point $u^*$ \label{line:admbasc_reverse_1}
\If{$u^*$ is not found}
    \State Use AdamBA($\epsilon, \frac{\|u^r - u^a\|}{4}, \Sigma, 1, u^a,  \frac{u^r - u^a}{\|u^r - u^a\|}, \textit{SAFE}, \textit{SAFE}$) to search for boundary point ${u^a}^*$
    \label{line:admbasc_reverse_2}
    \State Return ${u^a}^*$
\Else
    \State Return $u^*$
\EndIf
\EndProcedure
\end{algorithmic}
\end{algorithm}

Leveraging AdamBA and the safety index design rule, we construct the implicit safe set algorithm (ISSA) to safeguard the potentially unsafe reference control.  The proposed ISSA is summarized in \Cref{alg:adamsc}. Note that ISSA is presented under the context of discrete-time system with negligible sampling time, and $\mathcal{U}_S^D$ (line 3 of \Cref{alg:adamsc}) is the same as $\mathcal{U}_S$ when the sampling time is negligible.
 The key idea of ISSA is to use AdamBA for efficient search and grid sampling for worst cases. The inputs for ISSA are the approximation error bound ($\epsilon$), the learning rate ($\beta$), gradient vector covariance ($\Sigma$), gradient vector number ($n$) and reference unsafe control ($u^r$). Note that ISSA is only needed when the reference control is not safe. 

ISSA contains an offline stage and an online stage. In the offline stage, we synthesize the safety index according to the design rules. There are two major phases in the online stage for solving \eqref{eq:adamba_discrete}. In online-phase 1, we directly use AdamBA to find the safe controls on the boundary of $\mathcal{U}_S^D(x)$, and choose the control with minimum deviation from the reference control as the final output. In the case that no safe control is returned in online-phase 1 due to sparse sampling, online-phase 2 is activated. We uniformly sample the control space and deploy AdamBA again on these samples to find the safe control on the boundary of $\mathcal{U}_S^D(x)$. It can be shown in \Cref{appdx:proof2} that the ISSA algorithm is guaranteed to find a feasible solution of \eqref{eq:adamba_discrete} since the nonempty set of safe control is guaranteed by the safety index design rule in \eqref{eq:continuous_rule}. And we show in \Cref{sec:theory_contiuous} that the solution ensures \textit{forward invariance} 
in the set $\mathcal{S}=\mathcal{X}_S\cap \mathcal{X}_S^D$. Note that Phase 2 may consume significant computational resources, potentially compromising real-time performance. To avoid jerky movements caused by delayed computations, parallel processing and other acceleration techniques can be employed to speed up grid sampling.

Although the ISSA algorithm builds upon the safe set algorithm~\cite{liu2014control}, the proposed safety index synthesis and AdamBA algorithm can be applied to other \textit{energy function-based methods} to generate safe controls with or without an explicit analytical dynamics model. 


\section{Theoretical Results for ISSA When the Sampling Time is Negligible}
\label{sec:theory_contiuous}
\begin{theorem}[Forward Invariance for Continuous-Time System]
\label{thoem:main}
If the control system satisfies the conditions in \Cref{asm:ruleassumption} and \Cref{asm:continuous assumption}, and the safety index design follows the  rule described in \Cref{sec:synthesis}, the implicit safe set algorithm in \Cref{alg:adamsc} guarantees the \textit{forward invariance} 
to the set $\mathcal{S}\subseteq\mathcal{X}_S$. 
\end{theorem} 

To prove the main theorem, we introduce two important propositions to show that 1) the set of safe control is always nonempty if we choose a safety index that satisfies the design rule in \cref{sec:synthesis}; and 2) the proposed \cref{alg:adamsc} is guaranteed to find a safe control if there exists one.  With these two propositions, it is then straightforward to prove the \textit{forward invariance} 
to the set $S \subseteq \mathcal{X}_S$. 
In the following discussion, we discuss the two propositions in  \Cref{appdx:proof1} and \Cref{appdx:proof2}, respectively. 
Then, we prove \Cref{thoem:main} in \Cref{appdx:proofmain}.

\subsection{Feasibility of the Safety Index for Continuous-Time System}\label{appdx:proof1}

\begin{proposition}[Non-emptiness of the set of safe control]\label{prop:1}
If 1) the dynamic system satisfies the conditions in \Cref{asm:ruleassumption} and \Cref{asm:continuous assumption}; and 2) the safety index is designed according to the rule in  \Cref{sec:synthesis}, then the robot system in 2D plane has nonempty set of safe control at any state, i.e., $\mathcal{U}_S^D(x) \neq \emptyset, \forall x$.
\end{proposition}

Note that the set of safe control $\mathcal{U}_S^D(x):=\{u\in \mathcal{U}\mid \phi(f(x, u)) \leq \max\{\phi(x)-\eta, 0\} \}$ is non-empty if and only if it is non-empty in the following two cases: $\phi(x)-\eta < 0$ or $\phi(x)-\eta \geq 0$. In the following discussion, we first show that the safety index design rule guarantees a non-empty set of safe control if there's only one obstacle when $\phi(x)-\eta \geq 0$ (\Cref{lem:phi larger 0}). Then we show that the set of safe control is non-empty if there's only one obstacle when $\phi(x)-\eta < 0$ (\Cref{lem:phi less 0}). Finally, we leverage \Cref{lem:phi larger 0} and  \Cref{lem:phi less 0} to show $\mathcal{U}_S^D(x)$ is non-empty if there're multiple obstacles at any state.

\subsubsection{Preliminary Results}

\begin{lemma}
If the dynamic system satisfies the conditions in \Cref{asm:ruleassumption} and \Cref{asm:continuous assumption} and there is only one obstacle in the environment, then the safety index design rule in \Cref{sec:synthesis} ensures that $\mathcal{U}_S^D(x)\neq \emptyset$ for $x$ such that $\phi(x)-\eta \geq 0$.
\label{lem:phi larger 0}
\end{lemma}
\begin{proof}
For $x$ such that $\phi(x)-\eta \geq 0$, the set of safe control becomes
\begin{align}
    \label{condition on phi}
    \mathcal{U}_S^D(x) = \{u\in\mathcal{U}\mid \phi(f(x, u)) \leq \phi(x)-\eta\}
\end{align}

According to the first condition in \Cref{asm:continuous assumption}, we have $dt \rightarrow 0$. Therefore, the discrete-time approximation error approaches zero, i.e., $\phi(f(x,u)) = \phi(x) + dt\cdot \dot\phi(x,u) + \Delta$, where 
$\Delta \rightarrow 0$. Then we can rewrite \eqref{condition on phi} as:
\begin{align}
    \label{new condition on phi}
    \mathcal{U}_S^D(x) = \{u\in\mathcal{U}\mid \dot\phi \leq -\eta/dt\}
\end{align}

According to \eqref{eq: safety index}, $\dot\phi=-nd^{n-1}\dot d -k \ddot d$. We ignored the subscript $i$ since it is assumed that there is only one obstacle. Therefore, the non-emptiness of $\mathcal{U}_S^D(x)$ in \eqref{new condition on phi} is equivalent to the following condition
\begin{align}
    \label{eq:origin_condtion_lemma1}
    \forall x \text{ s.t. } \phi(x)\geq \eta, \exists u, \text{s.t. } \ddot d \geq \frac{\eta/dt - n d^{n-1}\dot d}{k}.
\end{align}

\rebuttal{


Note that in the 2D problem, $\ddot d = -a\cos(\alpha) + v\sin{(\alpha)}w$ and $\dot d = -v\cos(\alpha)$. According to \Cref{asm:ruleassumption}, there is a surjection from $u$ to $(a,w)\in W:=\{(a,w)\mid a_{min}\leq a\leq a_{max}, w_{min}\leq w\leq w_{max}\}$. Moreover, according to \eqref{eq: safety index}, $\phi$ for the 2D problem only depends on $\alpha$, $v$, and $d$. Hence  condition $\forall x \text{ s.t. } \phi(x)\geq \eta$ can be translated to $\forall (\alpha,v,d) \text{ s.t. } \sigma+d_{min}^n - d^n - kv\cos(\alpha) \geq \eta $. Denote the later set as
\begin{align}
    \Phi:=\{(\alpha,v,d) \mid \sigma+d_{min}^n - d^n - kv\cos(\alpha) \geq \eta, v\in[0,v_{max}], d\geq 0, \alpha \in[0,2\pi)\}.
\end{align} 
Consequently,  condition \eqref{eq:origin_condtion_lemma1} is equivalent to the following condition}
\begin{align}\label{eq:ori_condition1}
    \forall (\alpha, v,d)\in\Phi, \exists (a,w)\in W, \text{ s.t. }-a \cos(\alpha) + v \sin(\alpha) w & \geq  \frac{\eta/dt + n d^{n-1} v \cos(\alpha)}{k}.
\end{align}

According to the safety index design rule, we have $\eta = 0$. Then we show \eqref{eq:ori_condition1} holds in different cases. 

\textbf{Case 1}: $v=0$. In this case, we can simply choose $a=0$, then the inequality in \eqref{eq:ori_condition1} holds.

\textbf{Case 2}: $v\neq 0$ and $\cos(\alpha) \leq 0$. 
Note that velocity $v$ is always non-negative. Hence $v>0$. In this case, we just need to choose $a=w=0$, then the inequality in \eqref{eq:ori_condition1} holds, where the LHS becomes zero and the RHS becomes $\frac{nd^{n-1}v\cos(\alpha)}{k}$ which is non-positive.

\textbf{Case 3}: $v\neq 0$ and $\cos(\alpha)>0$. 
Dividing $v\cos(\alpha)$ on both sides of the inequality and rearranging the inequality,  \eqref{eq:ori_condition1} is equivalent to
\begin{align}\label{eq:closer_1}
    \forall (\alpha, v,d)\in\Phi, \exists (a,w)\in W, \text{ s.t. }- \frac{a}{v} + \tan(\alpha) w - \frac{ n d^{n-1}}{k}\geq 0,
\end{align}
and \eqref{eq:closer_1} can be verified by showing: 
\begin{align}\label{eq:closer_2}
    \min_{(\alpha,v,d)\in\Phi} \max_{(a,w)\in W} (- \frac{a}{v} + \tan(\alpha) w-\frac{ n d^{n-1}}{k}) \geq 0.
\end{align}

Now let us expand the LHS of \eqref{eq:closer_2}:
\begin{align}
    &\min_{(\alpha,v,d)\in\Phi} \max_{(a,w)\in W} (- \frac{a}{v} + \tan(\alpha) w-\frac{ n d^{n-1}}{k})\\
   = &\min_{(\alpha,v,d)\in\Phi}  (- \frac{a_{min}}{v} + [\tan(\alpha)]_{+} w_{max} +[\tan(\alpha)]_{-} w_{min} -\frac{ n d^{n-1}}{k})\\
   = &\min_{\alpha\in[0,2\pi),v\in(0,v_{max}]}  (- \frac{a_{min}}{v} + [\tan(\alpha)]_{+} w_{max} +[\tan(\alpha)]_{-} w_{min} -\frac{ n (\sigma+d_{min}^n+kv\cos(\alpha))^{\frac{n-1}{n}}}{k})\\
   = & -\frac{a_{min}}{v_{max}}-\frac{ n (\sigma+d_{min}^n+kv_{max})^{\frac{n-1}{n}}}{k}.\label{eq: LHS in case 3}
\end{align}
The first equality eliminates the inner maximization where $[\tan(\alpha)]_+ :=\max\{\tan(\alpha),0\}$ and $[\tan(\alpha)]_- :=\min\{\tan(\alpha),0\}$. The second equality eliminates $d$ according to the constraint in $\Phi$. The third equality is achieved when $\alpha = 0$ and $v=v_{max}$. According to the safety index design rule in \Cref{sec:synthesis}, \eqref{eq: LHS in case 3} is greater than or equal to zero. Hence \eqref{eq:closer_2} holds, which then implies that the inequality in \eqref{eq:ori_condition1} holds.


The three cases cover all possible situations. Hence \eqref{eq:ori_condition1} always hold and the claim in the lemma is verified.
\end{proof}

\begin{lemma}
If the dynamic system satisfies the contidions in \Cref{asm:ruleassumption} and \Cref{asm:continuous assumption} and there is only one obstacle in the environment, 
then the safety index design rule in \Cref{sec:synthesis} ensures that $\mathcal{U}_S^D(x)=\mathcal{U}$ for any $x$ that $\phi(x)-\eta < 0$.
\label{lem:phi less 0}
\end{lemma}
\begin{proof}
For $x$ such that $\phi(x)-\eta < 0$, the set of safe control becomes
\begin{align}
    \label{condition on phi lemma2}
    \mathcal{U}_S^D(x)=\{u\in\mathcal{U}\mid\phi(f(x, u)) \leq 0\}
\end{align}

According to the first assumption in \Cref{asm:continuous assumption}, we have $dt \rightarrow 0$. Therefore, the discrete-time approximation error approaches zero, i.e., $\phi(f(x,u)) = \phi(x) + dt\cdot \dot\phi(x,u) + \Delta$, where 
$\Delta \rightarrow 0$. Then we can rewrite \eqref{condition on phi lemma2} as:
\begin{align}
    \label{new condition on phi lemma2}
    \mathcal{U}_S^D(x)=\{u\in\mathcal{U}\mid\dot\phi \leq -\phi/dt\}
\end{align}

Note that $\eta = 0$ according to the safety index design rule, then $\phi(x) - \eta < 0$ implies that $\phi(x) < 0$. Hence $-\phi/dt\rightarrow \infty$ since $dt\rightarrow 0$. Then as long as $\dot\phi$ is bounded, $\mathcal{U}_S^D(x)=\mathcal{U}$.

Now we show that $\dot\phi$ is bounded. 
According to \eqref{eq: safety index}, $\dot\phi=-nd^{n-1}\dot d -k \ddot d$. We ignored the subscript $i$ since it is assumed that there is only one obstacle. 
According to \Cref{asm:ruleassumption}, we have the state space is bounded, thus both $d$ and $\dot d$ are bounded, which implies that $n d^{n-1}\dot d$ is bounded.
Moreover, 
we have for all possible values of $a$ and $w$, there always exists a control $u$ to realize such $a$ and $w$ according to \Cref{asm:ruleassumption}, which indicates the mapping from $u$ to $(a, w)$ is surjective. Since $a$ and $w$ are bounded and both can achieve zeros according to \Cref{asm:ruleassumption}, we have $\forall u$, the corresponding $(a, w)$ are bounded. Since $\ddot d = -a \cos(\alpha) + v \sin(\alpha) w$, then $ \ddot d$ is bounded. Hence $\mathcal{U}_S^D(x)=\mathcal{U}$ any $x$ that $\phi(x)-\eta < 0$ and the claim is true.
\end{proof}

\subsubsection{Proof of \Cref{prop:1}}
\begin{proof}
If there is one obstacle, then \cref{lem:phi larger 0} and \cref{lem:phi less 0} have proved that $\mathcal{U}_S^D(x)\neq \emptyset$ for all $x$. Now we need to consider the case where there are more than one obstacle but the environment is sparse in the sense that at any time step, there is at most one obstacle which is safety critical, i.e. $\phi_i\geq 0$. To show nonemptiness of $\mathcal{U}_S^D(x)$, we consider the following two cases. In the following discussion, we set $\eta=0$ according to the safety index design rule.

\textbf{Case 1}: $\phi(x) = \max_i\phi_i(x)\geq 0$. Denote $j :=\arg\max_i\phi_i(x)$. Since there is at most one obstacle that is safety critical, then $\phi_j(x) \geq 0$ and $\phi_k(x)<0$ for all $k\neq j$.
Denote ${\mathcal{U}_S^D}_j(x):=\{u\in \mathcal{U}\mid \phi_j(f(x, u)) \leq \phi_j(x) \}$. \Cref{lem:phi larger 0} ensures that ${\mathcal{U}_S^D}_j(x)$ is nonempty. 
Denote ${\mathcal{U}_S^D}_k(x):=\{u\in \mathcal{U}\mid \phi_k(f(x, u)) \leq 0 \}$ where $k \neq j$. Since $\phi_k(x) < 0$, \cref{lem:phi less 0} ensures that ${\mathcal{U}_S^D}_k(x) = \mathcal{U}$. 

Note that the set of safe control can be written as:
\begin{subequations}
\begin{align}
    {\mathcal{U}_S^D}(x)&:=\{u\in \mathcal{U}\mid \max_i\phi_i(f(x, u)) \leq \max_i\phi_i(x) \}\\
    &=\{u\in \mathcal{U}\mid \max_i\phi_i(f(x, u)) \leq \phi_j(x) \}\\
    &=\cap_i \{u\in \mathcal{U}\mid \phi_i(f(x, u)) \leq \phi_j(x) \}\\
    &= {\mathcal{U}_S^D}_j(x) \neq \emptyset
\end{align}
\end{subequations}
Note that the last equality is due to the fact that $\{u\in \mathcal{U}\mid \phi_i(f(x, u)) \leq \phi_j(x) \}\supseteq \{u\in \mathcal{U}\mid \phi_i(f(x, u)) \leq 0 \} = \mathcal{U}\supseteq {\mathcal{U}_S^D}_j(x)$ for $i\neq j$. 

\textbf{Case 2}: $\phi(x)= \max_i \phi_i(x) < 0$. Therefore, we have $\phi_i(x) < 0$ for all $i$. 
According to \Cref{lem:phi less 0}, $\{u\in\mathcal{U}\mid \phi_i(f(x,u))\leq 0\} = \mathcal{U}$. Hence the set of safe control satisfies the following relationship
\begin{subequations}
\begin{align}
        {\mathcal{U}_S^D}(x)&:=\{u\in \mathcal{U}\mid \max_i\phi_i(f(x, u)) \leq 0 \}\\
    &=\cap_i \{u\in \mathcal{U}\mid \phi_i(f(x, u)) \leq 0 \}\\
    &= \mathcal{U} \neq \emptyset
\end{align}
\end{subequations}

In summary, $\mathcal{U}_S^D(x) \neq \emptyset, \forall x$ and the claim holds.
\end{proof}

\subsection{Feasibility of ISSA}\label{appdx:proof2}
\begin{proposition}[Feasibility of \Cref{alg:adamsc}]\label{prop:2}
If the set of safe control is non-empty, \Cref{alg:adamsc} can always find a sub-optimal solution of \eqref{eq:adamba_discrete} with a finite number of iterations. 
\end{proposition}

 \Cref{alg:adamsc} executes two phases consecutively where the second phase will be executed if no solution of  \eqref{eq:adamba_discrete} is returned in the first phase. Hence, \Cref{alg:adamsc} can always find a sub-optimal solution of \eqref{eq:adamba_discrete} (safe control on the boundary of $\mathcal{U}_S^D$) if the solution of \eqref{eq:adamba_discrete} can always be found in phase 2. 
 
 Note that Phase 2 first finds an anchor safe control $u^a$, then use it with AdamBA (\Cref{alg:adamba}) to find the solution of  \eqref{eq:adamba_discrete}. In the following discussion, we first show that the safety index design rule guarantees $u^a$ can be found with finite iterations (\Cref{lem:finiteness}). Then we show that AdamBA will return a solution if it enters the \textit{exponential decay} phase (\Cref{lem:convergence}). Subsequently, we show that the evoked AdamBA procedures in phase 2 will definitely enter \textit{exponential decay} phase (\Cref{lem:feasibility}). Finally, we provide a upper bound of the computation iterations for \Cref{alg:adamsc} at \Cref{sec: proof of prop 2}.

\subsubsection{Preliminary Results.}

\begin{lemma}[Existence]
If the synthesized safety index can guarantee a non-empty set of safe control, then we can find an anchor point in phase 2 of \Cref{alg:adamsc} with finite iterations (line \ref{line grid sampling} in \cref{alg:adamsc}).
\label{lem:finiteness}
\end{lemma}

\begin{proof}

\begin{figure}
    \centering
    \includegraphics[width=0.5\textwidth]{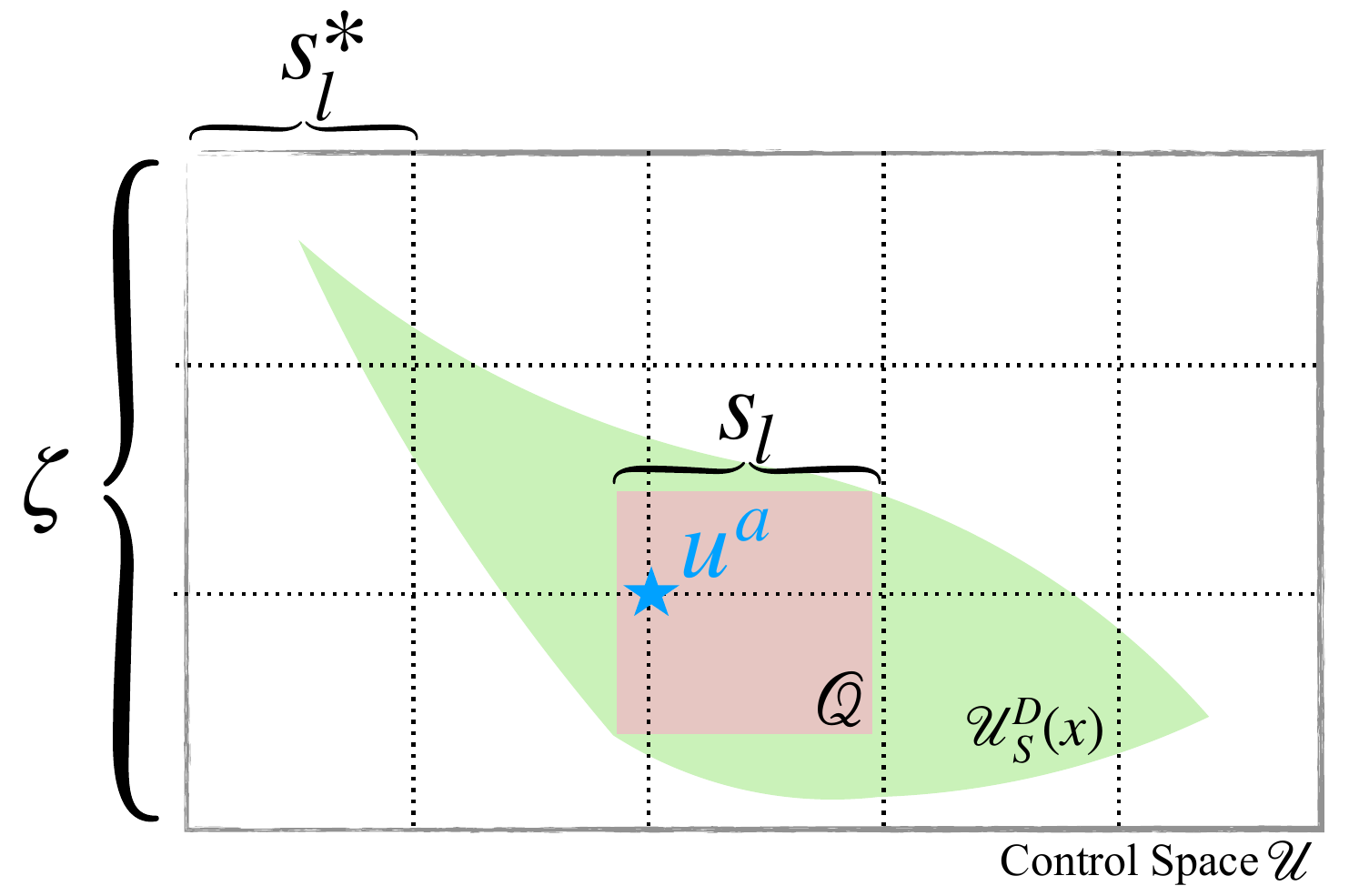}
    \caption{Illustration of the grid sampling to find an anchor control point.}
    \label{fig:lemma_2}
\end{figure}

If the synthesized safety index guarantees a non-empty set of safe control $\mathcal{U}^D_S$, then there exists a hypercube inside of $\mathcal{U}^D_S$, i.e. $\exists \mathcal{Q} \subset \mathcal{U}_S^D$, where $\mathcal{Q}$ is a $n_u$-dimensional hypercube with the same side length of $s_l > 0$.
Denote $\zeta_{[i]} = \max_{j,k}\| u_{[i]}^j -  u_{[i]}^k \|$, where $u_{[i]}$ denotes the $i$-th dimension of control $u$, and $u^j \in \mathcal{U}^D_S,  u^k \in \mathcal{U}^D_S$. 

By directly applying grid sampling in $\mathcal{U}^D_s$ with sample interval $s_l^*$ at each control dimension, such that $ 0 < s_l^* < s_l$, the maximum sampling iteration $T^a$ for finding an anchor point in phase 2 satisfies the following condition:
\begin{equation}\label{eq:max_sample}
    T^a < \prod\limits_{i=1}^{n_u} \lceil\frac{\zeta_{[i]}}{s^*_l}\rceil~,
\end{equation}
where $T^a$ is a finite number since $s_l^* > 0$. Then we have proved that we can find an anchor point in phase 2 of \Cref{alg:adamsc} with finite iteration (i.e., finite sampling time). The grid sampling to find an anchor control point is illustrated in \Cref{fig:lemma_2}.

\end{proof}

\begin{lemma}[Convergence]
If AdamBA enters the \textit{exponential decay} phase (line \ref{line:16} in \cref{alg:adamba}), then it can always return a boundary point approximation (with desired safety status) where the approximation error is upper bounded by $\epsilon$.
\label{lem:convergence}
\end{lemma}

\begin{proof}
    According to \Cref{alg:adamba}, \textit{exponential decay} phase applies Bisection method to locate the boundary point $P_b$ until $\|P_{NS} - P_S\| < \epsilon$. Denote the returned approximated boundary point as $P_{return}$, according to line \ref{line output}, $P_{return}$ is either $P_{NS}$ or $P_S$, thus the approximation error satisfies:
\begin{align}
    \|P_{return} - P_b\| &\leq \max\{\|P_{NS} - P_b\|,  \| P_{S} - P_b \|\} \\ \nonumber 
    &\leq \|P_{NS} - P_S\| \\ \nonumber
    & < \epsilon.
\end{align}

\end{proof}

\begin{lemma}[Feasibility]
If we enter the phase 2 of \Cref{alg:adamsc} with an anchor safe control being sampled, we can always find a local optimal solution of \eqref{eq:adamba_discrete}. 
\label{lem:feasibility}
\end{lemma}

\begin{proof}

According to line \ref{line:admbasc_reverse_1}-\ref{line:admbasc_reverse_2}, after an anchor safe control is being sampled, phase 2 of \Cref{alg:adamsc} will evoke at most two AdamBA processes. Hence, \Cref{lem:feasibility} can be proved by showing one of the two AdamBA will return a local optimal solution for \eqref{eq:adamba_discrete}. Next we show \Cref{lem:feasibility} holds in two cases.
\paragraph{Case 1: } line \ref{line:admbasc_reverse_1} of \Cref{alg:adamsc} finds a solution. 

In this case, the first AdamBA process finds a safe control $u^*$ solution (the return of AdamBA is a set, whereas the set here has at most one element). According to \Cref{alg:adamba}, a solution will be returned only if AdamBA enters \textit{exponential decay} stage. Hence, according to \Cref{lem:convergence}, $u^*$ is close to the boundary of the set of safe control with approximation error upper bounded by $\epsilon$.

\begin{figure}[h]
    \centering
    \includegraphics[width=0.5\textwidth]{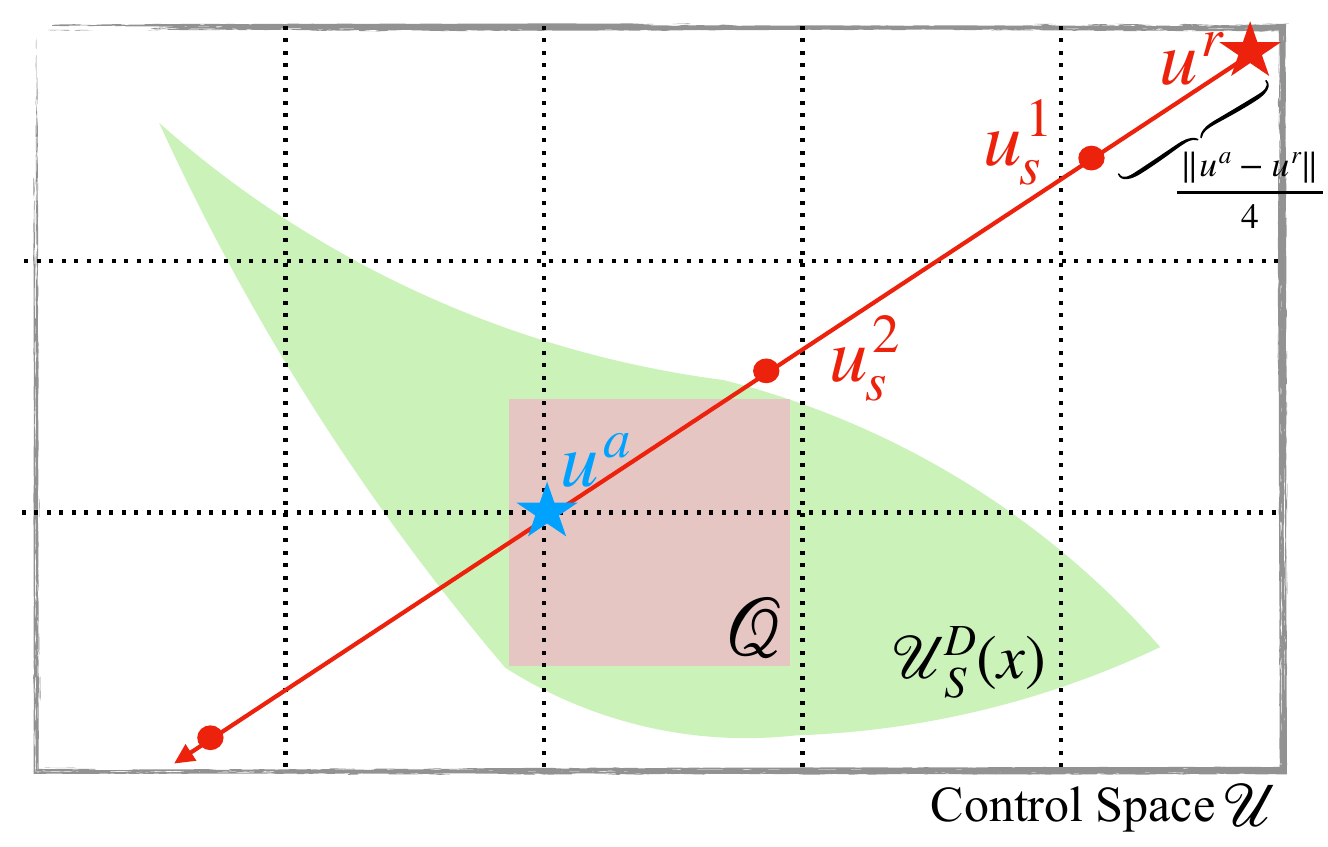}
    \caption{Illustration of the case when it is unable to find $u^*$.}
    \label{fig:no_ustar}
\end{figure}

\paragraph{Case 2: } line \ref{line:admbasc_reverse_1} of \Cref{alg:adamsc} fails to find a solution.

In this case, the second AdamBA process is evoked (line \ref{line:admbasc_reverse_2} of \Cref{alg:adamsc}). Since no solution is returned from the first AdamBA process (\ref{line:admbasc_reverse_1} of \Cref{alg:adamsc}), where we start from $u^r$ and exponentially outreach along the direction $\vec{v}_a = \frac{u^a - u^r}{||u^a - u^r||}$, then all the searched control point along $\vec{v}_a$ is \textit{UNSAFE}. 

Specifically, we summarize the aforementioned scenario in \Cref{fig:no_ustar}, the searched control points are represented as red dots along $\vec{v}_a$ (red arrow direction). Note that the exponential outreach starts with step size $\beta = \frac{\|u^a - u^r\|}{4}$, indicating two points $\{u_s^1$, $u_s^2\}$ are sampled between $u^r$ and $u^a$ such that 
\begin{equation}
    \begin{cases}
        u_s^1 = u^r + \frac{\|u^a - u^r\|}{4}\vec{v}_a = \frac{3u^r}{4} + \frac{u^a}{4} \\
        u_s^2 = u^r + \frac{3\|u^a - u^r\|}{4}\vec{v}_a = \frac{u^r}{4} + \frac{3u^a}{4}
    \end{cases}
\end{equation}
where the safety statuses of both $u_s^1$ and $u_s^2$ are \textit{UNSAFE}.

During the second AdamBA process (line \ref{line:admbasc_reverse_2} of \Cref{alg:adamsc}), we start from $u^a$ and exponentially outreach along the direction $\vec{v}_r = \frac{u^r - u^a}{||u^r - u^a||}$ with initial step size $\beta = \frac{\|u^r - u^a\|}{4}$. Hence, denote $\bar{u}_s^1$ as the first sample point along $\vec{v}_r$, and $\bar{u}_s^1$ satisfies
\begin{align}
    \bar{u}_s^1 = u^a + \frac{\|u^r - u^a\|}{4}\vec{v}_r = \frac{u^r}{4} + \frac{3 u^a}{4},
\end{align}

which indicates $\bar{u}_s^1 = u_s^2$, and the safety status of $\bar{u}_s^1$ is \textit{UNSAFE}. Since the safe status of $u^a$ is \textit{SAFE} and a \textit{UNSAFE} point can be sampled during the \textit{exponential outreach} stage, the second AdamBA process (line \ref{line:admbasc_reverse_2} of \Cref{alg:adamsc}) will enter \textit{exponential decay} stage. Therefore, according to \Cref{lem:convergence}, ${u^a}^*$ will always be returned and ${u^a}^*$ is close to the boundary of the set of safe control with approximation error upper bounded by $\epsilon$.

The two cases cover all possible situations. Hence, after an anchor safe control $u^a$ is sampled, phase 2 of \Cref{alg:adamsc} can always find a local optima of \eqref{eq:adamba_discrete}.

\end{proof}

\subsubsection{Proof of \Cref{prop:2}}
\label{sec: proof of prop 2}
\begin{proof}

According to \Cref{lem:finiteness} and \Cref{lem:feasibility}, \Cref{alg:adamsc} is able to find local optimal solution of \eqref{eq:adamba_discrete}. Next, we will prove \Cref{alg:adamsc} can be finished within finite iterations.

According to \Cref{alg:adamsc}, ISSA include  (i) one procedure to find anchor safe control $u_a$, and (ii) at most three AdamBA procedures. Firstly, based on \Cref{lem:finiteness}, $u_a$ can be found within finite iterations. Secondly, each AdamBA procedure can be finished within finite iterations due to:
\begin{itemize}
    \item \textit{exponential outreach} can be finished within finite iterations since the control space is bounded.
    \item \textit{exponential decay} can be finished within finite iterations since Bisection method  will exit within finite iterations.
\end{itemize}

Therefore, ISSA can be finished within finite iterations.

\end{proof}

\subsection{Proof of \Cref{thoem:main}}
\label{appdx:proofmain}

Before we prove \Cref{thoem:main}, we start with a preliminary result regarding $\mathcal{X}_S^D$ that is useful for proving the main theorem: 

\begin{lemma}[Forward Invariance of $\mathcal{X}_S^D$]
If the control system satisfies \Cref{asm:ruleassumption} and \Cref{asm:continuous assumption}, and the safety index design follows the rule described \Cref{sec:synthesis}, the implicit safe set algorithm guarantees the \textit{forward invariance} 
to the set $\mathcal{X}_S^D$. 
\label{lem:fi_safety_index}
\end{lemma}
\begin{proof}
If the control system satisfies the assumptions in \Cref{asm:ruleassumption} and \Cref{asm:continuous assumption}, and the safety index design follows the rule described \Cref{sec:synthesis}, then we can ensure the system has a nonempty set of safe control at any state by \Cref{prop:1}. By \Cref{prop:2}, the implicit safe set algorithm can always find a local optima solution of \eqref{eq:adamba_discrete}. The local optima solution always satisfies the constraint $\phi(f(x_t, u_t)) \leq \max\{\phi(x_t)-\eta, 0\}$, which indicates that 1) if $\phi(x_{t_0}) \leq 0$, then $\phi(x_t) \leq 0, \forall t \geq t_0$. 
Note that $\phi(x) \leq 0$ demonstrates that $x \in \mathcal{X}_S^D$.
\end{proof}

Now the proof the \Cref{thoem:main} is detailed below. 
\begin{proof}

Leveraging \Cref{lem:fi_safety_index}, we then proceed to prove that the \textit{forward invariance} 
to the set $\mathcal{X}_S^D$ guarantees the \textit{forward invariance} 
to the set $S \subseteq \mathcal{X}_S$. Recall that $\mathcal{S}=\mathcal{X}_S\cap \mathcal{X}_S^D$. Depending on the relationship between $\mathcal{X}_S^D$ and $\mathcal{X}_S$, there are two cases in the proof which we will discuss below.

\textbf{Case 1: $\mathcal{X}_S^D = \{x|\phi(x) \leq 0\}$ is a subset of $\mathcal{X}_S = \{x|\phi_0(x) \leq 0\}$.}

In this case, $\mathcal{S} = \mathcal{X}_S^D$. 
According to \Cref{lem:fi_safety_index}, If the control system satisfies the assumptions in \Cref{asm:ruleassumption} and \Cref{asm:continuous assumption}, and the safety index design follows the rule described \Cref{sec:synthesis}, the implicit safe set algorithm guarantees the \textit{forward invariance} 
to the set $\mathcal{X}_S^D$ and hence $\mathcal{S}$. 

\textbf{Case 2: $\mathcal{X}_S^D = \{x|\phi(x) \leq 0\}$ is NOT a subset of $\mathcal{X}_S = \{x|\phi_0(x) \leq 0\}$.}

\newtheory{
In this case, if $x_t \in \mathcal{S}$, we have $\phi_0(x_t) = \max_i {\phi_0}_i(x_t) \leq 0$, which indicates $\forall i, {\phi_0}_i \leq 0$. 

Firstly, we consider the case where ${\phi_0}_i(x_t) < 0$. Note that ${\phi_0}_i(x_{t+1}) = {\phi_0}_i(x_t) + \dot {\phi_0}_i(x_t)dt + \frac{\ddot {\phi_0}_i(x_t)dt^2}{2!} + \cdots$, since the state space and control space are both bounded, and $dt \rightarrow 0$ according to \Cref{asm:ruleassumption}, we have ${\phi_0}_i(x_{t+1}) \rightarrow {\phi_0}_i(x_t) \leq 0$.


Secondly, we consider the case where ${\phi_0}_i(x_t) = 0$. Since $x_t \in \mathcal{S}$, we have $\max_i \phi_i(x_t) \leq 0$, which indicates $\forall i, \sigma + d_{min}^n - d_i^n - k\dot d_i \leq 0$. Since ${\phi_0}_i(x_t) = 0$, we also have $d_i = d_{min}$. Therefore, the following condition holds: 
\begin{align}
    \sigma - k \dot d_i &\leq 0 \\ \nonumber
    \dot d_i &\geq \frac{\sigma}{k}
\end{align}
According to the safety index design rule, we have $k, \sigma \in \mathbb{R}^+$, which indicates $\dot d_i > 0$. Therefore, we have ${\phi_0}_i(x_{t+1}) < 0$.


Summarizing the above two cases, we have shown that if ${\phi_0}_i(x_t) \leq 0$ then ${\phi_0}_i(x_{t+1}) \leq 0$, which indicates if $\forall i, {\phi_0}_i(x_t) \leq 0$ then  $\forall i, {\phi_0}_i(x_{t+1}) \leq 0$. Note that $\forall i, {\phi_0}_i(x_{t+1}) \leq 0$ indicates that $\phi_0(x_{t+1}) = \max_i {\phi_0}_i(x_{t+1}) \leq 0$. Therefore, we have that if $x_t \in \mathcal{S}$ then $x_{t+1} \in  \mathcal{X}_S$. Thus, we also have $x_{t+1} \in \mathcal{S}$ by \Cref{lem:fi_safety_index}. By induction, we have if $x_{t_0} \in \mathcal{S}$, $x_t \in \mathcal{S}, \forall t > t_0$.

In summary, by discussing the two cases of whether $\mathcal{X}_S^D$ is the subset of $\mathcal{X}_S$, we have proven that if the control system satisfies the assumptions in \Cref{asm:ruleassumption} and \Cref{asm:continuous assumption}, and the safety index design follows the rule described in \Cref{sec:synthesis}, the implicit safe set algorithm guarantees the \textit{forward invariance} 
to the set $\mathcal{S}\subseteq\mathcal{X}_S$. }

\end{proof}


\section{Implicit Safe Set Algorithm for Systems with Non-Negligible Sampling Time}
\label{sec:discrete method}

The implicit safe set algorithm discussed in \Cref{sec:method} is for systems with negligible sampling time ($dt \rightarrow 0$). 
However, in reality, many systems are implemented in a discrete-time manner with non-negligible sampling time (e.g., digital twins, physical simulators), which means that the controller may not be able to respond immediately to potential violations of safety. What is even worse is that the safe control of continuous-time systems may lead to safety violations in discrete-time systems.
We demonstrate a toy problem in \Cref{appdx:toy} showing that the safe control of continuous-time systems (safe control from \eqref{eq: quadratic program for control}) actually violates the safety constraints when it is applied in discrete-time systems (conditions in \eqref{eq:adamba_discrete}). 
That is mainly because the higher order terms $O(dt^2)$ is not negligible in the transition from $\phi(x_t)$ to $\phi(x_{t+1})$ when $dt$ is not sufficiently small:
\begin{equation}\label{eq:discrepancy}
\begin{split}
     \phi(x_{t+1}) = \phi(x_t) + \dot \phi(x_t) dt  + O(dt^2), \dot\phi(x_t) = \lim_{\tau \rightarrow 0^+}\frac{\phi(x_{t+\tau/dt}) - \phi(x_t)}{\tau}.
\end{split}
\end{equation}

Therefore, the assumptions concerning sparse obstacle environment in \Cref{asm:continuous assumption} should also be justified. When $dt\rightarrow 0$, $\phi_{t+1} \rightarrow \phi_t$ according to \eqref{eq:discrepancy}, thus the safety-critical condition should be $\phi \ge 0$. However, when $dt$ is non-negligible, there will be a gap between $\phi_{t+1}$ and $\phi_t$ which can be represented as $\Delta_{\phi_{max}}$. Then the safety-critical condition should be $\phi + \Delta_{\phi_{max}} \ge 0$, where $\Delta_{\phi_{max}}$ is the maximum possible $\phi$ change magnitude at one time step, i.e. $\Delta_{\phi_{max}} = \max_{x,u,t} |\phi_{t+1} - \phi_{t}|$.

In fact, the continuous-time system can be seen as a special case of discrete-time systems where $dt \rightarrow 0$. To that end, studying the method of safe control in discrete-time systems is essential towards the real-world application of DRL.  

This section introduces the implicit safe set algorithm (ISSA) for discrete-time systems, which extends the ISSA introduced in \Cref{sec:method} to be applicable for systems with non-negligible sampling time. Specifically, ISSA for discrete-time systems shares the similar pipeline as introduced in \Cref{sec:method}:
\begin{itemize}
    \item \textbf{Offline:} design a safety index that ensures $\mathcal{U}_S^D(x)$ is nonempty for all $x$ for discrete-time system. 
    \item \textbf{Online:} project the nominal control to $\mathcal{U}_S^D(x)$ via a sample-efficient black-box optimization method during online robot maneuvers.
\end{itemize}

\subsection{Safety Index Design Rule for Discrete-Time System}
\label{sec:synthesis_discrete}

 We start our discussion by highlighting two critical restrictions of the safety index design rule for the continuous-time system from \Cref{sec:synthesis} as follows:
 \begin{itemize}
     \item the safety index design rule from \Cref{sec:synthesis} sets $\eta = 0$, hence the robot is unable to converge to the safe set (\textit{finite time convergence}) via solving \eqref{eq:adamba_discrete} if the robot starts from the unsafe state, i.e. $\phi(x_{t_0}) > 0$.
     \item the safety index design rule from \Cref{sec:synthesis} relies on the negligible sampling time assumption to guarantee the non-emptiness of the set of safe control, which doesn't hold in discrete-time system.
 \end{itemize}
 

In the following discussions, we will introduce the safety index design rule for the general discrete-time system that addresses the aforementioned restrictions. 

\subsubsection{Assumption}
The discrete-time system safety index for collision avoidance in 2D will be synthesized without referring to the specific dynamic model, but under the following assumptions.

\begin{asm}[2D Collision Avoidance for Discrete-Time System]
1) The discrete-time system time step satisfies the condition: $\frac{a_{min}}{2} + \frac{v_{max}}{4dt} > (a_m + v_{max}w_m)(-\frac{a_{min}}{v_{max}} + w_{m})dt$, where $a_m = \max\{-a_{min}, a_{max}\}$ and $w_m = \max\{-w_{min}, w_{max}\}$;
2) The acceleration and angular velocity keep constant between two consecutive time steps, i.e. $a_{[t:t+1)} = a_t$.
3) At  any  given  time,  there  can  at  most  be  one  obstacle  becoming  safety-critical, such that $\phi \geq - \Delta_{\phi_{max}}$ where $\Delta_{\phi_{max}}= \max_{x,u,t} |\phi_{t+1} - \phi_{t}|$ is the maximum possible $\phi$ change magnitude at one time step (Sparse Obstacle Environment). 
\label{asm:discrete assumption}
\end{asm}

These assumptions are easy to meet in practice. The first assumption ensures a bounded sampling time for discrete-time systems, which will be leveraged for the theoretical proofs in \Cref{sec:theory_discrete}. The second assumption assumes zero-order hold (ZOH) for discrete-time control signals. The third assumption 
enables safety index design rule applicable with multiple moving obstacles in discrete-time system. $\Delta_{\phi_{max}}$ is bounded mainly because the system dynamics are bounded and the sampling time is also bounded. 

\subsubsection{Safety Index Design Rule}
\label{sec:SI for discrete system}
Following the rules in \cite{liu2014control}, we parameterize the safety index as $\phi = \max_i\phi_i$, and $\phi_i = \sigma + d_{min}^n - d_i^n - k\dot d_i$, where all $\phi_i$ share the same set of tunable parameters $\sigma,n,k,\eta \in \mathbb{R}^+$. Our goal is to choose these parameters such that $\mathcal{U}_S^D(x)$ is always nonempty for all possible states in the discrete-time system. 
Although similar to the safety index design rule in continuous-time systems, the safety index design rule for discrete-time systems is updated mainly from three aspects. 
\begin{itemize}
     \item We use nontrivial positive $\sigma$ as an safety margin to enable the \textit{forward invariance} to the invariantly safe set $\mathcal{S}$ for discrete-time systems.
    \item We specifically pick $n=1$ to simplify the analysis for the higher order terms of \eqref{eq:discrepancy} (approximation error).
    \item We use nontrivial positive $\eta$  to enable the \textit{finite time convergence} to the invariantly safe set $\mathcal{S}$ when the discrete-time system starts from initially unsafe states or initially safe but inevitably unsafe states in $\mathcal{X}_S$.
\end{itemize}

\textbf{Safety Index Design Rule for Discrete-Time System:} 
By setting $n = 1$ and introducing an auxiliary parameter $\eta_0 \in \mathbb{R}^+$, the parameters $\sigma, \eta_0, k$ should be chosen such that 
\begin{subequations}
    \begin{align}
        \sigma &> - \dot{d}^*_{min} dt \label{eq: rule 0}\\
        \frac{\frac{\eta_0}{dt} + v_{\max}}{k} &\leq \min\{-a_{min}, a_{max}\}~, \label{eq: rule 1}
    \end{align}
\end{subequations}
where $\dot{d}^*_t = \frac{d_{t+1} - d_{t}}{dt}$, and $\dot{d}^*_{min}$ denotes the minimum achievable $\dot{d}_t^*$ in the system. Then the parameter $\eta$ is set as
\begin{align}
    \eta &= \eta_0|cos(\alpha)|~. \label{eq: rule 2}
\end{align}


For the discrete-time system, we will first synthesize the $\eta_0, n, \sigma, k$ offline first. During the online execution, $\eta$ is assigned on-the-fly based on $\alpha$ using \eqref{eq: rule 2}. We will show that the safety index design rule is valid to guarantee non-empty $\mathcal{U}_S^D(x)$ for discrete-time systems in \Cref{prop:discrete non-empty}.

\subsection{Implicit Safe Set Algorithm for Discrete-Time System}
\label{sec:ISSA_discrete}

The ultimate goal of the implicit safe set algorithm for discrete-time systems is to ensure (i) \textit{forward invariance} in $\mathcal{S} = \mathcal{X}_S\cap \mathcal{X}_S^D$ when systems start from safe states, and (ii) \textit{finite time convergence} to $\mathcal{S}$ when systems start from unsafe states. The discrete-time implicit safe set algorithm starts by applying ISSA (\Cref{alg:adamsc}), then applies an additional convergence trigger (CTrigger) algorithm to filter ISSA solution when $|\cos(\alpha)|$ is small and $\phi(x) > 0$. 

The main body of CTrigger algorithm is summarized in \Cref{alg:fctrigger}.
The inputs for CTrigger algorithm include:
\begin{enumerate}
    \item system state ($x$), 
    \item the safe control from ISSA ($u$), 
    \item the angle between the robot's heading vector and the vector from the robot to the obstacle 
($\alpha$), 
\item the minimum relative acceleration 
($a_{min}$), 
\item the maximum relative acceleration
($a_{max}$), 
\item the triggering angular velocity ($w_{trigger}$), where $w_{trigger} = \inf_{v \geq \frac{v_{max}}{2}}\sup_{u} |w|$, and 
\item the triggering angle ($\Delta_{\min}$), where 
$\Delta_{\min} = \inf_{|\cos(\alpha)| \leq \frac{\sqrt{3}}{2}, |w| \geq \frac{|w_{trigger}|}{2}} |\Delta \cos(\alpha)|$, and $\Delta \cos(\alpha)$ denotes the change magnitude of $\cos(\alpha)$ between two consecutive time steps.
\end{enumerate} Note that $a_{min}, a_{max}, w_{trigger}$ and $\Delta_{\min}$ are fundamental system properties, which can be evaluated offline.

\begin{algorithm}[h]
\newtheory{
\caption{Convergence Trigger}
\label{alg:fctrigger}
\begin{algorithmic}[1]
\Procedure{CTrigger}{$x, u, \alpha, a_{min}, a_{max},  w_{trigger}, \Delta_{\min}$} 
\If{$|\cos(\alpha)| < \min\{\frac{\sqrt{3}}{2}, \frac{\Delta_{\min}}{2}\}$ and $\phi(x) > 0$}
    \If{$v < \frac{v_{max}}{2}$}
        \State Use uniform sampling to find a safe control $u^{Trig}$, s.t. (i) $a \geq \frac{\min\{-a_{min}, a_{max}\}}{2}$ when $\cos(\alpha) < 0$, and (ii) $a \leq -\frac{\min\{-a_{min}, a_{max}\}}{2}$ when $\cos(\alpha) \geq 0$.
    \Else
        \State Use uniform sampling to find a safe control $u^{Trig}$, s.t. $|w| \geq \frac{|w_{trigger}|}{2}$ 
    \EndIf
    \State Return $u^{Trig}$
\Else
    \State Return $u$
\EndIf
\EndProcedure
\end{algorithmic}
}
\end{algorithm}

The core idea of CTrigger algorithm is to enable \textit{finite time convergence} via preventing $|\cos(\alpha)|$ from approaching zero (according to \eqref{eq: rule 2}). Intuitively, this is preventing the robot from moving in a circle around the obstacle at a constant distance, and thus preventing non-convergence. Specifically, CTrigger filters the safe control solution from ISSA when $|\cos(\alpha)|$ is less than a threshold, i.e. $\min\{\frac{\sqrt{3}}{2}, \frac{\Delta_{\min}}{2}\}$, which is accomplished via two main steps. I. Generate safe control with non-trivial acceleration to enable the system to gain enough speed (Line 4 of \Cref{alg:fctrigger}); II. Generate safe control with non-trivial angular velocity to push $|\cos(\alpha)|$ away from zero (Line 6 of \Cref{alg:fctrigger}). And we show in \Cref{thoem:discrete} that ISSA with CTrigger algorithm ensures \textit{forward invariance} and \textit{finite time convergence} to the set $\mathcal{S}=\mathcal{X}_S\cap \mathcal{X}_S^D$ for discrete-time systems.

 \begin{remark}
      Note that there are three heuristic based hyper-parameters in \Cref{alg:fctrigger}, including $\frac{\Delta_{min}}{2}$, $\frac{\min\{-a_{min}, a_{max}\}}{2}$, and $\frac{|w_{trigger}|}{2}$, and the selection heuristics are summarized as following: 1) $\frac{\Delta_{min}}{2}$ ensures that the triggering control will only be activated when $|\cos(\alpha)|$ is sufficiently small
 ; 2) $\frac{\min\{-a_{min}, a_{max}\}}{2}$ ensures that the candidate safe triggering control set for acceleration (Line 4 in \Cref{alg:fctrigger}) is non-trivial; 3) $\frac{|w_{trigger}|}{2}$ ensures that the candidate safe triggering control set for rotation (Line 6 in \Cref{alg:fctrigger}) is non-trivial. Although $a_{min}, a_{max}, w_{trigger}$ and $\Delta_{\min}$ are the fundamental system properties, it is possible to change the denominators of the hyper-parameters, e.g. $\frac{|w_{trigger}|}{3}$, or $\frac{\min\{-a_{min}, a_{max}\}}{5}$. However, there are trade-offs when we change the denominators of the hyper-parameters. Specifically, system will get out of \textit{Singularity State} faster if the denominator decreases, whereas it may take longer sampling time for finding a valid triggering control. In practice, we find setting the denominator as $2$ works well for balancing the aforementioned trade-offs. 
 \end{remark}

\section{Theoretical Results for ISSA When the Sampling Time is Non-Negligible}
\label{sec:theory_discrete}
\begin{theorem}[Forward Invariance and Finite Time Convergence for Discrete-Time System]
\label{thoem:discrete}
If the control system satisfies the assumptions in \Cref{asm:ruleassumption} and \Cref{asm:discrete assumption}, and the safety index design follows the rule described in \Cref{sec:synthesis_discrete}, the discrete-time implicit safe set algorithm described in \Cref{sec:ISSA_discrete} guarantees the \textit{forward invariance} and \textit{finite time convergence}
to the set $\mathcal{S}\subseteq\mathcal{X}_S$. 
\end{theorem} 

To prove the main theorem, we introduce two important propositions to show that 1) the set of safe control for discrete-time systems $\mathcal{U}_S^D(x)$ is always nonempty if we choose a safety index that satisfies the design rule in \cref{sec:synthesis_discrete}; and 2) the proposed ISSA for Discrete-Time System is guaranteed to find a safe control in finite time, i.e. either $u^{Trig}$ or $u$. With these two propositions, it is then straightforward to prove the \textit{forward invariance} and \textit{finite time convergence} to the set $\mathcal{S}\subseteq\mathcal{X}_S$. In the following discussion, we discuss the two propositions in  \Cref{sec:proof_prop3} and \Cref{sec: prop issa discrete}, respectively.
Then, we prove \Cref{thoem:discrete} in \Cref{sec:proof_discrete_main}.

\subsection{Feasibility of Safety Index for Discrete-Time System}
\label{sec:proof_prop3}

\begin{proposition}[Nonempty set of safe control for Discrete-Time System]\label{prop:discrete non-empty}
If the dynamic system satisfies the assumptions in and \Cref{asm:ruleassumption} and \Cref{asm:discrete assumption},
then the discrete-time safety index design rule in \Cref{sec:synthesis_discrete} ensures that the robot system in 2D plane has nonempty set of safe control at any state, i.e., $\mathcal{U}_S^D(x) \neq \emptyset, \forall x$.
\end{proposition}

Note that the set of safe control $\mathcal{U}_S^D(x):=\{u\in \mathcal{U}\mid \phi(f(x, u)) \leq \max\{\phi(x)-\eta, 0\} \}$ is non-empty if the following condition holds:
\begin{align}
\label{eq: sufficient condition for discrete}
    \forall x, \exists u, \text{s.t. } \phi(f(x, u)) \leq \phi(x)-\eta~,
\end{align}
where 
\begin{align}
    \label{eq: expand phi x u}
    \phi(f(x,u)) = \phi(x) + dt\cdot \dot\phi(x,u) + dt^2\cdot \frac{\ddot\phi(x,u)}{2} + \Delta,
\end{align}
and $\Delta$ is the non-negligible discrete-time approximation error. 

In the following discussion, we first analysis the upper bound of $\Delta$ if there's only one obstacle (\Cref{lem:bounded approximation error}). Then we leverage \Cref{lem:bounded approximation error}
show that the safety index design rule in \Cref{sec:synthesis_discrete} guarantees \eqref{eq: sufficient condition for discrete} hold if there's only one obstacle (\Cref{lem: exist safe control discrete time}). Finally, we leverage  \Cref{lem: exist safe control discrete time} to show $\mathcal{U}_S^D(x)$ is non-empty if there're multiple obstacles.

\subsubsection{Preliminary Results}
\label{sec: prelim prop3}

\begin{lemma}
If the dynamic system satisfies the assumptions in \Cref{asm:ruleassumption} and \Cref{asm:discrete assumption}, and there's only one obstacle in the environment, then the safety index design rule for discrete-time systems in \Cref{sec:synthesis_discrete} ensures the approximation error $\Delta$ for discrete-time systems is upper bounded by $\Delta_{max}dt^2 (e^{|w|dt}-1)$, where $\Delta_{max}$ is a positive constant, $dt$ is the discrete time step, and $w$ is the applied angular velocity.
\label{lem:bounded approximation error}
\end{lemma}
\begin{proof}

First, we will discuss the derivatives of $d$, which is the relative distance between the robot and the obstacle. According to the zero-order hold assumption for acceleration and angular velocity from \Cref{asm:discrete assumption}, we have $\dot a = 0$, and $\dot w = 0$. And we already have $\dot d = -v\cos(\alpha)$, then following condition holds:
\begin{align}
\label{eq: d i}
    d^{(i)} = (i-1)a(\pm\cos(\alpha) | \pm\sin(\alpha)) w^{i-2} + v(\pm\sin(\alpha) | \pm \cos(\alpha)) w^{i-1}
\end{align}
where $\pm\cos(\alpha) | \pm\sin(\alpha)$ denotes one of the terms from set \\
$\{\cos(\alpha), -\cos(\alpha), \sin(\alpha), -\sin(\alpha)\}$. $d^{(i)}$ denotes the $i$-th order derivatives of $d$.

According to the discrete-time system safety index design rule in \Cref{sec:synthesis_discrete}, we have $n = 1$, thus the derivatives of safety index satisfy the following conditions: 
\begin{align}
\label{eq: phi i}
    \phi^{(i)} = -d^{(i)} - kd^{(i+1)}
\end{align}

Without loss of generality, suppose $d^{(i)} = -(i-1)a\cos(\alpha) w^{i-2} + v\sin(\alpha) w^{i-1}$, and $d^{(i+1)} = ia\sin(\alpha)w^{i-1} + v\cos(\alpha) w^i$, we have: 
\begin{align}
    \phi^{(i)} &= -d^{(i)} - kd^{(i+1)} \\ \nonumber
    &= -(-(i-1)a\cos(\alpha) w^{i-2} + v\sin(\alpha) w^{i-1}) - k(ia\sin(\alpha)w^{i-1} + v\cos(\alpha) w^i) \\ \nonumber
    &=w^{i-2}\bigg((i-1)a\cos(\alpha) - (v + kia)\sin(\alpha)w - kv\cos(\alpha) w^2\bigg)
\end{align}

Next, we use the Taylor expansion to represent the safety index at next time step as following: 
\begin{align}
\label{eq:taylor_phi}
    \phi_{t+1} = \phi_t + \frac{\dot \phi_t}{1!} dt + \frac{\ddot \phi_t}{2!}dt^2 + \frac{\dddot \phi_t}{3!}dt^3 + \cdots
\end{align}
where each term from \eqref{eq:taylor_phi} except $\phi_t$ can be represented in the form of $\frac{\phi_t^{(i)}}{i!}dt^i, i = 1, 2, \cdots$. When $i \geq 2$, we have an important property for $\frac{\phi_t^{(i)}}{i!}dt^i$ as following: 
\begin{align}
    \frac{\phi_t^{(i)}}{i!}dt^i = dt^2 \frac{(i-1)a\cos(\alpha) - (v + kia)\sin(\alpha)w - kv\cos(\alpha) w^2}{i!} (wdt)^{i-2}
\end{align}
Next, we can find the upper bound of one part of the RHS as following: 
\begin{align}
\label{eq:upperbound_mid_term}
    &\frac{(i-1)a\cos(\alpha) - (v + kia)\sin(\alpha)w - kv\cos(\alpha) w^2}{i!}\\ \nonumber
    & \leq \bigg|\frac{a\cos(\alpha)}{(i-1)!}\bigg| + \bigg| \frac{(v + k|a|)\sin(\alpha)w}{(i-1)!}\bigg| + \bigg|\frac{kv\cos(\alpha) w^2}{(i-1)!} \bigg| \\ \nonumber
    & \leq \frac{a_{m} + v_{max}w_m + ka_mw_m + kv_{max}w_m^2}{(i-1)!} \\ \nonumber
    & = \frac{(a_{m} + v_{max}w_m)(1 + kw_{m})}{(i-1)!}
\end{align}

where $a_m = \max\{-a_{min}, a_{max}\}$ and $w_m = \max\{-w_{min}, w_{max}\}$. Let's define $\Delta_{max} = (a_{m} + v_{max}w_m)(1 + kw_{m})$, and $\Delta_{max}$ is a constant given a specific $k$ value. Based on \eqref{eq:upperbound_mid_term} and \eqref{eq:taylor_phi}, the upper bound of $\frac{\phi_t^{(i)}}{i!}dt^i$ is defined as following:

\begin{align}
\label{eq: upper_bound_phi_i_dt}
    &\frac{\phi_t^{(i)}}{i!}dt^i \\ \nonumber
    &\leq dt^2 \frac{\Delta_{max}}{(i-1)!} (wdt)^{i-2}\\ \nonumber
    & \leq  dt^2 \frac{\Delta_{max}}{(i-1)!} (|w|dt)^{i-2}\\ \nonumber
    & \leq \Delta_{max}dt^2\frac{(|w|dt)^{i-2}}{(i-2)!}
\end{align}

Recall that $\phi(x_{t+1}) = \phi(x_t) + dt\cdot \dot\phi(x_t) + dt^2\cdot \frac{\ddot\phi(x_t)}{2} + \Delta$, where $\Delta$ is the approximation error due to discrete-time systems, and $ \Delta = \frac{\dddot \phi_t}{3!}dt^3 + \frac{\ddddot \phi_t}{4!}dt^4 + \cdots$. According to the Taylor series of exponential function $e^x = 1 + x + \frac{x^2}{2!} + \frac{x^3}{3!} + \cdots$, and leverage the result from \eqref{eq: upper_bound_phi_i_dt}, the upper bound of $\Delta$ can be derived as following:
\begin{align}
    \Delta &= \frac{\dddot \phi_t}{3!}dt^3 + \frac{\ddddot \phi_t}{4!}dt^4 + \cdots \\ \nonumber
    &\leq \sum_{i=1}^\infty\Delta_{max}dt^2\frac{(|w|dt)^{i}}{i!} \\ \nonumber
    &= \Delta_{max}dt^2 (e^{|w|dt}-1)
\end{align}

\end{proof}

\begin{lemma}
If the dynamic system satisfies the assumptions in \Cref{asm:ruleassumption} and \Cref{asm:discrete assumption}, and there's only one obstacle in the environment,
then the safety index design rule for discrete-time systems in \Cref{sec:synthesis_discrete} ensures that the robot system in 2D plane has nonempty set of safe control at any state.
\label{lem: exist safe control discrete time}
\end{lemma}
\begin{proof}

According to \eqref{eq: safety index}, $\dot\phi=-nd^{n-1}\dot d -k \ddot d$. We ignored the subscript $i$ since it is assumed that there is only one obstacle. Therefore, according to \eqref{eq: expand phi x u}, condition \eqref{eq: sufficient condition for discrete} is equivalent to the following condition

\begin{align}
    \label{eq:origin_condtion_discrete}
    \forall x,  \exists u, \text{s.t. } \ddot d \geq \frac{\frac{\eta + dt^2\cdot \frac{\ddot\phi(x,u)}{2} + \Delta}{dt} - n d^{n-1}\dot d}{k}~.
\end{align}

To prove the non-empty set of safe control for all system state, we will prove that there always exists a safe control such that \eqref{eq:origin_condtion_discrete} holds. According to the safety index design rule for discrete-time systems in \Cref{sec:synthesis_discrete}, we have the parameter $n$ is chosen to be $1$. Then combined with \eqref{eq: d i} and \eqref{eq: phi i}, the condition for \eqref{eq:origin_condtion_discrete} can be rewrote as: 
\begin{align}
\label{eq: real fundamental_detail}
     \forall (\alpha, v), \exists (a,w), &\text{ s.t. }\\ \nonumber
     -a \cos(\alpha) + v \sin(\alpha) w &\geq \frac{\frac{\eta}{dt} + dt\cdot \frac{a\cos(\alpha) - (v+2ka)\sin(\alpha)w - kv\cos(\alpha)w^2}{2} + \frac{\Delta}{dt} +  v\cos(\alpha)}{k}
\end{align}

According to \Cref{lem:bounded approximation error}, we have $ \Delta \leq \Delta_{max}dt^2 (e^{|w|dt}-1)$ and \eqref{eq: real fundamental_detail} can be verified by showing: 
\begin{align}
\label{eq: real fundamental_detail_upper_bound}
     \forall (\alpha, v), \exists (a,w), &\text{ s.t. }\\ \nonumber
     -a \cos(\alpha) + v \sin(\alpha) w & \geq \frac{ \frac{\eta}{dt} + dt\cdot \frac{a\cos(\alpha) - (v+2ka)\sin(\alpha)w - kv\cos(\alpha)w^2}{2}}{k} + \\ \nonumber 
     & \frac{\Delta_{max}dt (e^{|w|dt}-1) +  v\cos(\alpha)}{k}
\end{align}

By selecting $w \to 0$, the condition for \eqref{eq: real fundamental_detail_upper_bound} becomes: 
\begin{align}
\label{eq: real fundamental_detail_simplified}
    \forall (\alpha, v), \exists (a), \text{ s.t. } -a \cos(\alpha) & \geq \frac{\frac{\eta}{dt} + dt\cdot \frac{a\cos(\alpha)}{2} +  v\cos(\alpha)}{k} 
\end{align}
 
There are only two scenarios $\cos(\alpha) < 0$ and $\cos(\alpha) \geq 0$. 

\textbf{Case 1:} $\cos(\alpha) < 0$. According to safety index design rule described in \Cref{sec:synthesis_discrete}, we have $\eta = \eta_0 |\cos(\alpha)|$ and $a_{max} \geq \min\{-a_{min}, a_{max}\} \geq  \frac{\frac{\eta_0}{dt} + v_{max}}{k}\geq \frac{\eta_0}{kdt}$. Thus, the following condition holds: 
\begin{align}
\label{eq:lem2_result1}
     \forall (\alpha, v), \exists (a > 0), \text{ s.t. } -a \cos(\alpha) &= a |\cos(\alpha)| \\ \nonumber
     &\geq \frac{\eta_0 |\cos(\alpha)|}{k dt}\\ \nonumber
     &\geq \frac{\frac{\eta_0 |\cos(\alpha)|}{dt} + dt\cdot \frac{a\cos(\alpha)}{2} +  v\cos(\alpha)}{k} 
\end{align}
which indicates \eqref{eq: real fundamental_detail_simplified} holds when $\cos(\alpha) < 0$.

\textbf{Case 2:} $\cos(\alpha) \geq 0$. 
According to the safety index design rule, we have $\frac{\frac{\eta_0}{dt} + v_{\max}}{k} \leq \min\{-a_{min}, a_{max}\}$. Thus, by selecting $a < 0$, the following inequality holds: 
\begin{align}
\label{eq:before_case2}
    \max_{a} -a = - a_{min} \geq \frac{\frac{\eta_0}{dt} + v_{\max}}{k} \geq \frac{\frac{\eta_0}{dt} + dt\cdot \frac{a}{2} +  v_{\max}}{k} \geq \frac{\frac{\eta_0}{dt} + dt\cdot \frac{a}{2} +  v}{k} 
\end{align}

Note that $\cos(\alpha) = |\cos(\alpha)|$, when $\cos(\alpha) \geq 0$. Therefore, \eqref{eq:before_case2} indicates the following condition holds:
\begin{align}
\label{eq:lem2_result2}
     \forall (\alpha, v), \exists (a < 0), \text{ s.t. } -a \cos(\alpha) &\geq \frac{\frac{\eta_0}{dt} + dt\cdot \frac{a}{2} +  v}{k} \cos(\alpha) \\ \nonumber
     &= \frac{\frac{\eta_0 |\cos(\alpha)|}{dt} + dt\cdot \frac{a\cos(\alpha)}{2} +  v\cos(\alpha)}{k} 
\end{align}

which further indicates \eqref{eq: real fundamental_detail_simplified} holds when $\cos(\alpha) \geq 0$. 

Combine these two cases, we have proved that if the control system satisfies the assumptions in \Cref{asm:ruleassumption} and \Cref{asm:discrete assumption},
then the safety index design rule in \Cref{sec:synthesis_discrete} ensures that the robot system in 2D plane has nonempty set of safe control at any state.

\end{proof}

\subsubsection{Proof of \Cref{prop:discrete non-empty}}\label{subsec: si rule}

\begin{proof}

Note that the set of safe control $\mathcal{U}_S^D(x_t):=\{u\in \mathcal{U}\mid \phi(f(x, u)) \leq \max\{\phi(x)-\eta, 0\} \}$ is non-empty if it is non-empty in the following two cases: $\phi(x) \geq -\Delta_{\phi_{max}}$ and $ \phi(x) < -\Delta_{\phi_{max}}$.

\textbf{Case 1:} Firstly, we consider the case where $\phi(x) \geq -\Delta_{\phi_{max}}$.
We have $\phi(x) = \max_i \phi_i(x) \geq -\Delta_{\phi_{max}}$, where $\phi_i$ is the safety index with respect to the $i$-th obstacle. According to \cref{asm:discrete assumption}, we have that at any given time, there can at most be one obstacle becoming safety critical, such that $\phi \geq - \Delta_{\phi_{max}}$ (Sparse Obstacle Environment). Therefore, $\max_i \phi_i(x) > 0$ indicates there's only one obstacle ($j$-th obstacle) in the environment that $\phi_j(x) \geq -\Delta_{\phi_{max}}$. Whereas for the rest of the obstacles, we have $\phi_k(x) < - \Delta_{\phi_{max}}, k \neq j$.

 Denote ${\mathcal{U}_S^D}_j(x):=\{u\in \mathcal{U}\mid \phi_j(f(x, u)) \leq \max\{\phi_j(x)-\eta, 0\} \}$
. According to \Cref{lem: exist safe control discrete time}, we have if the dynamic system satisfies the assumptions in \Cref{asm:discrete assumption},
then the safety index design rule in \Cref{sec:synthesis_discrete} ensures ${\mathcal{U}_S^D}_j(x)$ is nonempty. Since $\Delta_{\phi_{max}}$ is the maximum possible $\phi$ change magnitude at one time step, we also have that $\phi_k(f(x, u)) \leq 0 \leq \max\{\phi_j(x)-\eta, 0\}$, for $u \in {\mathcal{U}_S^D}_j(x)$.
 
Therefore, we further have that if $\phi(x) > 0$, by applying $u \in {\mathcal{U}_S^D}_j(x)$, the following condition holds: 
\begin{align}
    \phi(f(x, u)) = \max_i \phi_i(f(x, u)) \leq \max\{\phi_j(x)-\eta, 0\} = \max\{\phi(x)-\eta, 0\}
\end{align}

\textbf{Case 2:} Secondly, we consider the case where $\phi(x) < -\Delta_{\phi_{max}}$. We have  $\phi(x) = \max_i \phi_i(x) < -\Delta_{\phi_{max}}$, where $\phi_i$ is the safety index with respect to the $i$-th obstacle. Therefore, we have $\forall i, \phi_i(x) < -\Delta_{\phi_{max}}$.
Since $\Delta_{\phi_{max}}$ is the maximum possible $\phi$ change magnitude at one time step, we have that by applying $u \in \mathcal{U}$, the following condition holds: 
 
 \begin{align}
    \phi(f(x, u)) = \max_i \phi_i(f(x, u)) \leq 0 = \max\{\phi(x)-\eta, 0\}
\end{align}

In summary,
 if the dynamic system satisfies the assumptions in \Cref{asm:ruleassumption} and \Cref{asm:discrete assumption},
then the discrete-time safety index design rule in \Cref{sec:synthesis_discrete} ensures that the robot system in 2D plane has nonempty set of safe control at any state, i.e., $\mathcal{U}_S^D(x) \neq \emptyset, \forall x$.

\end{proof}

\subsection{Feasibility of Implicit Safe Set Algorithm for Discrete-Time System}
\label{sec: prop issa discrete}

Here we define set $\mathcal{X}_S^D:= \{x|\phi(x) \leq 0\}$.

\begin{proposition}[Finite Time Convergence to $\mathcal{X}_S^D$ with One Obstacle]
If the control system satisfies the assumptions in \Cref{asm:ruleassumption} and \Cref{asm:discrete assumption}, and there's only one obstacle in the environment,
the safety index design follows the rule described in \Cref{sec:synthesis_discrete},
the implicit safe set algorithm with convergence trigger algorithm together guarantee the \textit{finite time convergence} to the set $\mathcal{X}_S^D$. 
\label{lem:ftc_one_obs}
\end{proposition}

In the following discussion, we first proof that one important branch in \Cref{alg:fctrigger} (Line 6) can always find solution $u^{Trig}$. And then we can prove the finite time convergence described in \Cref{lem:ftc_one_obs}.

\subsubsection{Preliminary Results}
\label{sec: prelim prop4}

\begin{lemma}[Persistent Existence of Safe Angular Velocity]
If the dynamic system satisfies the assumptions in \Cref{asm:ruleassumption} and \Cref{asm:discrete assumption}, and there's only one obstacle in the environment,
then the safety index design rule in \Cref{sec:synthesis_discrete} ensures that the robot system in 2D plane can always generate safe control such that angular velocity is non-zero for states where $|\sin(\alpha)| \geq \frac{1}{2}, v \geq \frac{v_{max}}{2}$. 
\label{lem:exsitence_angular}
\end{lemma}
\begin{proof}

According to \eqref{eq: real fundamental_detail_upper_bound}, the non-empty set of safe control for discrete-time systems can be verified by showing the following condition: 
\begin{align*}
     \forall (\alpha, v), \exists (a,w), &\text{ s.t. }\\ \nonumber
     -a \cos(\alpha) + v \sin(\alpha) w & \geq \frac{ \frac{\eta}{dt} + dt\cdot \frac{a\cos(\alpha) - (v+2ka)\sin(\alpha)w - kv\cos(\alpha)w^2}{2}}{k} + \\ \nonumber 
     & \frac{\Delta_{max}dt (e^{|w|dt}-1) +  v\cos(\alpha)}{k}
\end{align*}

The proof of \Cref{lem: exist safe control discrete time} has shown that by setting $w \to 0$, the condition \eqref{eq: real fundamental_detail_simplified} holds: 
\begin{align*}
    \forall (\alpha, v), \exists (a), \text{ s.t. } -a \cos(\alpha) & \geq \frac{\frac{\eta}{dt} + dt\cdot \frac{a\cos(\alpha)}{2} +  v\cos(\alpha)}{k}
\end{align*}

Thus, by subtracting the components of \eqref{eq: real fundamental_detail_simplified} from \eqref{eq: real fundamental_detail_upper_bound}, we can verify there exists non-empty set of safe control where angular velocity is non-zero at states where $|\sin(\alpha)| \geq \frac{1}{2}, v \geq \frac{v_{max}}{2}$, by showing the following condition holds: 

\begin{align}
\label{eq: non zero angular velocity condition}
    \forall (|\sin(\alpha)| \geq \frac{1}{2}, &v \geq \frac{v_{max}}{2}, a), \exists (|w| > 0), \text{ s.t. },  \\ \nonumber
     &v\sin(\alpha) w \geq \frac{dt\cdot \frac{- (v+2ka)\sin(\alpha)w - kv\cos(\alpha)w^2}{2} + \Delta_{max}dt (e^{|w|dt}-1)}{k} \\ \nonumber
\end{align}

Note that $w$ can either be positive or negative, by selecting $w$ such that $\sin(\alpha)w > 0$, showing \eqref{eq: non zero angular velocity condition} is equivalent to show the following condition holds: 
\begin{align}
\label{eq: absolute non zero angular velocity condition}
    \forall (|\sin(\alpha)| \geq \frac{1}{2}, &v \geq \frac{v_{max}}{2}, a), \exists (|w| > 0), \text{ s.t. },  \\ \nonumber
    & (\frac{kv}{dt} + \frac{v + 2ka}{2}) |\sin(\alpha)| |w| + \frac{kv\cos(\alpha)|w|^2}{2} - \Delta_{max} (e^{|w|dt}-1)\geq 0 \\ \nonumber
\end{align}

Denote $F(|w|) = (\frac{kv}{dt} + \frac{v + 2ka}{2}) |\sin(\alpha)| |w| + \frac{kv\cos(\alpha)|w|^2}{2} - \Delta_{max} (e^{|w|dt}-1)$, we have $\nabla_{|w|}F(|w|) = (\frac{kv}{dt} + \frac{v + 2ka}{2}) |\sin(\alpha)| + kv\cos(\alpha)|w| - dt\Delta_{max}e^{|w|dt}$. Since $\Delta_{max} = (a_{m} + v_{max}w_m)(1 + kw_{m})$, We have the following condition hold:
\begin{align}
\label{eq: grad_F before}
    \nabla_{|w|}F(0) &= (\frac{kv}{dt} + \frac{v + 2ka}{2}) |\sin(\alpha)| - dt\Delta_{max} \\ \nonumber
    &> k(\frac{\frac{v}{dt} + a}{2} - dt (a_{m} + v_{max}w_m)(\frac{1}{k} + w_{m}))
\end{align}

According to the safety index design rule in \Cref{sec:synthesis_discrete}, we have $ \frac{v_{\max}}{k} \leq \frac{\frac{\eta_0}{dt} + v_{\max}}{k} \leq \min\{-a_{min}, a_{max}\} \leq -a_{min}$, which indicates $0 < \frac{1}{k} \leq \frac{-a_{min}}{v_{max}}$. We also have $\frac{a_{min}}{2} + \frac{v_{max}}{4dt} > (a_m + v_{max}w_m)(-\frac{a_{min}}{v_{max}} + w_{m})dt$ according to \Cref{asm:discrete assumption}. Therefore, when $v > \frac{v_{max}}{2}$, the lower bound for \eqref{eq: grad_F before} is summarized as following: 

\begin{align}
    \nabla_{|w|}F(0) &> k(\frac{\frac{v}{dt} + a}{2} - dt (a_{m} + v_{max}w_m)(\frac{1}{k} + w_{m})) \\ \nonumber
    &> k(\frac{\frac{v}{dt} + a}{2} - dt (a_{m} + v_{max}w_m)(\frac{-a_{min}}{v_{max}} + w_{m})) \\ \nonumber
    &> k(\frac{\frac{v_{max}}{2dt} + a_{min}}{2} - dt (a_{m} + v_{max}w_m)(\frac{-a_{min}}{v_{max}} + w_{m})) \\ \nonumber
    &> 0
\end{align}

Since $F(0) = 0$, $\nabla_{|w|}F(0) > 0 $, and $\nabla_{|w|}F(|w|)$ is differentiable  everywhere, there exists $|w_{trigger}| > 0$, such that $\forall |w| \in [0, |w_{trigger}|]$, \eqref{eq: absolute non zero angular velocity condition} holds.

\end{proof}

\subsubsection{Proof of \Cref{lem:ftc_one_obs}}
\label{sec: proof prop:4}
\begin{proof}

According to the safety index design rule, we have that $\eta = \eta_0|cos(\alpha)|$. Next, we will discuss the only two situations for $|\cos(\alpha)|$: 

\textbf{Case 1:} $|\cos(\alpha)| \geq \min\{\frac{\sqrt{3}}{2}, \frac{\Delta_{\min}}{2}\}$. In this case, \Cref{alg:fctrigger} won't be triggered, thus the properties of finite iterations convergence follows \Cref{alg:adamsc}.

\textbf{Case 2.1:} $|\cos(\alpha)| < \min\{\frac{\sqrt{3}}{2}, \frac{\Delta_{\min}}{2}\}$ and $v<\frac{v_{max}}{2}$. In this case, CTrigger algorithm is activated to filter the control solution from ISSA, which seeks to find a safe control $u^{Trig}$ such that $a_t \geq \frac{\min\{-a_{min}, a_{max}\}}{2}$ when $\cos(\alpha) < 0$ and $a_t \leq -\frac{\min\{-a_{min}, a_{max}\}}{2}$ when $\cos(\alpha) \geq 0$ and $w \rightarrow 0$. The existence of $u^{Trig}$ is guaranteed by the \eqref{eq:lem2_result1}, \eqref{eq:lem2_result2} from the proof of \Cref{lem: exist safe control discrete time}, and can be found through uniform sampling within finite iterations by \Cref{lem:finiteness}.

\textbf{Case 2.2:} $|\cos(\alpha)| < \min\{\frac{\sqrt{3}}{2}, \frac{\Delta_{\min}}{2}\}$ and $v\geq\frac{v_{max}}{2}$. According to \Cref{asm:discrete assumption}, we have that $|a_t|$ is constant between two consecutive time steps, which indicates the following property holds:

\begin{enumerate}[label=\textnormal{(P\arabic*)}]
    \item When $|\cos(\alpha)| < \min\{\frac{\sqrt{3}}{2}, \frac{\Delta_{\min}}{2}\}$, CTrigger algorithm ensures $v \geq \frac{v_{max}}{2}$ within at most $\frac{v_{max}}{\min\{-a_{min}, a_{max}\}}$  iterations \label{property 1}
\end{enumerate}

In this case, we have CTrigger algorithm seeks to find a safe control $u^{Trig}$ such that $|w_t| \geq \frac{|w_{trigger}|}{2}$. Note that $|\cos(\alpha)| < \frac{\sqrt{3}}{2}$ indicates $|\sin(\alpha)| > \frac{1}{2}$, thus the existence of $u^{Trig}$ is guaranteed by the \cref{lem:exsitence_angular}, and can be found through uniform sampling within finite iterations by \Cref{lem:finiteness}.

Suppose at time step $t$, $|\cos(\alpha_t)| < \min\{\frac{\sqrt{3}}{2}, \frac{\Delta_{\min}}{2}\}$ and $v_t \geq \frac{v_{max}}{2}$. Denote $\Delta_t \cos(\alpha)$ as the change of $\cos(\alpha)$ at time step $t$ after applying $u^{Trig}$, we have $|\Delta_t \cos(\alpha)| \geq  \Delta_{\min} > \frac{\Delta_{\min}}{2} > |\cos(\alpha_t)|$, which indicates:
\begin{align}
    |\cos(\alpha_{t+1})| &= |\cos(\alpha_t) + \Delta_t \cos(\alpha)|\\ \nonumber
    &\geq \Big| |\Delta_t \cos(\alpha)| - |\cos(\alpha_t)| \Big|\\ \nonumber
    &\geq \Big|\Delta_{\min} - |\cos(\alpha_t)| \Big|\\ \nonumber
    &> \min\{\frac{\sqrt{3}}{2}, \frac{\Delta_{\min}}{2}\}
\end{align}
which further shows that $\eta_{t+1} > \eta_0 \min\{\frac{\sqrt{3}}{2}, \frac{\Delta_{\min}}{2}\}$.

Therefore, according to property \cref{property 1}, ISSA with CTrigger ensures \\
$\eta \geq \eta_0 \min\{\frac{\sqrt{3}}{2}, \frac{\Delta_{\min}}{2}\}$ in at most every $\frac{v_{max}}{\min\{-a_{min}, a_{max}\}}+1$ time steps. Thus, if $\phi(x_{t_0}) > 0$, then $\phi(x_{t_1}) \leq 0$ for finite time $t_1 > t_0$, where $t_1 - t_0 < \frac{\phi(x_{t_0})}{\eta_0 \min\{\frac{\sqrt{3}}{2}, \frac{\Delta_{\min}}{2}\}}(\frac{v_{max}}{\min\{-a_{min}, a_{max}\}}+1)$.

\end{proof}

\subsection{Proof of \Cref{thoem:discrete}}
\label{sec:proof_discrete_main}

\subsubsection{Preliminary Results}

\begin{lemma}[Forward Invariance and Finite Time Convergence of $\mathcal{X}_S^D$]
If the control system satisfies the assumptions in \Cref{asm:ruleassumption} and \Cref{asm:discrete assumption}, and
the safety index design follows the rule described in \Cref{sec:synthesis_discrete},
the implicit safe set algorithm with convergence trigger algorithm together guarantee the \textit{forward invariance} and \textit{finite time convergence} to the set $\mathcal{X}_S^D$. 
\label{lem:fi_ftc_safety_index}
\end{lemma}
\begin{proof}

Firstly, we will prove the \textit{forward invariance} of set $\mathcal{X}_S^D$. If the control system satisfies the assumptions in \Cref{asm:ruleassumption} and \Cref{asm:discrete assumption}, and the safety index design follows the rule described \Cref{sec:synthesis_discrete}, then we can ensure the system has nonempty set of safe control at any state by \Cref{prop:discrete non-empty}. By \Cref{prop:2}, implicit safe set algorithm can always find local optima solution of \eqref{eq:adamba_discrete}. The local optima solution always satisfies the constraint $\phi(f(x_t, u_t)) \leq \max\{\phi(x_t)-\eta, 0\}$, which indicates that if $\phi(x_{t_0}) \leq 0$, then $\phi(x_t) \leq 0, \forall t \geq t_0$ since $\eta\geq0$. Note that $\phi(x) \leq 0$ demonstrates that $x \in \mathcal{X}_S^D$. Thus the forward invariance of $\mathcal{X}_S^D$ is proved.

Secondly, we will prove the \textit{finite time convergence} to $\mathcal{X}_S^D$. 
Suppose at time step $t_0$, we have $\phi(x_{t_0}) = \max_i \phi_i(x_{t_0}) > 0$. According to Sparse Obstacle Environment assumption from \Cref{asm:discrete assumption}, there's only one obstacle ($j$-th obstacle) in the environment that $\phi_j(x_{t_0}) > 0$. Whereas for the rest of the obstacles, we have $\phi_k(x_{t_0}) < - \Delta_{\phi_{max}}, k \neq j$.

According to \Cref{lem:ftc_one_obs}, we have implicit safe set algorithm together with convergence trigger algorithm guarantee that $\phi_j(x_{t_1}) \leq 0$ for a finite time $t_1 > t_0$ while $\phi_j(x_{t_1-1}) > 0$. By Sparse Obstacle Environment assumption, we also have $\forall k\neq j, \phi_k(x_{t_1 - 1}) < - \Delta_{\phi_{max}}$, hence $\forall k\neq j, \phi_k(x_{t_1}) \leq 0$. Therefore, $\phi(x_{t_1}) =  \max_i \phi_i(x_{t_1}) \leq 0$ and it demonstrates that $x_{t_1} \in \mathcal{X}_S^D$.

\end{proof}

\subsubsection{Proof the \Cref{thoem:discrete}}
\label{sec: proof theo2}
\begin{proof}

Leveraging \Cref{lem:fi_ftc_safety_index}, we then proceed to prove that the \textit{forward invariance} and \textit{finite time convergence} to the set $\mathcal{X}_S^D$ guarantees the \textit{forward invariance} and \textit{finite time convergence} to the set $S \subseteq \mathcal{X}_S$. Recall that $\mathcal{S}=\mathcal{X}_S\cap \mathcal{X}_S^D$. Depending on the relationship between $\mathcal{X}_S^D$ and $\mathcal{X}_S$, there are two cases in the proof which we will discuss below.

\textbf{Case 1: $\mathcal{X}_S^D = \{x|\phi(x) \leq 0\}$ is a subset of $\mathcal{X}_S = \{x|\phi_0(x) \leq 0\}$.}

In this case, $\mathcal{S} = \mathcal{X}_S^D$. 
According to \Cref{prop:1} and \Cref{lem:fi_ftc_safety_index}, if the safety index design follows the rule described in \Cref{sec:synthesis}, the implicit safe set algorithm guarantees the \textit{forward invariance} and \textit{finite time convergence} to the set $\mathcal{X}_S^D$ and hence $\mathcal{S}$. 

\textbf{Case 2: $\mathcal{X}_S^D = \{x|\phi(x) \leq 0\}$ is NOT a subset of $\mathcal{X}_S = \{x|\phi_0(x) \leq 0\}$.}

1) To prove \textit{finite time convergence} in this case, recall the forms of safety index are $\phi_i = \sigma + d^n_{min} - d_i^n - k \dot{d_i}$ and $\phi_{0_i} = d_{min} - d_i$. 
Since the finite time convergence to $\mathcal{X}_S^D$ is proved in \Cref{lem:fi_ftc_safety_index}, we then discuss the finite time convergence of $\mathcal{X}_S$ based on the fact that $\phi(x) \leq 0$.
Note that the set $\mathcal{X}_S^D$ can be divided into two subsets $ \mathcal{X}_S^D\setminus \mathcal{S} = \{x\mid\phi(x) \leq 0, \phi_0(x) > 0 \}$ and $\mathcal{S} = \{x\mid\phi(x) \leq 0, \phi_0(x) \leq 0 \}$. If $x_{t_0} \in \mathcal{X}_S^D\setminus \mathcal{S}$, meaning the following two conditions hold:
\begin{align}
    \max_i d_{min} - d_i > 0 \\
    \forall i, \sigma + d^n_{min} - d_i^n - k \dot{d_i} \leq 0
\end{align}
Thus, the following condition hold:
\begin{align}
\label{eq:convergence_theorem1}
    \forall i_{{\phi_0}_i>0}, \sigma - k \dot{d_i}&\leq d_i^n - d_{min}^n < 0 \\ \nonumber
    \dot d_i &> \frac{\sigma}{k}
\end{align}

\eqref{eq:convergence_theorem1} indicates that $\forall i, \phi_{0_i}(x_{t}) \leq 0$ for some finite time $t > t_0$, which shows $\phi_0(x_t) \leq 0$ for some finite time $t > t_0$. Therefore, we have proved \textit{finite time convergence} to the set $\mathcal{S} \subseteq\mathcal{X}_S$.

\begin{figure}[htbp]
    \centering
    \includegraphics[width=0.5\textwidth]{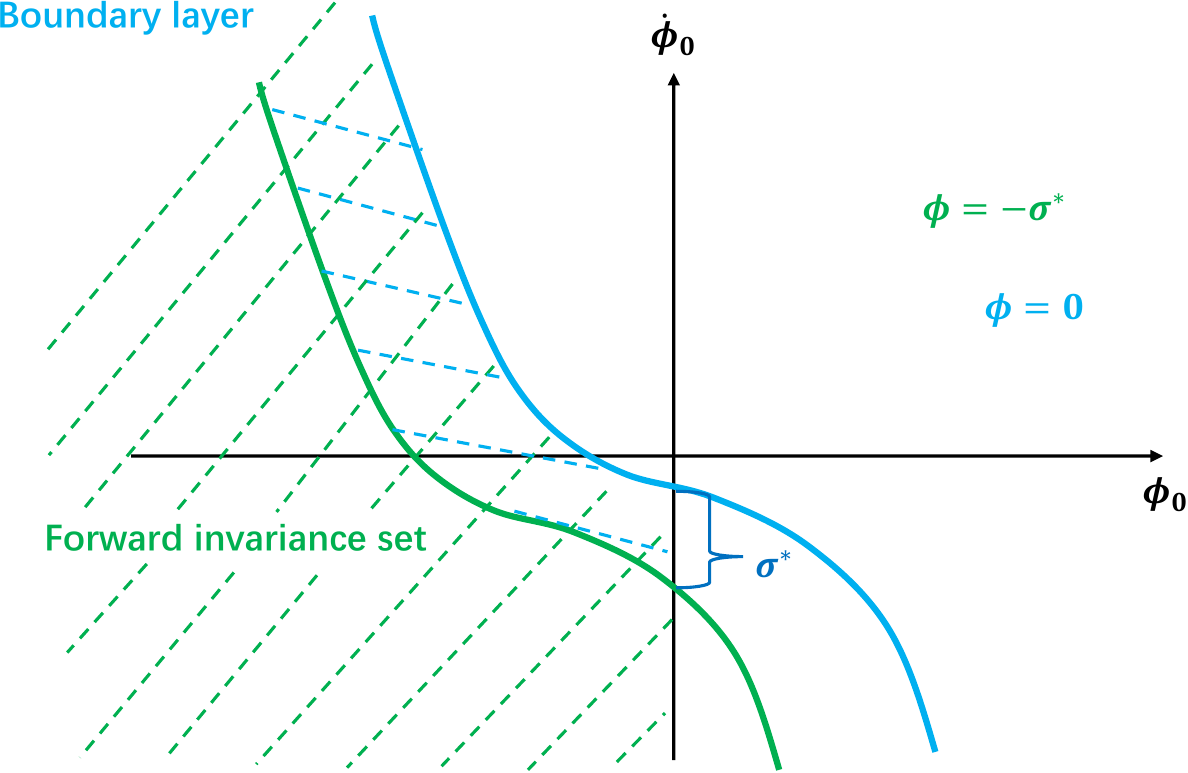}
    \caption{The illustration of \textit{forward invariance} under the constraint of $\phi(x_{t+1})\leq\max \{\phi(x_t)-\eta, 0\}) - \sigma^*$.}
    \label{fig:main_theorem}
\end{figure}

2) To prove the \textit{forward invariance} in this case, we first need to prove that if $x_t \in \mathcal{S}$, then $x_{t+1} \in \mathcal{X}_{\mathcal{S}}$. 

we first define $\sigma^*$ such that $(\sigma + d^n_{min})^{\frac{1}{n}} > d_{min} + \sigma^* $.
For $x_{t_0} \in \mathcal{S}$ and for each $d_{{t_0}_i}$, there are only two situations for $d_{{t_0}_i}$: 

2.1) $ d_{min} + \sigma^* \geq d_{{t_0}_i} \geq d_{min}$. And we have $x_t \in \mathcal{S}$, thus the following condition holds:
\begin{align}
    \phi(x_t) \leq 0 &\Rightarrow \sigma + d^n_{min} - d_{t_i}^n - k \dot{d_{t_i}} \leq 0 \\
    d_{t_i} \leq d_{min} + \sigma^* &\Rightarrow \nonumber
    -d_{t_i}^n \geq -(d_{min} + \sigma^*)^n =  -(\sigma + d^n_{min}) + \iota
\end{align}

where $\iota = \sigma + d^n_{min} - (d_{min} + \sigma^*)^n$ is a positive constant. Thus, the following condition holds: 
\begin{align}
\label{eq:increase_dotd}
     \dot{d_{t_i}} &\geq \frac{\sigma + d^n_{min} - d_{t_i}^n}{k} \\ \nonumber
     & \geq \frac{\sigma + d^n_{min} -(\sigma + d^n_{min}) + \iota}{k} \\ \nonumber
     & \geq \frac{\iota}{k} > 0
\end{align}

which indicates the $d_i$ will increase at next time step, then $ \phi_{0_i}(x_{t+1}) \leq \phi_{0_i}(x_{t}) \leq 0$. Thus in this case, if $x_t \in \mathcal{S}$, then $x_{t+1} \in \mathcal{X}_{\mathcal{S}}$.

2.2) $d_{{t_0}_i} > d_{min} + \sigma^*$. Note that $x_{t} \in \mathcal{S}$ and $d_{t_i} > d_{min} + \sigma^*$ indicate $\sigma + d^n_{min} - d_{t_i}^n - k \dot{d_{t_i}} \leq 0$ and $d_{min} - d_{t_i} < 0$, whereas $\dot d$ can either $< 0$ or $\geq 0$. 

If $\dot d_{t_i} \geq 0$, ${\phi_0}_i$ keeps decreasing, then $\phi_{0_i}(x_{t+1}) \leq \phi_{0_i}(x_{t}) \leq 0$.

If $\dot d_{t_i} < 0$, $\phi_{0_i}(x_{t+1}) \leq 0$
can be guaranteed if $d_{{t+1}_i} = d_{t_i} + \dot{d_i}^*dt \geq d_{min}$, where $\dot{d_i}^* \neq \dot{d_i}$ due to discrete-time systems. Denote $\dot{d}^*_{min}$ as the minimum $\dot{d}^*$ can be achieved in the system, the following condition holds:
\begin{align}
    d_{{t+1}_i} &= d_{t_i} + \dot{d_i}^*dt\\ \nonumber
            &\geq d_{t_i} + \dot{d}^*_{min}dt
\end{align}

According to \eqref{eq: rule 0} from the safety index design rule for discrete-time system in \Cref{sec:synthesis_discrete}, we have $\dot{d}^*_{min}dt > -\sigma^*$, then:
\begin{align}
    d_{{t+1}_i} &\geq d_{t_i} + \dot{d}^*_{min}dt \\\nonumber
    &> d_{t_i} - \sigma^* \\ \nonumber
    &> d_{min} + \sigma^* - \sigma^* \\ \nonumber
    &> d_{min}
\end{align}
which also indicates $\phi_{0_i}(x_{t+1}) \leq 0$. Thus in this case, if $x_t \in \mathcal{S}$, then $x_{t+1} \in \mathcal{X}_{\mathcal{S}}$.

Summarizing the above content, we have shown that if $x_t \in \mathcal{S}$, then $\forall i, {\phi_0}_i(x_{t+1}) \leq 0$, which indicates $\phi_0(x_{t+1}) = \max_i {\phi_0}_i(x_{t+1}) \leq 0$, hence $x_{t+1} \in \mathcal{X}_S$. We also have if $x_t \in \mathcal{S}$, $x_{t+1} \in \mathcal{X}_S^D$ by \Cref{lem:fi_ftc_safety_index}. Therefore, if $x_t \in \mathcal{S}$, $x_{t+1} \in \mathcal{S}$. By induction, we have shown if $x_{t_0} \in \mathcal{S}$, $x_t \in \mathcal{S}, \forall t > t_0$. Thus the forward invariance of $\mathcal{S}$ is proved.

Leveraging the results introduced above, in practice, we can ensure the \textit{forward invariance} of $\mathcal{S} = \{x\mid\phi(x) \leq 0, \phi_0(x) \leq 0 \}$ by using the safe control generation constraint  $\phi(x_{t+1})\leq\max \{\phi(x_t)-\eta, 0\}) - \sigma^*$ through adding an extra safety boundary $\sigma^*$. The resulting \textit{forward invariance} set and newly added boundary layer are illustrated in \Cref{fig:main_theorem}, where the \textit{forward invariance} set is the green area and the boundary layer is the blue area.

In summary, by discussing the two cases of whether $\mathcal{X}_S^D$ is the subset of $\mathcal{X}_S$, we have proved that if the safety index design follows the rule described in \Cref{sec:synthesis}, the implicit safe set algorithm guarantees the \textit{forward invariance} and \textit{finite time convergence} to the set $S \subseteq \mathcal{X}_S$.

\end{proof}

\section{Experimental Results}
\label{sec:experiment}
In our experiments, we aim to answer the following questions:
\begin{itemize}
    \item[\textbf{Q1}:] How does the discrepancy between continuous-time safe control and discrete-time safe control affect the evolution of the safety index in discrete-time systems? (Answered in \Cref{appdx:toy})
    \item[\textbf{Q2}:] How does ISSA compare with other state-of-the-art methods for safe RL? Can ISSA achieve zero-violation of the safety constraint? (Answered in \Cref{sec:q2a})
    \item[\textbf{Q3}:] How does the design of the safety index affect the set of safe control? (Answered in \Cref{sec:exp_feasibility})
    \item[\textbf{Q4}:] How do the hyper-parameters of ISSA \rebuttal{and the dimensionality of the system} impact its performance? (Answered in \Cref{sec:exp_ablation})
\end{itemize}

\subsection{Toy Problem Experiment Details}
\label{appdx:toy}

To demonstrate the discrepancy between continuous-time systems and discrete-time systems, we build a toy problem environment to show that directly applying the result from \eqref{eq: quadratic program for control} in discrete-time systems will lead to safety violations. The toy robot is a 3-state unicycle model with state $x = [p_x, p_y, \theta]$, where $p_x,p_y$ denote the coordinates on the x-axis and y-axis, respectively, and  $\theta$ is the heading direction. The control inputs of the toy environment are $[v, w]$, where $v$ denotes the velocity of the robot, and $w$ denotes the angular velocity of the robot. Next, we define the underlying dynamics of the robot for the toy environment in control affine form as follows: 
\begin{equation}\label{eq:dynamics}
\begin{split}
    x_{t+1} &= x_{t} + \dot{x}_tdt = x_{t} + \begin{bmatrix} \cos{(\theta)}dt & 0 \\ \sin{(\theta)}dt & 0 \\ 0 & dt \end{bmatrix} u
\end{split}
\end{equation}

\begin{wrapfigure}{r}{0.4\textwidth}
    \centering
    \includegraphics[width=0.35\textwidth]{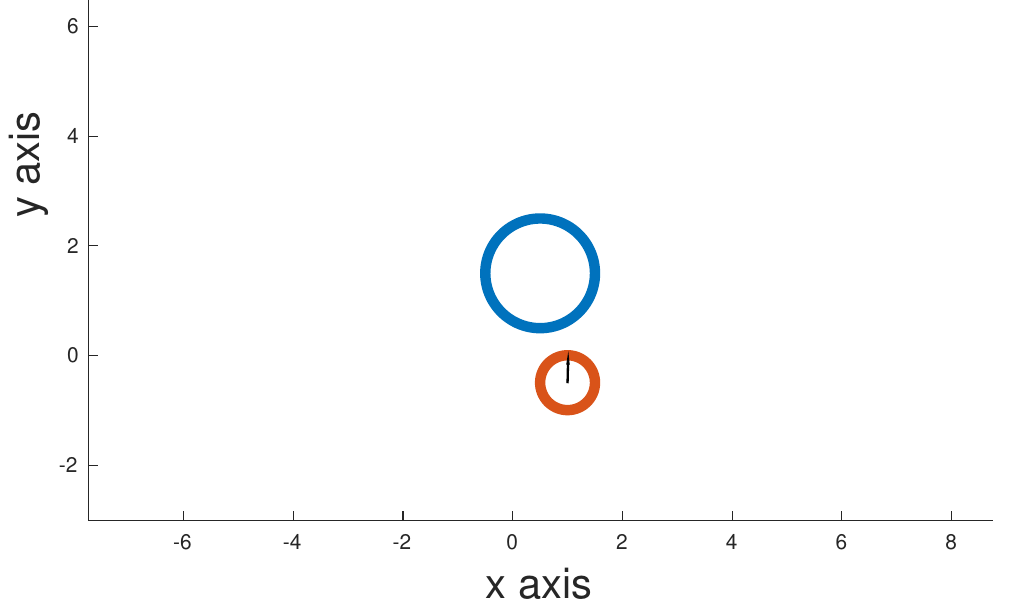}
    \caption{The toy experimental platform. The blue circle is a static obstacle, and red circle is the robot, whose heading direction is illustrated as the black arrow.}
    \label{fig:setup}
\end{wrapfigure}

\subsubsection{Experimental Setup: Toy Problem}
\label{sec:setup}
The toy environment is illustrated in \Cref{fig:setup}. Note that the environment may contain multiple obstacles, and the robot is required to be collision-free with all the obstacles. However, we can always consider the safety constraint between the robot and its closest obstacle at each time step. The justification is summarized in the following remark.
\begin{remark}
If the robot is collision-free with its closest obstacle at each time step, then the robot is collision-free with environmental obstacles at each time step. 
\end{remark}
Therefore, in \Cref{fig:setup} we only consider one obstacle. For simplicity, we currently assume the obstacle is static, which can be easily extended to moving obstacles.  

Next, we define the safety index $\phi(x)$ by constraining the perpendicular line from the center of the obstacle to the robot heading direction to be larger than $(R + r)$, which is shown as follows: 
\begin{equation}\label{eq:si}
    \phi(x) = (r + R)^2 - (({p_x}_0-p_x)\sin(\theta) - ({p_y}_0 - p_y)\cos(\theta))^2
\end{equation}

where $r, R$ are the radii of the robot and obstacle, respectively. $[{p_x}_0, {p_y}_0]$ is the location of the obstacle. It can be easily shown that $\phi(x) < 0$ will ensure the robot is collision-free with environmental obstacles. Finally, we design a dummy nominal policy to control the point robot to move forward with constant velocity, which has no safety guarantee. Therefore, if the initial system state is unsafe, our proposed methods are expected to generate the safe control to gradually drive the system state to the safe set where $\phi(x) < 0$, and remain in the safe set.

\subsubsection{Results: Toy Problem}

\begin{wrapfigure}{r}{0.4\textwidth}
    \centering
    \includegraphics[width=0.35\textwidth]{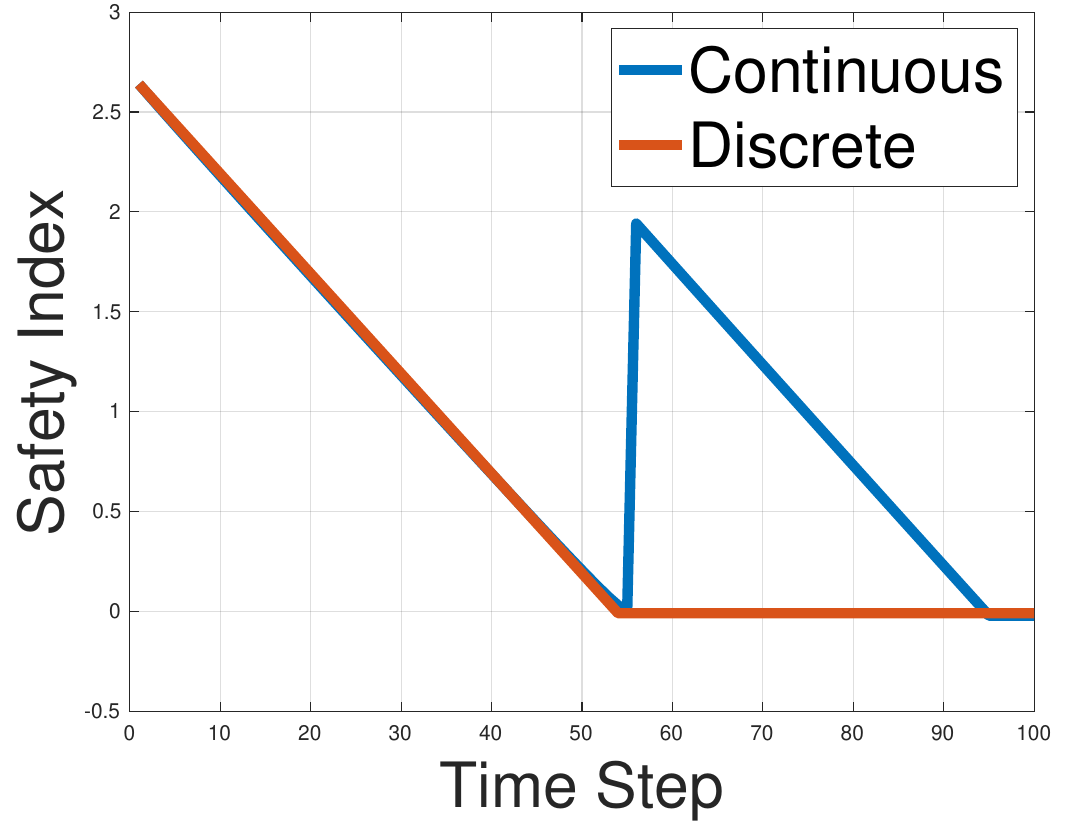}
    \caption{Safety Index evolution curves comparison.}
    \label{fig:comp}
    \vspace{-20pt}
\end{wrapfigure}

For toy problem, we simulate the system for $100$ time steps, and system $dt_{system} = 0.01$. 
We compare the safety index evolution for the safe control generated by: 1) solving \eqref{eq:adamba_discrete} with ISSA (i.e., $dt_{ISSA} = 0.01 = dt_{system}$ for ISSA simulation) and 2) sloving \eqref{eq: quadratic program for control} with ISSA (i.e., $dt_{ISSA} = 0.00001 \ll dt_{system}$ for ISSA simulation).
The comparison result is demonstrated in \Cref{fig:comp}.


We can see that the safe control generated by solving continuous-time system problem \eqref{eq: quadratic program for control} fails to ensure the safety index is monotonically decreasing due to \eqref{eq:discrepancy}, where the safety index suddenly increased to a large positive value at the time step between 50 and 60. However, the safe control generated by solving discrete-time system problem \eqref{eq:adamba_discrete} ensures the safety index is monotonically decreasing until below zero. Therefore, \Cref{fig:comp} strongly supports the discrepancy between the discrete-time system and the continuous-time system.

\subsection{Safety Gym Experiment Details}
In recent years, numerous benchmark environments for safe control have emerged \cite{ray2019benchmarking,ji2023omnisafeinfrastructureacceleratingsafe,zhao2024guardsafereinforcementlearning,sun2025sparkmodularbenchmarkhumanoid}. We adopt Safety Gym \cite{ray2019benchmarking} as our testing platform to evaluate the effectiveness of the proposed implicit safe set algorithms.
\begin{figure*}
     \centering
    \begin{subfigure}[t]{0.19\textwidth}
        \raisebox{-\height}{\includegraphics[width=\textwidth]{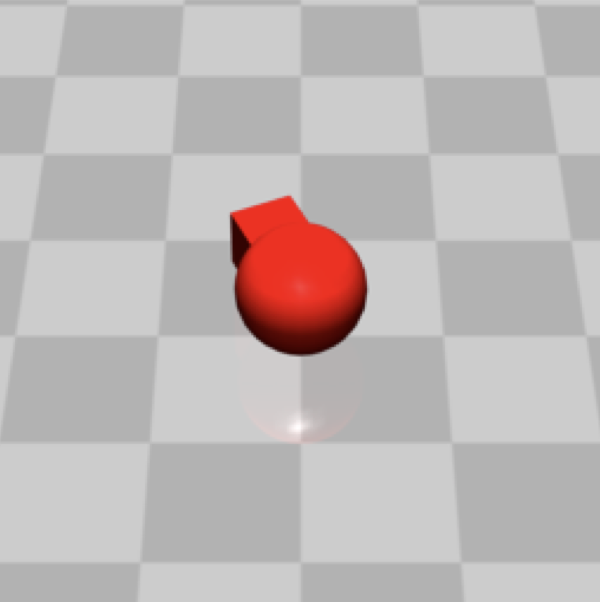}}
        \caption{Point robot: a simple 2D robot that can turn and move.}
        \label{fig:sg_point}
    \end{subfigure}
    \hfill
    \begin{subfigure}[t]{0.19\textwidth}
        \raisebox{-\height}{\includegraphics[width=\textwidth]{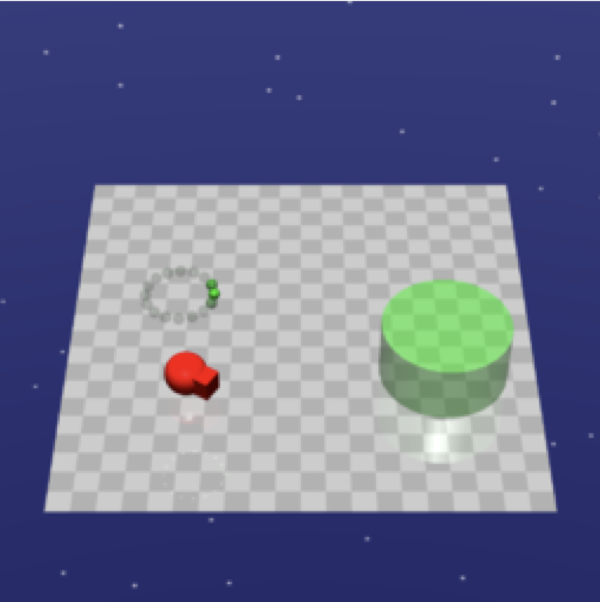}}
        \caption{Goal: navigating the robot inside the green goal area.}
        \label{fig:sg_goal}
    \end{subfigure}
    \hfill
    \begin{subfigure}[t]{0.19\textwidth}
        \raisebox{-\height}{\includegraphics[width=\textwidth]{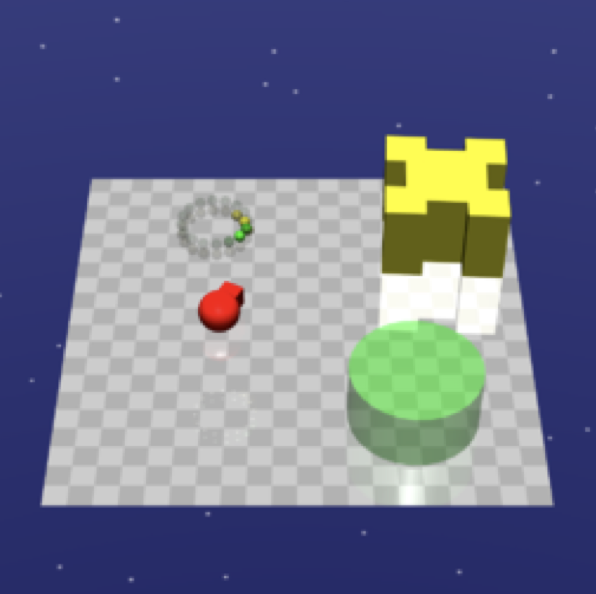}}
    \caption{Push: pushing the yellow box inside the green goal area.} 
    \label{fig:sg_push}
    \end{subfigure}
    \hfill
    \begin{subfigure}[t]{0.19\textwidth}
        \raisebox{-\height}{\includegraphics[width=\textwidth]{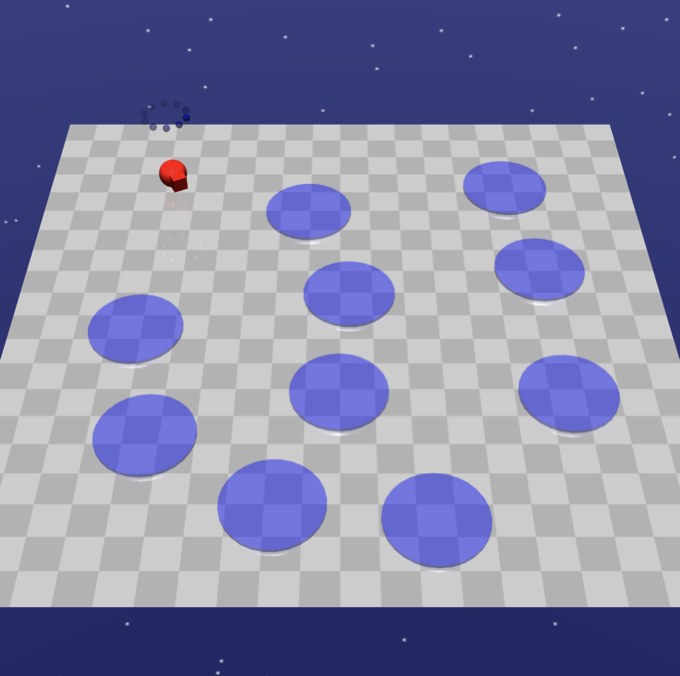}}
    \caption{Hazards: non-physical dangerous areas.} 
    \label{fig:sg_hazard}
    \end{subfigure}
    \hfill
    \begin{subfigure}[t]{0.19\textwidth}
        \raisebox{-\height}{\includegraphics[width=\textwidth]{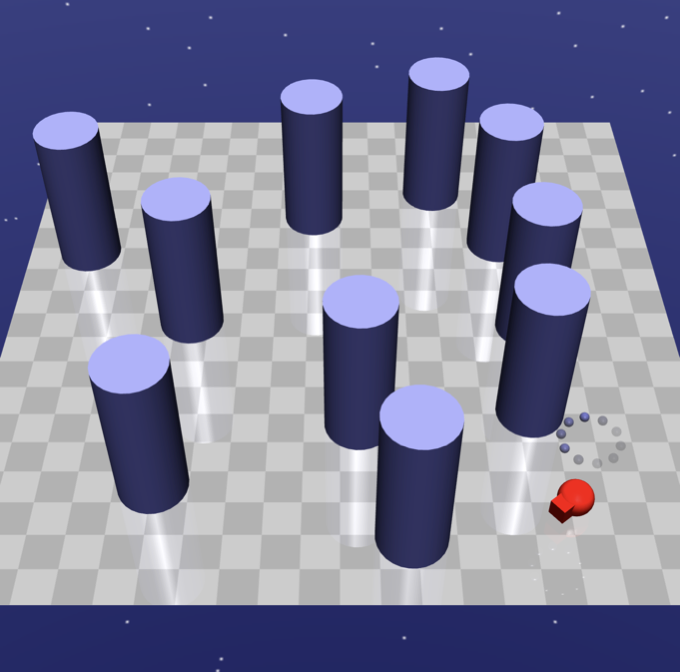}}
    \caption{Pillars: fixed dangerous obstacles} 
    \label{fig:sg_pillar}
    \end{subfigure}
    
     \caption{The environmental settings for benchmark problems in 
     Safety Gym.}
     \label{settings}
\end{figure*}
Our experiments adopt the Point robot ($\mathcal{U} \subseteq \mathbb{R}^2$) as shown in \Cref{fig:sg_point} and the Doggo robot ($\mathcal{U} \subseteq \mathbb{R}^{12}$) as shown in \Cref{fig:doggo_rule}.
We design 8 experimental environments with different task types, constraint types, constraint numbers and constraint sizes. We name these environments as \texttt{\{Task\}-\{Constraint Type\}\{Constraint Number\}-\{Constraint Size\}}. Note that \texttt{Constraint Size} equals $d_{min}$ in the safety index design. Two tasks are considered:
\begin{itemize}
    \item \texttt{Goal}: The robot must navigate to a goal as shown in~\Cref{fig:sg_goal}.
    \item \texttt{Push}: The robot must push a box to a goal as shown in~\Cref{fig:sg_push}.
\end{itemize}
And two different types of constraints are considered:
\begin{itemize}
    \item \texttt{Hazard}: Dangerous (but admissible) areas as shown in~\Cref{fig:sg_hazard}. Hazards are circles on the ground. The agent is penalized for entering them.
    \item \texttt{Pillar}: Fixed obstacles as shown in ~\Cref{fig:sg_pillar}. The agent is penalized for hitting them.
\end{itemize}

\begin{wrapfigure}{r}{0.4\textwidth}
    \centering
    \includegraphics[width=0.38\textwidth]{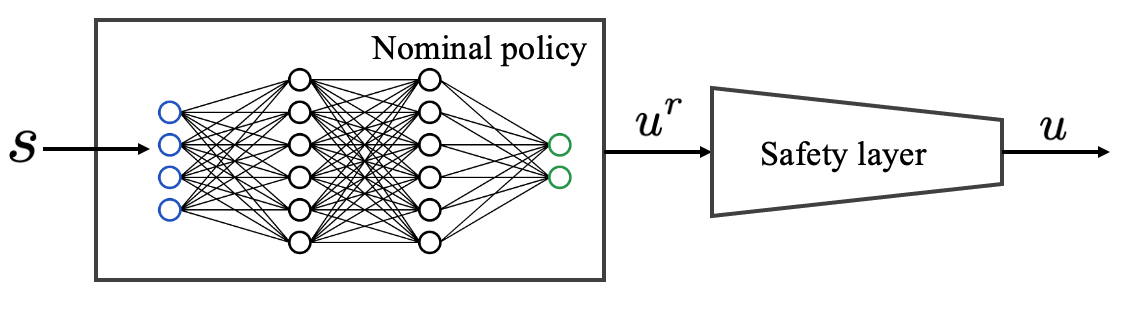}
    \caption{The PPO-ISSA structure.}
    \label{fig:agent}
\end{wrapfigure}

The methods in the comparison group include: unconstrained RL algorithm PPO~\cite{schulman2017proximal} and constrained safe RL algorithms PPO-Lagrangian, CPO~\cite{achiam2017constrained} \rebuttal{and PPO-SL (PPO-Safety Layer)~\cite{dalal2018safe}. 
We select PPO as our baseline method since it is state-of-the-art and already has safety-constrained derivatives that can be tested off-the-shelf.}
We set the limit of cost to 0 for both PPO-Lagrangian and CPO since we aim to avoid any violation of the constraints. To make sure ISSA can complete tasks while guaranteeing safety, we use a PPO agent as the nominal policy and ISSA as a safety layer to solve \eqref{eq:adamba_discrete}, we call this structure as PPO-ISSA, and it is illustrated in \Cref{fig:agent}. \rebuttal{Such safety layer structure has also been used in PPO-SL~\cite{dalal2018safe} which leverages offline dataset to learn a linear safety-signal model and then construct a safety layer via analytical optimization.}

For all experiments, we use neural network policies with separate feedforward MLP policy and value networks of size (256, 256) with tanh activations. 
More details are as follows.

\subsubsection{Environment Settings}
\paragraph{Goal Task}
In the Goal task environments, the reward function is:
\begin{equation}\notag
\begin{split}
    & r(x_t) = d^{g}_{t-1} - d^{g}_{t} + \mathbbm{1}[d^g_t < R^g]~,\\
\end{split}
\end{equation}
where $d^g_t$ is the distance from the robot to its closest goal and $R^g$ is the size (radius) of the goal. When a goal is achieved, the goal location is randomly reset to someplace new while keeping the rest of the layout the same.

\paragraph{Push Task}
In the Push task environments, the reward function is:
\begin{equation}\notag
\begin{split}
    & r(x_t) = d^{r}_{t-1} - d^{r}_{t} + d^{b}_{t-1} - d^{b}_{t} + \mathbbm{1}[d^b_t < R^g]~,\\
\end{split}
\end{equation}
where $d^r$ and $d^b$ are the distance from the robot to its closest goal and the distance from the box to its closest goal, and $R^g$ is the size (radius) of the goal. The box size is $0.2$ for all the Push task environments. Like the goal task, a new goal location is drawn each time a goal is achieved. 


\paragraph{Hazard Constraint}
In the Hazard constraint environments, the cost function is:
\begin{equation}\notag
\begin{split}
    & c(x_t) = \max(0, R^h - d^h_t)~,\\
\end{split}
\end{equation}
where $d^h_t$ is the distance to the closest hazard and $R^h$ is the size (radius) of the hazard.
\paragraph{Pillar Constraint}
In the Pillar constraint environments, the cost $c_t = 1$ if the robot contacts with the pillar otherwise $c_t = 0$. 

\paragraph{Black-box Dynamics}
The underlying dynamics of Safety Gym is directly handled by MuJoCo physics simulator~\cite{todorov2012mujoco}. This indicates the dynamics is not explicitly accessible but rather can be implicitly evaluated, which is suitable for our proposed implicit safe set algorithm. The implementation of block-box dynamics for ISSA is through simulation in the MuJoCo physics simulator and recovering to the pre-simulated state. 

\paragraph{State Space}
The state space is composed of various physical quantities from standard robot sensors (accelerometer, gyroscope, magnetometer, and velocimeter) and lidar (where each lidar sensor perceives objects of a single kind). The state spaces of all the test suites are summarized in \Cref{tab:state_space}. Note that Vase is another type of constraint in Safety Gym \cite{ray2019benchmarking} and all the returns of vase lidar are zero vectors (i.e., $[0, 0, \cdots, 0] \in \mathbb{R}^{16}$) in our experiments since none of our eight test suites environments have vases.

\begin{table*}[t]
\vskip 0.15in
\caption{The state space components of different test suites environments.}
\begin{center}
\begin{tabular}{c|ccc}
\toprule
\textbf{State Space Option} &  Goal-Hazard & Goal-Pillar & Push-Hazard\\
\hline
Accelerometer ($\mathbb{R}^3$) & \Checkmark & \Checkmark & \Checkmark\\
Gyroscope ($\mathbb{R}^3$) & \Checkmark & \Checkmark & \Checkmark\\
Magnetometer ($\mathbb{R}^3$) & \Checkmark & \Checkmark & \Checkmark\\
Velocimeter ($\mathbb{R}^{3}$) & \Checkmark & \Checkmark & \Checkmark\\
Goal Lidar ($\mathbb{R}^{16}$) & \Checkmark & \Checkmark & \Checkmark\\
Hazard Lidar ($\mathbb{R}^{16}$) & \Checkmark & \XSolid & \Checkmark\\
Pillar Lidar ($\mathbb{R}^{16}$) & \XSolid & \Checkmark & \XSolid\\
Vase Lidar ($\mathbb{R}^{16}$) & \Checkmark & \Checkmark & \Checkmark\\
Box Lidar ($\mathbb{R}^{16}$) & \XSolid & \XSolid & \Checkmark\\
\bottomrule
\end{tabular}
\label{tab:state_space}
\end{center}
\end{table*}

\paragraph{Control Space}
For all the experiments, the control space $\mathcal{U} \subset \mathbb{R}^2$. The first dimension $u_1 \in [-10, 10]$ is the control space of moving actuator, and second dimension $u_2 \in [-10, 10]$ is the control space of turning actuator. For each actuator the maximum torque corresponds to $a_{\max}=2.5\,\text{m/s}^2$ and $\omega_{\max}=4.0\,\text{rad/s}$; within these limits the inner-loop PID reaches the desired $(a,w)$ in $\le 6\,\text{ms}$, satisfying Assumption~1.

\subsubsection{Policy Settings}
Detailed parameter settings are shown in \Cref{tab:policy_setting}. All the policies in our experiments use the default hyper-parameter settings hand-tuned by Safety Gym~\cite{ray2019benchmarking} except the $\text{cost limit} = 0$ for PPO-Lagrangian and CPO. 
\begin{table*}[htbp]
\vskip 0.15in
\caption{Important hyper-parameters of PPO, PPO-Lagrangian, CPO, PPO-SL and PPO-ISSA}
\begin{center}
\begin{tabular}{c|cccc}
\toprule
\textbf{Policy Parameter} & PPO & PPO-Lagrangian & CPO & PPO-SL \& PPO-ISSA\\
\hline
Timesteps per iteration & 30000 & 30000 & 30000 & 30000 \\
Policy network hidden layers & (256, 256) & (256, 256) & (256, 256) & (256, 256) \\
Value network hidden layers & (256, 256) & (256, 256) & (256, 256) & (256, 256) \\
Policy learning rate & 0.0004 & 0.0004 & (N/A) & 0.0004 \\
Value learning rate & 0.001 & 0.001 & 0.001 & 0.001 \\
Target KL & 0.01 & 0.01 & 0.01 & 0.01 \\
Discounted factor $\gamma$ & 0.99 & 0.99 & 0.99 & 0.99 \\
Advantage discounted factor $\lambda$ & 0.97 & 0.97& 0.97 & 0.97 \\
PPO Clipping $\epsilon$ & 0.2 & 0.2  & (N/A) & 0.2 \\
TRPO Conjugate gradient damping & (N/A) & (N/A) & 0.1 & (N/A) \\
TRPO Backtracking steps & (N/A) & (N/A) & 10 & (N/A) \\
Cost limit & (N/A) & 0 & 0 & (N/A) \\
\bottomrule
\end{tabular}
\label{tab:policy_setting}
\end{center}
\end{table*}
\subsubsection{Safety Index Settings}
The parameters of safety index design are summarized in \Cref{tab:index_setting}, where we adopt $k=0.375$ for test suites with constraint size of 0.05 and $k=0.5$ for test suites with constraint size of 0.15.
\begin{table*}[htbp]
\vskip 0.15in
\caption{Experiment-specific parameters of safety index design for PPO-ISSA.}
\begin{center}
\begin{tabular}{c|cc}
\toprule
\textbf{Safety Index Parameter} & Constraint size = 0.05 & Constraint size = 0.15\\
\hline
n & 1 & 1 \\
k & 0.375 & 0.5 \\
$\eta$ & 0 & 0  \\
\bottomrule
\end{tabular}
\label{tab:index_setting}
\end{center}
\end{table*}

\subsection{Evaluating PPO-ISSA and Comparison Analysis}
\label{sec:q2a}
\begin{figure*}
    \centering
    \begin{subfigure}[t]{0.24\textwidth}
    \begin{subfigure}[t]{1.00\textwidth}
        \raisebox{-\height}{\includegraphics[width=\textwidth]{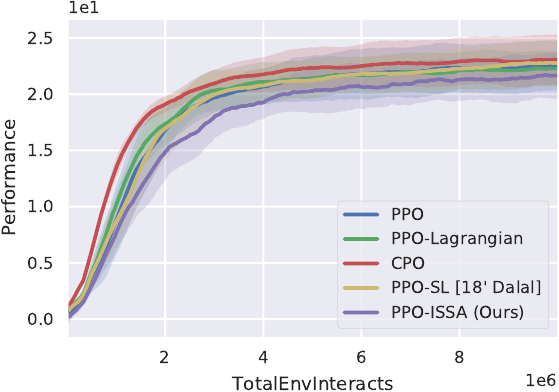}}
        \label{fig:goal-hazard4-0.05-Performance}
    \end{subfigure}
    \hfill
    \begin{subfigure}[t]{1.00\textwidth}
        \raisebox{-\height}{\includegraphics[width=\textwidth]{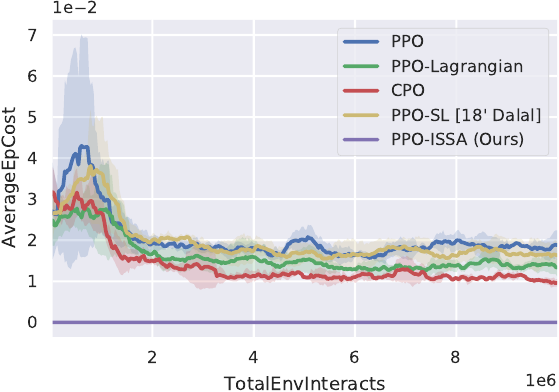}}
    \label{fig:goal-hazard4-0.05-AverageEpCost}
    \end{subfigure}
    \hfill
    \begin{subfigure}[t]{1.00\textwidth}
        \raisebox{-\height}{\includegraphics[width=\textwidth]{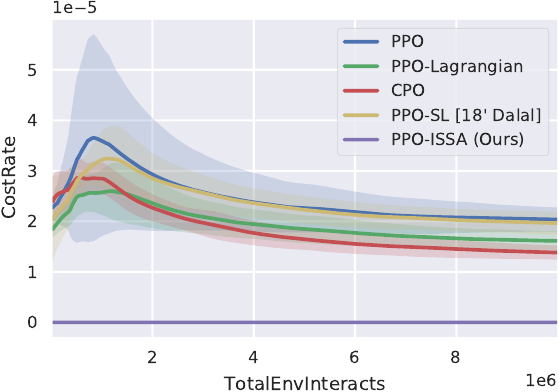}}
    \label{fig:goal-hazard4-0.05-CostRate}
    \end{subfigure}
    \caption{Goal-Hazard4-0.05}
    \label{fig:goal-hazard4-0.05}
    \end{subfigure}
    \begin{subfigure}[t]{0.24\textwidth}
    \begin{subfigure}[t]{1.00\textwidth}
        \raisebox{-\height}{\includegraphics[width=\textwidth]{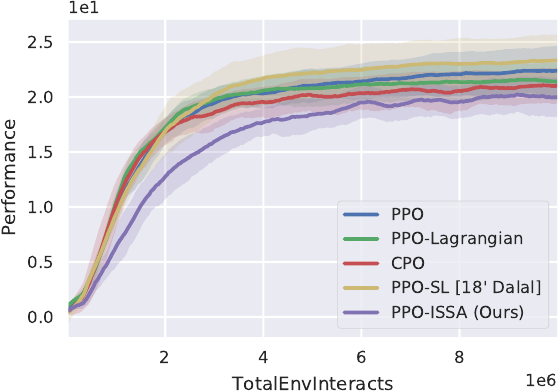}}
        \label{fig:goal-hazard4-0.15-Performance}
    \end{subfigure}
    \hfill
    \begin{subfigure}[t]{1.00\textwidth}
        \raisebox{-\height}{\includegraphics[width=\textwidth]{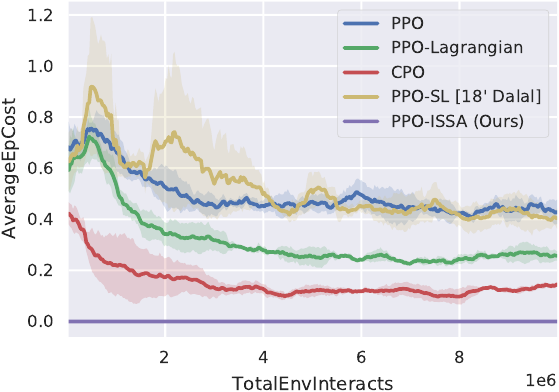}}
    \label{fig:goal-hazard4-0.15-AverageEpCost}
    \end{subfigure}
    \hfill
    \begin{subfigure}[t]{1.00\textwidth}
        \raisebox{-\height}{\includegraphics[width=\textwidth]{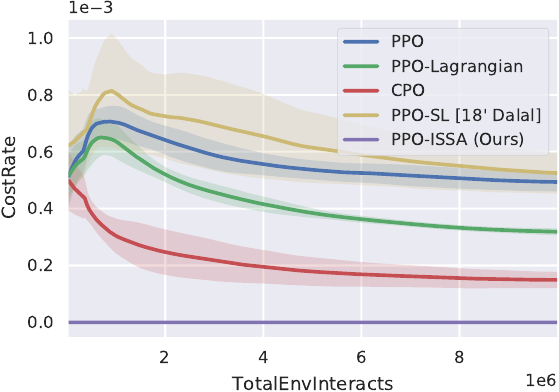}}
    \label{fig:goal-hazard4-0.15-CostRate}
    \end{subfigure}
    \caption{Goal-Hazard4-0.15}
    \label{fig:goal-hazard4-0.15}
    \end{subfigure}
    \begin{subfigure}[t]{0.24\textwidth}
    \begin{subfigure}[t]{1.00\textwidth}
        \raisebox{-\height}{\includegraphics[width=\textwidth]{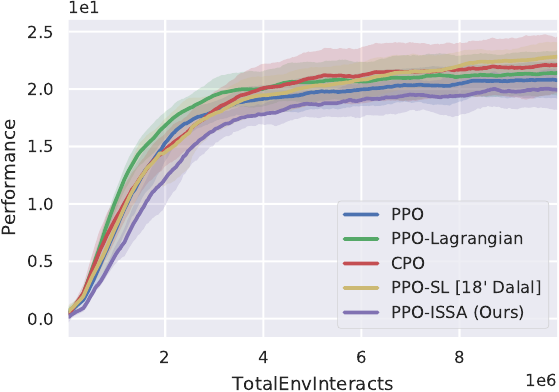}}
        \label{fig:goal-pillar4-0.15-Performance}
    \end{subfigure}
    \hfill
    \begin{subfigure}[t]{1.00\textwidth}
        \raisebox{-\height}{\includegraphics[width=\textwidth]{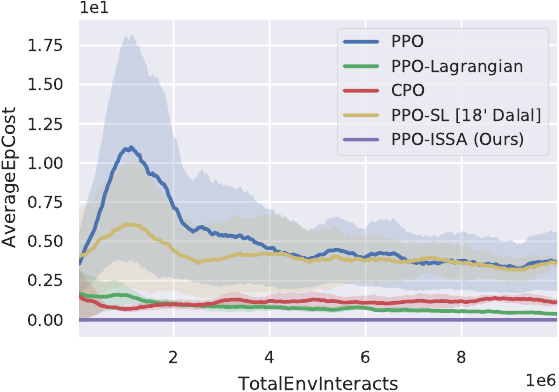}}
    \label{fig:goal-pillar4-0.15-AverageEpCost}
    \end{subfigure}
    \hfill
    \begin{subfigure}[t]{1.00\textwidth}
        \raisebox{-\height}{\includegraphics[width=\textwidth]{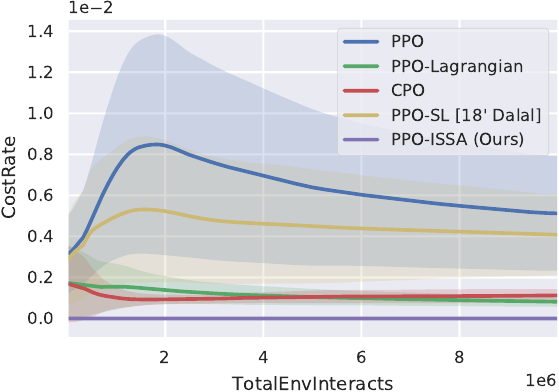}}
    \label{fig:goal-pillar4-0.15-CostRate}
    \end{subfigure}
    \caption{Goal-Pillar4-0.15}
    \label{fig:goal-pillar4-0.15}
    \end{subfigure}
    \begin{subfigure}[t]{0.24\textwidth}
    \begin{subfigure}[t]{1.00\textwidth}
        \raisebox{-\height}{\includegraphics[width=\textwidth]{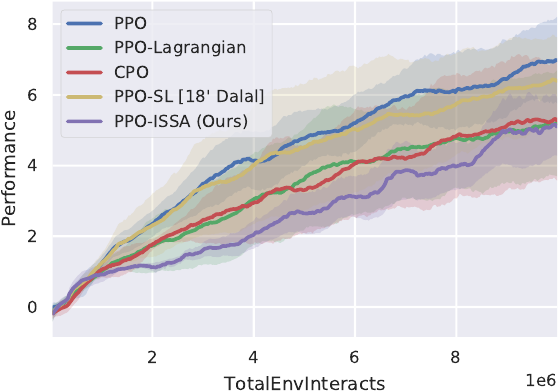}}
        \label{fig:push-hazard4-0.15-Performance}
    \end{subfigure}
    \hfill
    \begin{subfigure}[t]{1.00\textwidth}
        \raisebox{-\height}{\includegraphics[width=\textwidth]{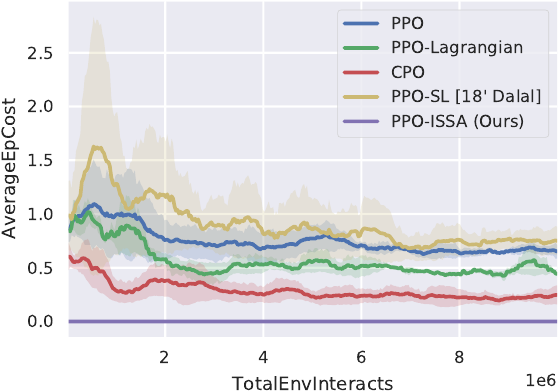}}
    \label{fig:push-hazard4-0.15-AverageEpCost}
    \end{subfigure}
    \hfill
    \begin{subfigure}[t]{1.00\textwidth}
        \raisebox{-\height}{\includegraphics[width=\textwidth]{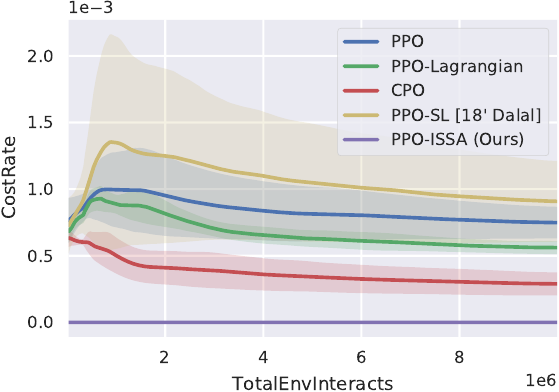}}
    \label{fig:push-hazard4-0.15-CostRate}
    \end{subfigure}
    \caption{Push-Hazard4-0.15}
    \label{fig:push-hazard4-0.15}
    \end{subfigure}
    \caption{
    \rebuttal{Average episodic return, episodic cost and overall cost rate of constraints of PPO-ISSA and baseline methods on 4-constraint environments over five seeds.}} 
    \label{fig:exp1-constraint-4}
\end{figure*}

\begin{figure*}
    \centering
    \begin{subfigure}[t]{0.24\textwidth}
    \begin{subfigure}[t]{1.00\textwidth}
        \raisebox{-\height}{\includegraphics[width=\textwidth]{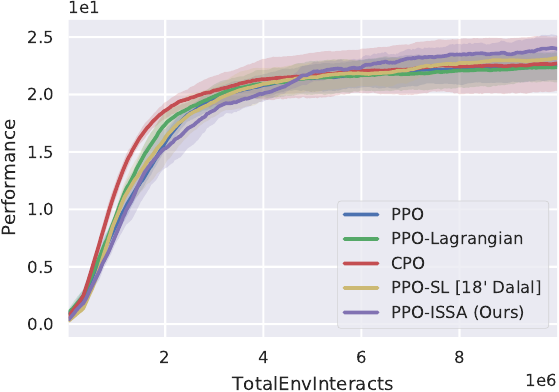}}
        \label{fig:goal-hazard1-0.05-Performance}
    \end{subfigure}
    \hfill
    \begin{subfigure}[t]{1.00\textwidth}
        \raisebox{-\height}{\includegraphics[width=\textwidth]{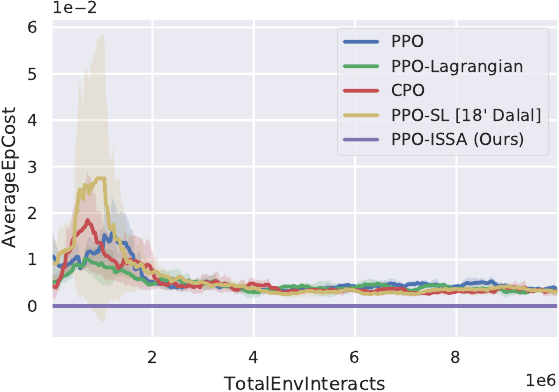}}
    \label{fig:goal-hazard1-0.05-AverageEpCost}
    \end{subfigure}
    \hfill
    \begin{subfigure}[t]{1.00\textwidth}
        \raisebox{-\height}{\includegraphics[width=\textwidth]{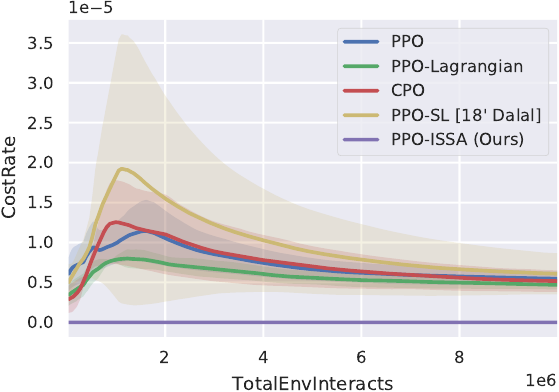}}
    \label{fig:goal-hazard1-0.05-CostRate}
    \end{subfigure}
    \caption{Goal-Hazard1-0.05}
    \label{fig:goal-hazard1-0.05}
    \end{subfigure}
    \begin{subfigure}[t]{0.24\textwidth}
    \begin{subfigure}[t]{1.00\textwidth}
        \raisebox{-\height}{\includegraphics[width=\textwidth]{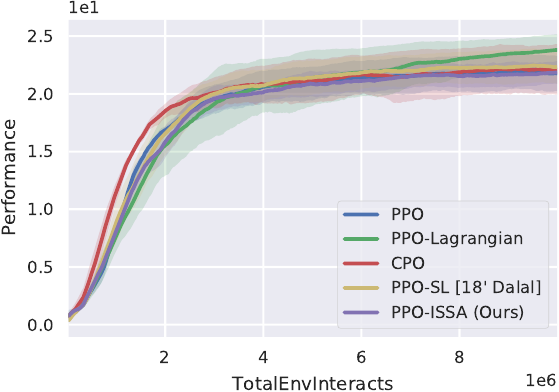}}
        \label{fig:goal-hazard1-0.15-Performance}
    \end{subfigure}
    \hfill
    \begin{subfigure}[t]{1.00\textwidth}
        \raisebox{-\height}{\includegraphics[width=\textwidth]{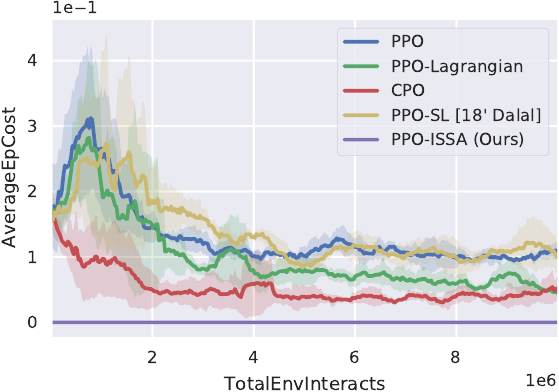}}
    \label{fig:goal-hazard1-0.15-AverageEpCost}
    \end{subfigure}
    \hfill
    \begin{subfigure}[t]{1.00\textwidth}
        \raisebox{-\height}{\includegraphics[width=\textwidth]{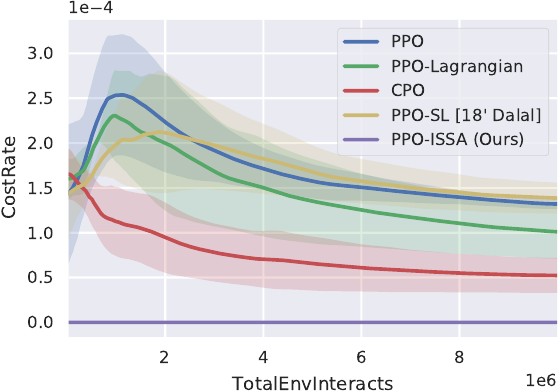}}
    \label{fig:goal-hazard1-0.15-CostRate}
    \end{subfigure}
    \caption{Goal-Hazard1-0.15}
    \label{fig:goal-hazard1-0.15}
    \end{subfigure}
    \begin{subfigure}[t]{0.24\textwidth}
    \begin{subfigure}[t]{1.00\textwidth}
        \raisebox{-\height}{\includegraphics[width=\textwidth]{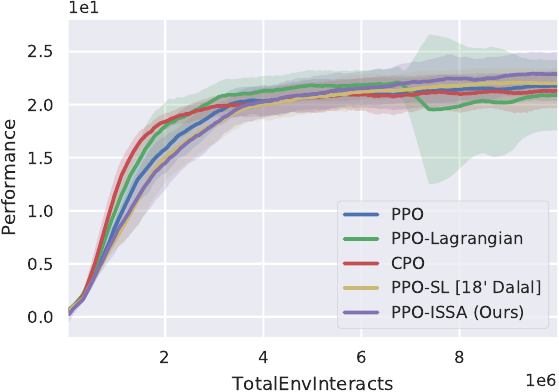}}
        \label{fig:goal-pillar1-0.15-Performance}
    \end{subfigure}
    \hfill
    \begin{subfigure}[t]{1.00\textwidth}
        \raisebox{-\height}{\includegraphics[width=\textwidth]{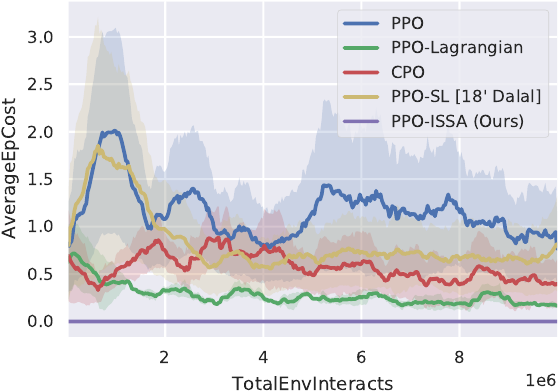}}
    \label{fig:goal-pillar1-0.15-AverageEpCost}
    \end{subfigure}
    \hfill
    \begin{subfigure}[t]{1.00\textwidth}
        \raisebox{-\height}{\includegraphics[width=\textwidth]{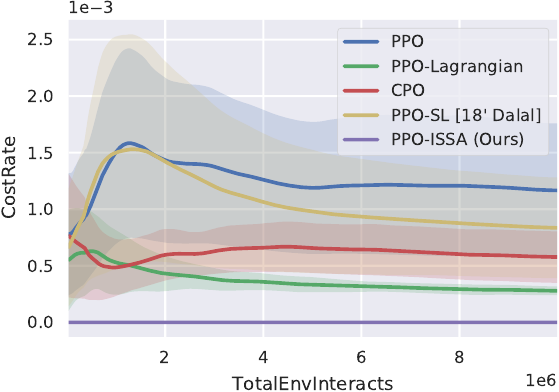}}
    \label{fig:goal-pillar1-0.15-CostRate}
    \end{subfigure}
    \caption{Goal-Pillar1-0.15}
    \label{fig:goal-pillar1-0.15}
    \end{subfigure}
    \begin{subfigure}[t]{0.24\textwidth}
    \begin{subfigure}[t]{1.00\textwidth}
        \raisebox{-\height}{\includegraphics[width=\textwidth]{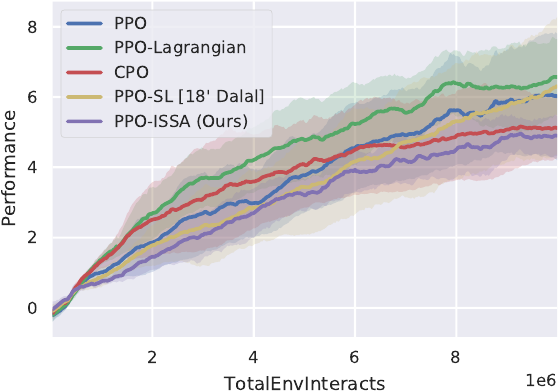}}
        \label{fig:push-hazard1-0.15-Performance}
    \end{subfigure}
    \hfill
    \begin{subfigure}[t]{1.00\textwidth}
        \raisebox{-\height}{\includegraphics[width=\textwidth]{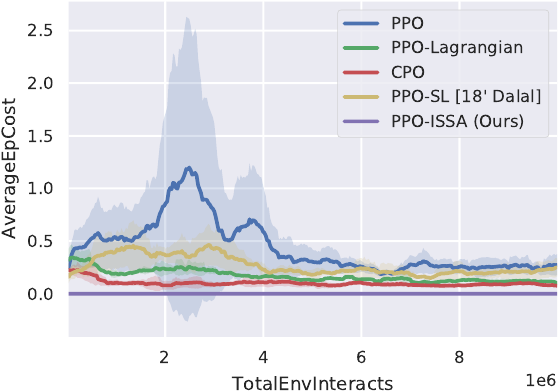}}
    \label{fig:push-hazard1-0.15-AverageEpCost}
    \end{subfigure}
    \hfill
    \begin{subfigure}[t]{1.00\textwidth}
        \raisebox{-\height}{\includegraphics[width=\textwidth]{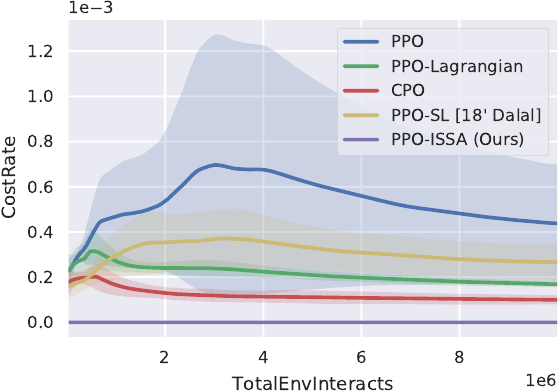}}
    \label{fig:push-hazard1-0.15-CostRate}
    \end{subfigure}
    \caption{Push-Hazard1-0.15}
    \label{fig:push-hazard1-0.15}
    \end{subfigure}
    \caption{
    \rebuttal{Average performance of PPO-ISSA and baseline methods on 1-constraint environments over five random seeds. The three rows represent average episodic return, average episodic cost and overall cost rate of constraints. The safety advantage of safe RL baseline methods over unconstrained RL method (PPO) becomes trivial as the constraint number and constraint size decrease, where the cost rate and average episode cost of PPO-Lagrangian, CPO and PPO are nearly the same when there is only one constraint with size 0.05. }}
    \label{fig:exp1-constraint-1}
\end{figure*}

To compare the reward and safety performance of PPO-ISSA to the baseline methods in different tasks, constraint types, and constraint sizes, we design 
4 test suites with 4 constraints which are summarized in \Cref{fig:exp1-constraint-4}.
The comparison results reported in \Cref{fig:exp1-constraint-4} demonstrate that 
\rebuttal{PPO-ISSA is able to achieve zero average episode cost and zero cost rate across all experiments while slightly sacrificing the reward performance.}
The baseline soft safe RL methods (PPO-Lagrangian and CPO) fail to achieve zero-violation safety even when the cost limit is set to be 0. 
Moreover, the safety advantage of safe RL baseline methods over unconstrained RL method (PPO) becomes trivial as the constraint number and constraint size decrease as shown in \Cref{fig:goal-hazard1-0.05}, where the cost rate and average episode cost of PPO-Lagrangian, CPO and PPO are  nearly the same when there is only one constraint with  size 0.05.

\begin{wrapfigure}{r}{0.4\textwidth}
    \vspace{-10pt}
    \centering
    \includegraphics[width=0.28\textwidth]{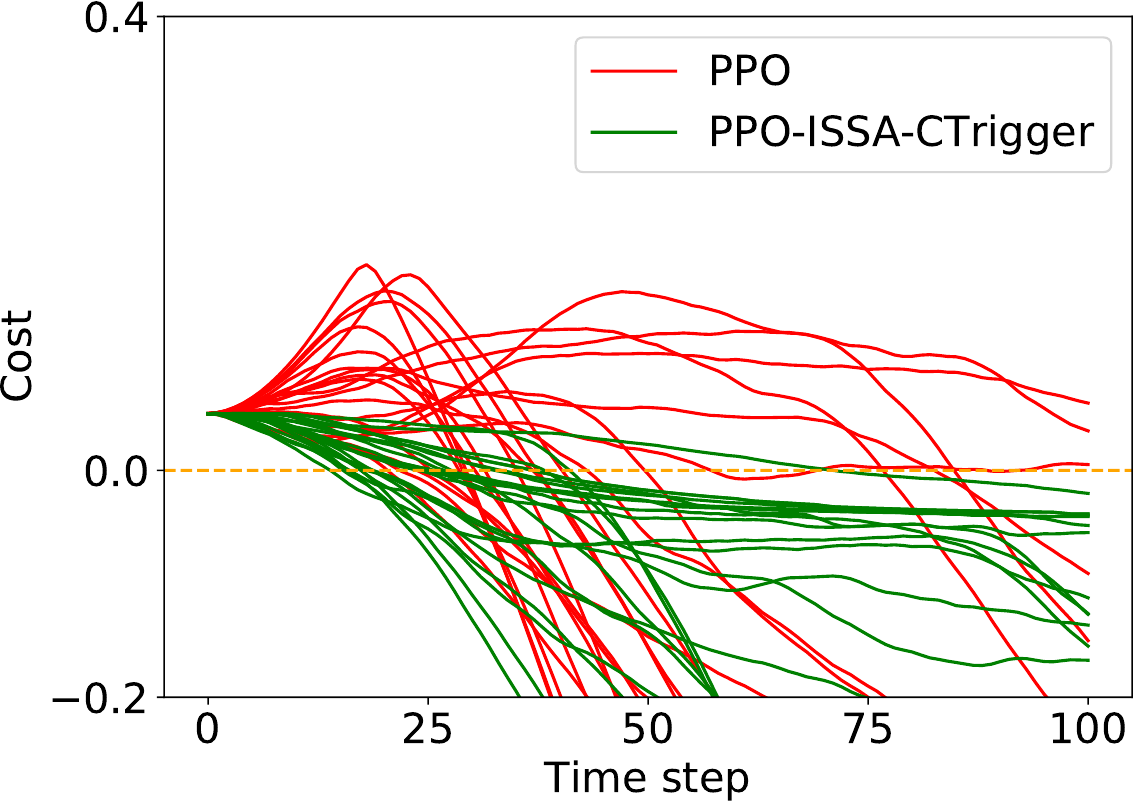}
    \caption{Cost changes over 100 time steps of PPO and PPO-ISSA-CTrigger starting from the same unsafe state over 20 trails.}
    \label{fig:finite}
    \vspace{-10pt}
\end{wrapfigure}

PPO-Lagrangian and CPO fail since both methods rely on trial-and-error to enforce constraints while ISSA is able to guarantee \textit{forward invariance} by \Cref{thoem:main}. \rebuttal{We also observe that PPO-SL fails to lower the violation during training, due to the fact that the linear approximation of cost function $ c(x_{t+1}) \approx c(x_t) + g(x_t, w)^T u $~\cite{dalal2018safe} becomes inaccurate when the dynamics are highly nonlinear like the ones we used in MuJoCo~\cite{todorov2012mujoco}. More importantly, PPO-SL cannot guarantee that these always exist a feasible safe control to lower the cost, since they directly use the user defined cost function which cannot always ensure feasibility.} More detailed metrics for comparison and experimental results on test suites with 1 constraint are summarized in \Cref{sec:metrics}.

\subsubsection{Metrics Comparison}
\label{sec:metrics}
In this section, we report all the results of eight test suites by three metrics defined in Safety Gym \cite{ray2019benchmarking}:
\begin{itemize}
    \item The average episode return $J_r$.
    \item The average episodic sum of costs $M_c$.
    \item The average cost over the entirety of training $\rho_c$.
\end{itemize}
The average episode return $J_r$ and the average episodic sum of costs $M_c$ were obtained by averaging over the last five epochs of training to reduce noise. Cost rate $\rho_c$ was just taken from the final epoch. We report the results of these three metrics in \Cref{tab:metrics} normalized by PPO results. \rebuttal{We calculate the converged reward $\bar{J}_r$ percentage of PPO-ISSA compared to other three safe RL baseline methods (PPO-Lagrangian, CPO and PPO-SL) over eight control suites. The computed mean reward percentage is $95\%$, and the standard deviation is $9 \%$. Therefore we conclude that PPO-ISSA is able to gain $95\% \pm 9\% $ cumulative reward compared to state-of-the-art safe DRL methods.} 

\rebuttal{Note that in safety-critical environments, there is always a tradeoff between reward performance and safety, where safety guarantees prevent aggressive strategies for seeking high reward. On the other hand, the provable safety is prominently weighted in the tradeoff, since any safety violation may lead to property loss, life danger in the real robotics applications. Therefore, PPO-ISSA does a better job in terms of balancing the tradeoff between provable safety and task performance.}

\begin{table}[htbp]
\caption{Normalized metrics obtained from the policies at the end of the training process, which is averaged over eight test suits environments and five random seeds.}
\begin{subtable}[t]{0.4\textwidth}
\caption{Goal-Hazard1-0.05}
\vspace{-0.1in}
\begin{center}
\resizebox{!}{1.3cm}{
\begin{tabular}{c|ccc}
\toprule
\textbf{Algorithm} & $\bar{J}_r$ & $\bar{M}_c$ & $\bar{\rho}_c$\\
\hline
PPO & 1.00 & 1.00 & 1.00\\
PPO-Lagrangian & 1.003  & 1.587  & 0.859\\
CPO & 1.012 & 1.052  & 0.944 \\
PPO-SL [18' Dalal] & 1.038  & 1.031 & 1.110\\
PPO-ISSA (Ours) & \textbf{1.077}  & \textbf{0.000} & \textbf{0.000}\\
\bottomrule
\end{tabular}
}
\label{tab:goal_1_hazard_0.05}
\end{center}
\end{subtable}
\hfill
\begin{subtable}[t]{0.4\textwidth}
\caption{Goal-Hazard4-0.05}
\vspace{-0.1in}
\begin{center}
\resizebox{!}{1.3cm}{
\begin{tabular}{c|ccc}
\toprule
\textbf{Algorithm} & $\bar{J}_r$ & $\bar{M}_c$ & $\bar{\rho}_c$\\
\hline
PPO & 1.000 & 1.000 & 1.000\\
PPO-Lagrangian & 0.983  & 0.702  & 0.797\\
CPO & \textbf{1.022} & 0.549  & 0.676 \\
PPO-SL [18' Dalal] & 1.014  & 0.923 & 0.963\\
PPO-ISSA (Ours) & 0.961  & \textbf{0.000} & \textbf{0.000}\\
\bottomrule
\end{tabular}
}
\label{tab:goal_4_hazard_0.05}
\end{center}
\end{subtable}

\begin{subtable}[t]{0.4\textwidth}
\caption{Goal-Hazard1-0.15}
\vspace{-0.1in}
\begin{center}
\resizebox{!}{1.3cm}{
\begin{tabular}{c|ccc}
\toprule
\textbf{Algorithm} & $\bar{J}_r$ & $\bar{M}_c$ & $\bar{\rho}_c$\\
\hline
PPO & 1.000 & 1.000 & 1.000\\
PPO-Lagrangian & \textbf{1.086}  & 0.338  & 0.760\\
CPO & 1.011 & 0.553  & 0.398 \\
PPO-SL [18' Dalal] & 1.018  & 0.898 & 1.048\\
PPO-ISSA (Ours) & 1.008  & \textbf{0.000} & \textbf{0.000}\\
\bottomrule
\end{tabular}
}
\label{tab:goal_1_hazard_0.15}
\end{center}
\end{subtable}
\hfill
\begin{subtable}[t]{0.4\textwidth}
\caption{Goal-Hazard4-0.15}
\vspace{-0.1in}
\begin{center}
\resizebox{!}{1.3cm}{
\begin{tabular}{c|ccc}
\toprule
\textbf{Algorithm} & $\bar{J}_r$ & $\bar{M}_c$ & $\bar{\rho}_c$\\
\hline
PPO & 1.000 & 1.000 & 1.000\\
PPO-Lagrangian & 0.948  & 0.581  & 0.645\\
CPO & 0.932 & 0.328  & 0.303 \\
PPO-SL [18' Dalal] & \textbf{1.038}  & 0.948 & 1.063\\
PPO-ISSA (Ours) & 0.895  & \textbf{0.000} & \textbf{0.000}\\
\bottomrule
\end{tabular}
}
\label{tab:goal_4_hazard_0.15}
\end{center}
\end{subtable}

\begin{subtable}[t]{0.4\textwidth}
\caption{Goal-Pillar1-0.15}
\vspace{-0.1in}
\begin{center}
\resizebox{!}{1.3cm}{
\begin{tabular}{c|ccc}
\toprule
\textbf{Algorithm} & $\bar{J}_r$ & $\bar{M}_c$ & $\bar{\rho}_c$\\
\hline
PPO & 1.000 & 1.000 & 1.000\\
PPO-Lagrangian & 0.968  & 0.196  & 0.239\\
CPO & 0.976 & 0.328  & 0.494 \\
PPO-SL [18' Dalal] & 1.017  & 0.948 & 1.063\\
PPO-ISSA (Ours) & \textbf{1.056}  & \textbf{0.000} & \textbf{0.000}\\
\bottomrule
\end{tabular}
}
\label{tab:goal_1_pillar_0.15}
\end{center}
\end{subtable}
\hfill
\begin{subtable}[t]{0.4\textwidth}
\caption{Goal-Pillar4-0.15}
\vspace{-0.1in}
\begin{center}
\resizebox{!}{1.3cm}{
\begin{tabular}{c|ccc}
\toprule
\textbf{Algorithm} & $\bar{J}_r$ & $\bar{M}_c$ & $\bar{\rho}_c$\\
\hline
PPO & 1.000 & 1.000 & 1.000\\
PPO-Lagrangian &1.035  & 0.105  & 0.159\\
CPO & 1.060 & 0.304  & 0.221 \\
PPO-SL [18' Dalal] & \textbf{1.094} & 1.055 &0.780\\
PPO-ISSA (Ours) & 0.965  & \textbf{0.000} & \textbf{0.000}\\
\bottomrule
\end{tabular}
}
\label{tab:goal_4_pillar_0.15}
\end{center}
\end{subtable}
\hfill
\begin{subtable}[t]{0.4\textwidth}
\caption{Push-Hazard1-0.15}
\vspace{-0.1in}
\begin{center}
\resizebox{!}{1.3cm}{
\begin{tabular}{c|ccc}
\toprule
\textbf{Algorithm} & $\bar{J}_r$ & $\bar{M}_c$ & $\bar{\rho}_c$\\
\hline
PPO & 1.000 & 1.000 & 1.000\\
PPO-Lagrangian & \textbf{1.124}  & 0.356  & 0.384\\
CPO & 0.872 & 0.231  & 0.228 \\
PPO-SL [18' Dalal] & 1.107 & 0.685 & 0.610\\
PPO-ISSA (Ours) & 0.841  & \textbf{0.000} & \textbf{0.000}\\
\bottomrule
\end{tabular}
}
\label{tab:push_1_hazard_0.15}
\end{center}
\end{subtable}
\hfill
\begin{subtable}[t]{0.4\textwidth}
\caption{Push-Hazard4-0.15}
\vspace{-0.1in}
\begin{center}
\resizebox{!}{1.3cm}{
\begin{tabular}{c|ccc}
\toprule
\textbf{Algorithm} & $\bar{J}_r$ & $\bar{M}_c$ & $\bar{\rho}_c$\\
\hline
PPO & \textbf{1.000} & 1.000 & 1.000\\
PPO-Lagrangian & 0.72  & 0.631  & 0.748\\
CPO & 0.758 & 0.328  & 0.385 \\
PPO-SL [18' Dalal] & 0.914 & 1.084 & 1.212\\
PPO-ISSA (Ours) & 0.727  & \textbf{0.000} & \textbf{0.000}\\
\bottomrule
\end{tabular}
}
\label{tab:push_4_hazard_0.15}
\end{center}
\end{subtable}
\label{tab:metrics}
\end{table}

\begin{wrapfigure}{r}{0.33\textwidth}
    \vspace{-10pt}
    \centering
    \includegraphics[width=0.26\textwidth]{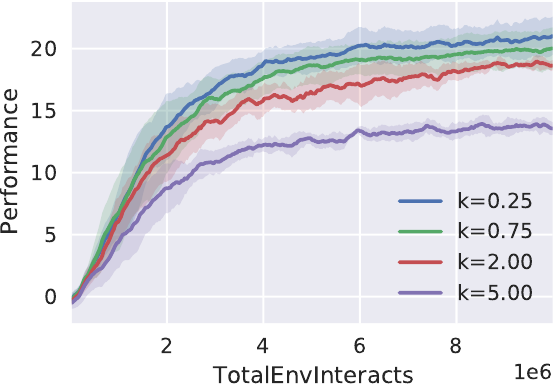}
    \caption{Average return of PPO-ISSA with different safety index design on Goal-Hazard4-0.15.}
    \label{fig:different_k}
    \vspace{-20pt}
\end{wrapfigure}

To validate the \textit{finite time convergence} of \Cref{thoem:discrete} that ISSA combined with CTrigger algorithm (PPO-ISSA-CTrigger) can ensure the \textit{finite time convergence} to $\mathcal{X}_S$ for discrete-time system, we further compare the cost evolution of PPO and PPO-ISSA-CTrigger agents when starting from the same unsafe state (i.e., cost $> 0$). The comparison results are shown in \Cref{fig:finite}, which shows that PPO-ISSA-CTrigger
can converge to $\mathcal{X}_S$ within 100 time steps across all experiments while cost evolution of PPO agents fluctuates wildly without preference to converge to safe set. The cost changes of PPO-ISSA-CTrigger aligns with ours theory of \textit{finite time convergence}.

\subsection{Feasibility of Safety Index Synthesis}
\label{sec:exp_feasibility}
\begin{figure*}
    \centering
    \resizebox{0.9\textwidth}{!}{
    \centering
    \begin{subfigure}[t]{0.23\textwidth}
        \centering
        \raisebox{-1.1\height}{\includegraphics[width=0.96\textwidth]{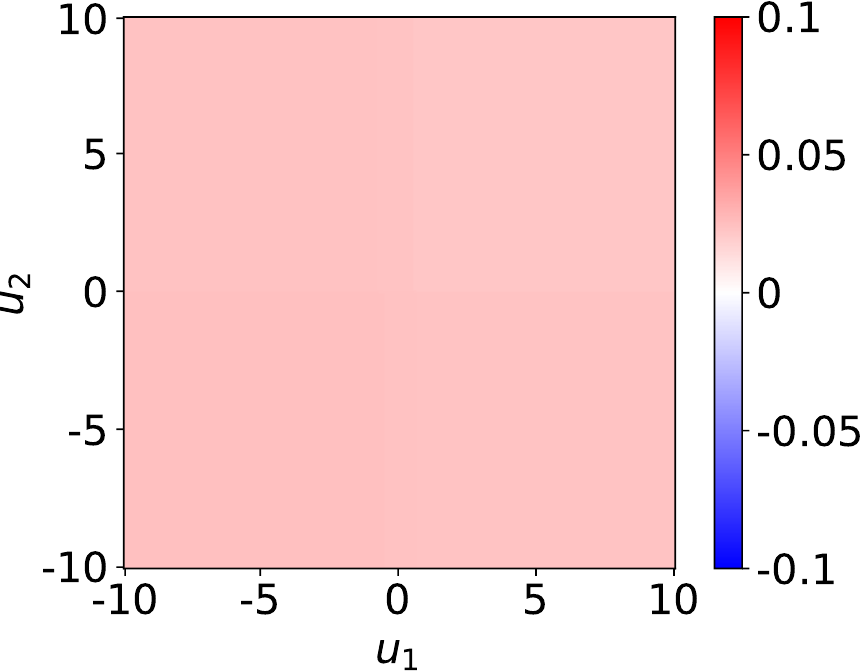}}
        \caption{Distance Safety Index}
        \label{fig:heatmap_dis}
    \end{subfigure}
    \hfill
    \begin{subfigure}[t]{0.75\textwidth}
        \raisebox{-\height}{\includegraphics[width=\textwidth]{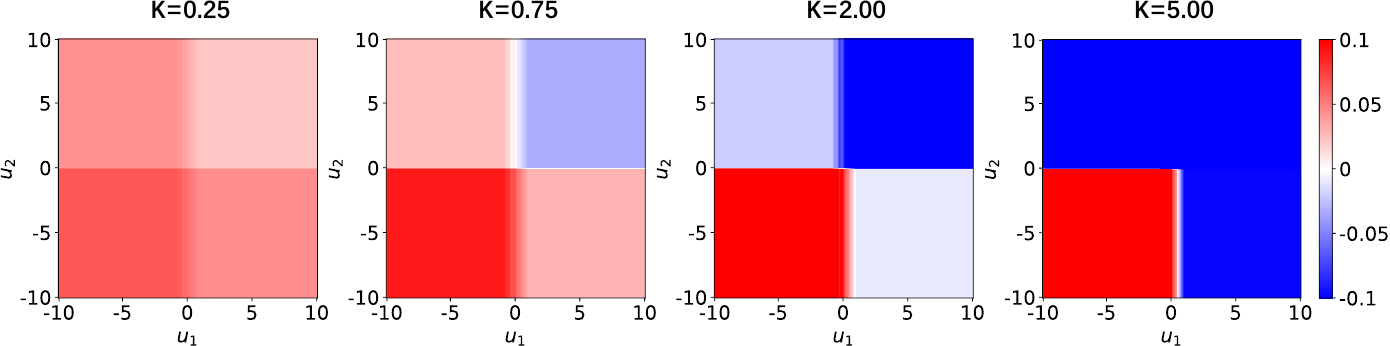}}
        \caption{Synthesized Safety Index}
        \label{fig:heatmap_k}
    \end{subfigure}}
    \hfill
     \caption{
     Heat maps of the difference of safety index $\Delta \phi = \phi(f(x, u)) - \phi(x)$. The x-axis $u_1$ represents the control space of moving actuator, and the y-axis $u_2$ represents the control space of turning actuator.}
     \label{fig:heatmap}
\end{figure*}

To demonstrate how the set of safe control is impacted by different safety index definition,  we randomly pick an unsafe state $x^*$ such that $\phi(x^*) > 0$, and visualize the corresponding set of safe control ${ \mathcal{U}}_S^D$ under different safety index definitions, which are shown in \Cref{fig:heatmap}. 
Red area means $\Delta \phi > 0$ (i.e. unsafe control) and blue area means $\Delta \phi < 0$ (i.e. safe control). 
\Cref{fig:heatmap_dis} shows the set of safe control of distance safety index $ \phi_d = \sigma + d_{min} - d$, which is the default cost definition of Safety Gym. The heatmap is all red in \Cref{fig:heatmap_dis}, which means that the set of safe control under the default $\phi_d$ is empty. \Cref{fig:heatmap_k} shows the set of safe control of the synthesized safety index $\phi = \sigma + d_{min}^2 -d^2 - k\dot{d}$ with different value of $k$. With the synthesized safety index, \Cref{fig:heatmap_k} demonstrates that the size of the set of safe control grows as $k$ increases, which aligns with the safety index synthesis rule discussed in \Cref{sec:synthesis} as larger $k$ is easier to satisfy \eqref{eq:continuous_rule}.
To demonstrate the reward performance of PPO-ISSA under different safety index designs, we select Goal-Hazard4-0.15 test suite. \Cref{fig:different_k} demonstrates the average return of PPO-ISSA under different value of $k$, which shows that the reward performance of PPO-ISSA deteriorates as $k$ value increases (since larger $k$ makes the control more conservative). Note that the set of safe control increases as the $k$ value increases, thus the optimal $k$ should be the smallest $k$ that makes the set of safe control nonempty for all states. Our safety index synthesis rule in \eqref{eq:continuous_rule} provides the condition to pick the optimal $k$. 
\subsection{Ablation Study}
\label{sec:exp_ablation}
\subsubsection{Sensitivity Analysis}


\begin{table}
\begin{center}
\resizebox{0.6\textwidth}{!}{
\begin{tabular}{c|ccc}
\toprule
Number of vectors & Simulation Time $T_{sim}$ &  Overall ISSA Time $T_{all}$  & Return $\bar{J}_r$\\
\hline
$n=3$ & 0.297 & 0.301 & 0.738\\
$n=5$ & 0.504 & 0.511  & 0.826\\
$n=10$ & 0.987 & 1.000  & 1.000 \\
\bottomrule
\end{tabular}
}
\vskip 5pt
\caption{Normalized computation time and return under different number of vectors in ISSA. These results are average on 100 ISSA runs over five random seeds on Goal-Hazard4-0.15.}
\label{tab:ablation}
\end{center}
\vskip -0.3in
\end{table}

To demonstrate the scalability and the performance of PPO-ISSA when ISSA chooses different parameters, we conduct additional tests using the test suite Goal-Hazard4-0.15. 
Among all input parameters of ISSA, the gradient vector number $n$ is critical to impact the quality of the solution of \eqref{eq:adamba_discrete}. Note in the limit when $n \rightarrow \infty$, ISSA is able to traverse all boundary points of the set of safe control, hence able to find the global optima of \eqref{eq:adamba_discrete}. We pick three different $n$ values: 3, 5, 10; and report the average episode reward of PPO-ISSA, and the computation time of ISSA when solving \eqref{eq:adamba_discrete}, which includes the normalized average ISSA computation time and the normalized average simulation time for each run. The results are summarized in \Cref{tab:ablation}, which demonstrates that the reward performance of PPO-ISSA would improve as $n$ gets bigger since we get better optima of \Cref{eq:adamba_discrete}. 
In practice, we find that the reward performance will stop improving when $n$ is big enough ($n>10$). 
\begin{figure*}
    \centering
    \begin{subfigure}[b]{0.4\textwidth}
        \raisebox{-\height}{\includegraphics[width=\textwidth]{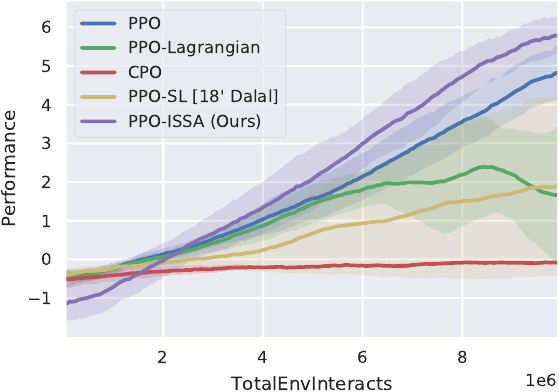}}
        \caption{Average reward}
    \end{subfigure}
    \hspace{2cm}
    \begin{subfigure}[b]{0.4\textwidth}
        \raisebox{-\height}{\includegraphics[width=\textwidth]{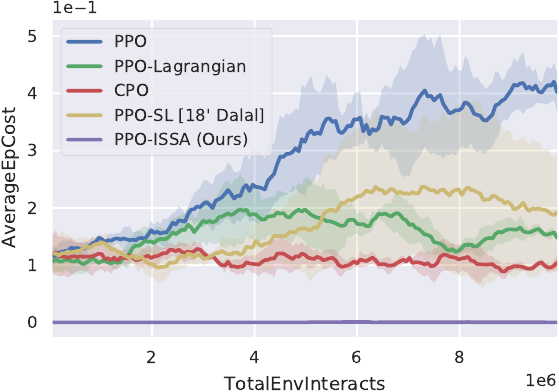}}
        \caption{Average cost}
    \end{subfigure}
    \hfill
     \caption{
     \rebuttal{Average episodic return and episodic cost of PPO-ISSA and baseline methods on Goal-hazard1-0.15 environment of a doggo robot over five seeds.}} 
     \label{fig:doggo_goal-hazard1-0.15}
\end{figure*}
The computation time scales linearly with respect to $n$ while the majority ($98\%$) of computation cost is used for environment simulation, which can be improved in the future by replacing the simulator with a more computationally efficient surrogate model.


\subsubsection{Scalability Analysis}
\label{sec:scalability}
The safe control algorithm ISSA is based on the sampling method \cref{alg:adamba} AdamBA. In this section, we demonstrate the scalability of AdamBA and ISSA in systems with higher dimensions of state and control.

\begin{figure*}[htbp]
    \centering
    \begin{subfigure}[b]{0.19\textwidth}
     
        \raisebox{-\height}{\includegraphics[width=1\textwidth]{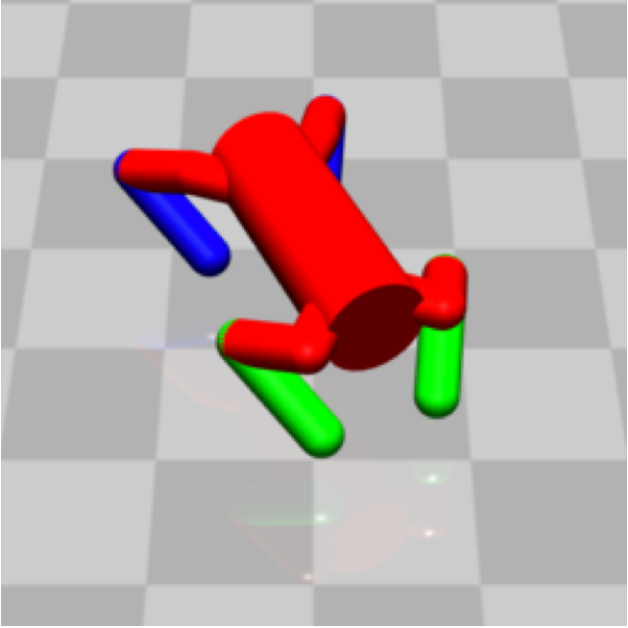}}
    \end{subfigure}
    \hspace{2cm}
    \begin{subfigure}[b]{0.19\textwidth}
        \raisebox{-\height}{\includegraphics[width=1\textwidth]{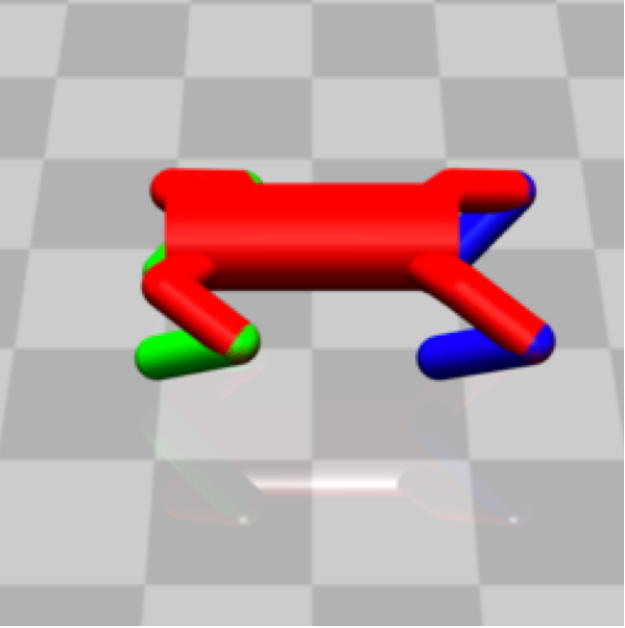}}
    \end{subfigure}
    
     \caption{
     \rebuttal{doggo robot: a quadrupedal robot with bilateral symmetry with 12-dimensional control space.}}
     \label{fig:doggo_robot}
\end{figure*}

\begin{figure*}
    \centering
    \begin{subfigure}[b]{0.30\textwidth}
        \raisebox{-\height}{\includegraphics[width=1\textwidth]{fig/rebuttal/doggo-goal-hazard1-0.15/_performance-crop.pdf}}
        \caption{Average episode reward}
    \end{subfigure}
    \hfill
    \begin{subfigure}[b]{0.30\textwidth}
        \raisebox{-\height}{\includegraphics[width=1\textwidth]{fig/rebuttal/doggo-goal-hazard1-0.15/_averageepcost-crop.pdf}}
        \caption{Average episode cost}
    \end{subfigure}
    \hfill
    \begin{subfigure}[b]{0.30\textwidth}
        \raisebox{-\height}{\includegraphics[width=1\textwidth]{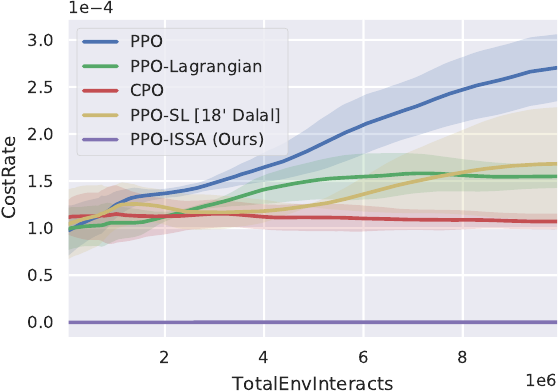}}
        \caption{Average episode costrate}
    \end{subfigure}
     \caption{
     \rebuttal{Average performance of PPO-ISSA and baseline methods on Goal-hazard1-0.15 environment of doggo robot over five seeds. The three columns represent average episode return, average episode cost and overallcost rate of constraints.}}
     
     \label{fig:doggo_goal-hazard1-0.15_complete}
\end{figure*}

\subsubsection{ISSA in higher dimensional constrol systems}
\label{sec: ISSA-scalability}
We test ISSA with a doggo robot as shown in \Cref{fig:doggo_robot} with 12 dimensional control space and 80 dimensional state space. We evaluate ISSA with the doggo robot in the Goal-Hazard1-0.15 suite. As shown in \Cref{fig:doggo_goal-hazard1-0.15_complete}, ISSA is able to guarantee zero-violation in higher dimensional control systems. We notice that in system with higher control dimensions, safe RL methods like PPO-Lagrangian and CPO perform poorly compared to PPO, which demonstrates the constrained RL algorithms struggle to learn good reward performance for complex locomotion behavior. Similar comparison results are also reported in Safety Gym benchmarks~\cite{ray2019benchmarking}. In contrast, PPO-ISSA is able to achieve the best reward performance while guaranteeing zero safety violation, showing the scalability of ISSA to achieve satisfying reward performance in systems with higher control dimensions.

To illustrate the computation cost of searching for a safe control using ISSA in systems with higher control dimensions, we apply ISSA to find safe controls for both Point robot and doggo robot in the Goal-Hazard1-0.15 suite with different number of unit gradient vectors generated by guassian distribution. Note that we desire to encourage ISSA phase 1 to find safe control, while the functionality of ISSA phase 2 is the fail-safe strategy for ISSA phase 1 to ensure feasibility of ISSA since 1) ISSA phase 2 is relatively more computational expensive than ISSA phase 1 due to random sampling, and 2) ISSA phase 2 can only return one safe control candidate. Thus, we are especially interested in analyzing the performance of ISSA phase 1. 

We report the average computation time, ISSA phase 1 success rate and number of safe control candidates found by ISSA phase 1 under different robot types and number of unit gradient vectorss in \Cref{tab:ablation_doggo}, where the success of ISSA phase 1 is defined as it returns at least 1 safe control which may lead to a large deviation from the original nominal control. As demonstrated in \Cref{tab:ablation_doggo}, we only need 20 unit gradient vectors for 12 dimensional control space doggo robot to achieve satisfying success rate in ISSA phase 1 and number of safe control candidates. On the other hand, we need 10 unit gradient vectors for 2 dimensional control space Point robot. Therefore, higher control dimensionality will not necessarily increase the computation cost exponentially for ISSA to find safe control.  We also notice that, even with the same number of vectors, the computaion time of doggo robot is 3 times longer than that of Point robot due to the doggo robot simulation takes longer than point robot simulaton per step in MuJoCo simulator. Here we also highlight a fact that the process of AdamBA outreach/decay for each unit gradient vector is independent from each other, thus we can always accelerate ISSA by parallel computation, which is then discussed in \Cref{sec:adamba_scalability}.


\begin{table*}[t]
\vskip 0.15in
\caption{\rebuttal{Average Computation time, succuss ratio of ISSA phase 1 and number of safe control candidates of 200 ISSA runs on Goal-Hazard1-0.15.}}
\begin{center}
\resizebox{\textwidth}{!}{%
\begin{tabular}{c|c|ccc}
\toprule
& Number of vectors & Overall (non-parallel) ISSA Time $T_{all}$ (s)  & ISSA phase 1 success rate & Number of safe control\\
\hline
& $n=3$ & 0.023 & 0.962 & 1.193\\
& $n=5$ & 0.038 & 1.000  & 2.258\\
Point robot & \textbf{$n=10$ (used)} & 0.076 & 1.000  & 4.576 \\
Control dimension = 2& $n=20$ & 0.160  & 1.000 & 9.838 \\
& $n=40$ & 0.302 & 1.000  & 18.055 \\
& $n=100$ & 0.737 & 1.000 & 25.740 \\
\hline
& $n=3$ &0.068 & 0.802 & 2.041\\
& $n=5$ & 0.116 & 0.903  & 3.051\\
Doggo robot & $n=10$ & 0.228 & 0.925  & 5.127 \\
Control dimension = 12& \textbf{$n=20$ (used)} & 0.461 & 0.981  & 10.673 \\
& $n=40$ & 0.910 & 0.990  & 19.373 \\
& $n=100$ & 2.190 & 1.000  & 33.910 \\

\bottomrule
\end{tabular}
}
\label{tab:ablation_doggo}
\end{center}
\end{table*}

\subsubsection{AdamBA in higher dimensional space}
\label{sec:adamba_scalability}
\begin{figure*}[htbp]
    \centering
    \begin{subfigure}[b]{0.40\textwidth}
        \raisebox{-\height}{\includegraphics[width=1\textwidth]{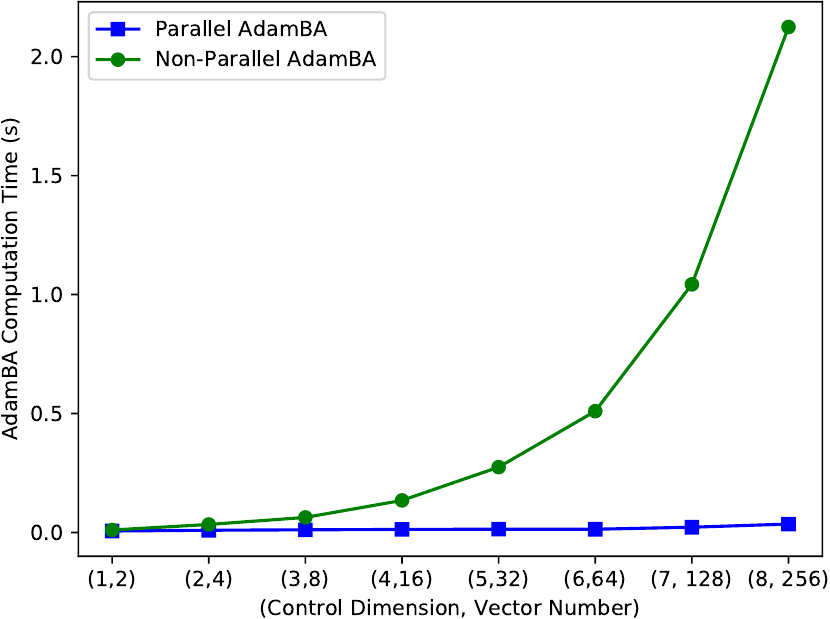}}
    \end{subfigure}
    \hfill
    \centering
    \begin{subfigure}[b]{0.40\textwidth}
        \raisebox{-\height}{\includegraphics[width=1\textwidth]{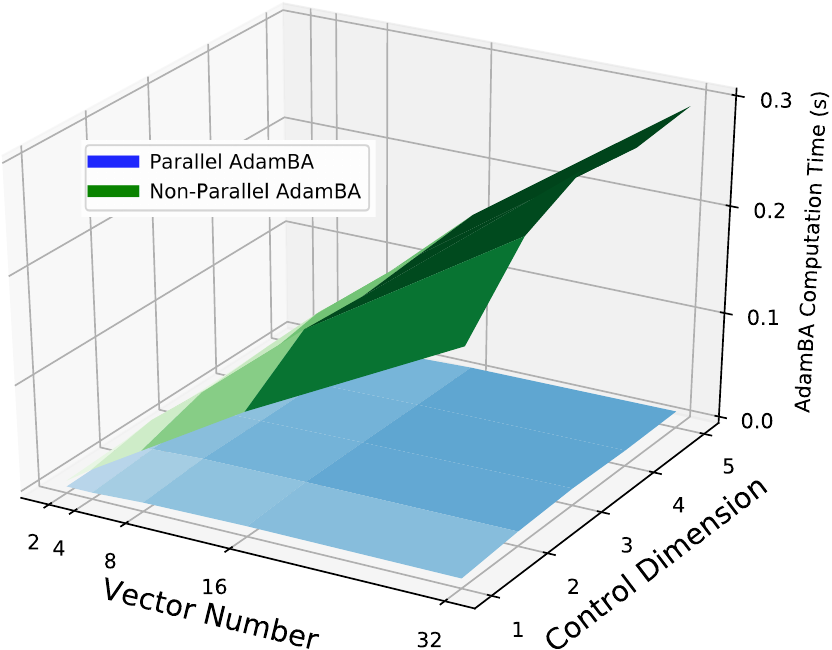}}
    \end{subfigure}
    
    \vspace{0.5cm}
    \caption{\rebuttal{The average computation time of non-parallel AdamBA and parallel AdamBA on toy problem with different control dimensions and number of vectors.}}
    \label{fig:parallel}
\end{figure*}

As reported in \Cref{sec: ISSA-scalability}, in MuJoCo environment, the sampling cost of AdamBA is not exponentially increasing as the control dimensions increase, where we only need 10 vectors for point robot (2-dimensional control space) and only need 20 vectors for doggo robot (12-dimensional control space). However, let's consider a worse case for AdamBA, where the relative size of safe control space in the entire control space is exponentially decreasing with dimension linearly increases, which could result in the computation cost of AdamBA to find the boundary exponentially increasing. We synthesis a simple control problem, where only half of the control space is safe for each dimension, which means the portion of safe control space to the entire control space is $\frac{1}{2^n}$ where $n$ is the number of control space dimensions. Note that for every unit gradient vector, the process of AdamBA outraech/decay is independent of the other unit gradient vectors, which means we could utilize parallel computation in practice when applying real-time robots. We report the average computation time of AdamBA on the prescribed toy control problem in \Cref{fig:parallel}, where we can see that the computation cost of parallel AdamBA remains nearly the same as the number of vectors exponentially increases, which shows the potential capability of AdamBA scaling to higher dimensional real-time control systems.

\section{Conclusion and Future Work}
\label{sec:conclusion}
Safety guarantee is critical for robotic applications in real world, such that robots can persistently satisfy safety constraints. This paper presents a model-free safe control strategy to synthesize safeguards for DRL agents, which will ensure zero safety violation during training. In particular, we present an implicit safe set algorithm as a safeguard, which synthesizes the safety index (also called the barrier certificate) and the subsequent safe control law only by querying a black-box dynamics function (e.g., a digital twin simulator).  The theoretical results indicate that the synthesized safety index guarantees nonempty set of safe control for all system states, and ISSA guarantees \textit{finite time convergence} and \textit{forward invariance} 
to the safe set for both continuous-time and discrete-time system. We further validate the proposed safeguard with DRL on state-of-the-art safety benchmark Safety Gym. Our proposed method achieves zero safety violation and \rebuttal{$95\% \pm 9\% $} reward performance compared to state-of-the-art safe DRL methods.

There are three major directions for future work. Firstly, we will further generalize the safety index synthesis rule to cover a wider range of applications other than collision avoidance in 2D. Secondly, we will further speed up the implicit model evaluation step by replacing the physical engine based simulator with a learned surrogate model while taking the learned dynamics error into account. Third, obstacle avoidance tasks in densely cluttered environments — where multiple safety-critical constraints may be active simultaneously — still require further theoretical advancements and algorithmic exploration.


\begin{acks}
This work is in part supported by Amazon Research Award and National Science Foundation through Grant \#2144489. We also thank Tairan He for assistance with experiments. 
\end{acks}

\printbibliography

\appendix

\section{Reproducibility Checklist for JAIR}

Select the answers that apply to your research -- one per item. 

\subsection*{All articles:}

\begin{enumerate}
    \item All claims investigated in this work are clearly stated. 
    [yes]
    \item Clear explanations are given how the work reported substantiates the claims. 
    [yes]
    \item Limitations or technical assumptions are stated clearly and explicitly. 
    [yes]
    \item Conceptual outlines and/or pseudo-code descriptions of the AI methods introduced in this work are provided, and important implementation details are discussed. 
    [yes]
    \item 
    Motivation is provided for all design choices, including algorithms, implementation choices, parameters, data sets and experimental protocols beyond metrics.
    [yes]
\end{enumerate}

\subsection*{Articles containing theoretical contributions:}
Does this paper make theoretical contributions? 
[yes] 

If yes, please complete the list below.

\begin{enumerate}
    \item All assumptions and restrictions are stated clearly and formally. 
    [yes]
    \item All novel claims are stated formally (e.g., in theorem statements). 
    [yes]
    \item Proofs of all non-trivial claims are provided in sufficient detail to permit verification by readers with a reasonable degree of expertise (e.g., that expected from a PhD candidate in the same area of AI). 
    [yes]
    \item
    Complex formalism, such as definitions or proofs, is motivated and explained clearly.
    [yes]
    \item 
    The use of mathematical notation and formalism serves the purpose of enhancing clarity and precision; gratuitous use of mathematical formalism (i.e., use that does not enhance clarity or precision) is avoided.
    [yes]
    \item 
    Appropriate citations are given for all non-trivial theoretical tools and techniques. 
    [yes]
\end{enumerate}

\subsection*{Articles reporting on computational experiments:}
Does this paper include computational experiments? [yes]

If yes, please complete the list below.
\begin{enumerate}
    \item 
    All source code required for conducting experiments is included in an online appendix 
    or will be made publicly available upon publication of the paper.
    The online appendix follows best practices for source code readability and documentation as well as for long-term accessibility.
    [yes]
    \item The source code comes with a license that
    allows free usage for reproducibility purposes.
    [yes]
    \item The source code comes with a license that
    allows free usage for research purposes in general.
    [yes]
    \item 
    Raw, unaggregated data from all experiments is included in an online appendix 
    or will be made publicly available upon publication of the paper.
    The online appendix follows best practices for long-term accessibility.
    [yes]
    \item The unaggregated data comes with a license that
    allows free usage for reproducibility purposes.
    [yes]
    \item The unaggregated data comes with a license that
    allows free usage for research purposes in general.
    [yes]
    \item If an algorithm depends on randomness, then the method used for generating random numbers and for setting seeds is described in a way sufficient to allow replication of results. 
    [yes]
    \item The execution environment for experiments, the computing infrastructure (hardware and software) used for running them, is described, including GPU/CPU makes and models; amount of memory (cache and RAM); make and version of operating system; names and versions of relevant software libraries and frameworks. 
    [yes]
    \item 
    The evaluation metrics used in experiments are clearly explained and their choice is explicitly motivated. 
    [yes]
    \item 
    The number of algorithm runs used to compute each result is reported. 
    [yes]
    \item 
    Reported results have not been ``cherry-picked'' by silently ignoring unsuccessful or unsatisfactory experiments. 
    [yes]
    \item 
    Analysis of results goes beyond single-dimensional summaries of performance (e.g., average, median) to include measures of variation, confidence, or other distributional information. 
    [yes]
    \item 
    All (hyper-) parameter settings for 
    the algorithms/methods used in experiments have been reported, along with the rationale or method for determining them. 
    [yes]
    \item 
    The number and range of (hyper-) parameter settings explored prior to conducting final experiments have been indicated, along with the effort spent on (hyper-) parameter optimisation. 
    [yes]
    \item 
    Appropriately chosen statistical hypothesis tests are used to establish statistical significance
    in the presence of noise effects.
    [yes]
\end{enumerate}

\subsection*{Articles using data sets:}
Does this work rely on one or more data sets (possibly obtained from a benchmark generator or similar software artifact)? 
[no]

If yes, please complete the list below.
\begin{enumerate}
    \item 
    All newly introduced data sets 
    are included in an online appendix 
    or will be made publicly available upon publication of the paper.
    The online appendix follows best practices for long-term accessibility with a license
    that allows free usage for research purposes.
    [yes/partially/no/NA]
    \item The newly introduced data set comes with a license that
    allows free usage for reproducibility purposes.
    [yes/partially/no]
    \item The newly introduced data set comes with a license that
    allows free usage for research purposes in general.
    [yes/partially/no]
    \item All data sets drawn from the literature or other public sources (potentially including authors' own previously published work) are accompanied by appropriate citations.
    [yes/no/NA]
    \item All data sets drawn from the existing literature (potentially including authors’ own previously published work) are publicly available. [yes/partially/no/NA]
    \item All new data sets and data sets that are not publicly available are described in detail, including relevant statistics, the data collection process and annotation process if relevant.
    [yes/partially/no/NA]
    \item 
    All methods used for preprocessing, augmenting, batching or splitting data sets (e.g., in the context of hold-out or cross-validation)
    are described in detail. [yes/partially/no/NA]
\end{enumerate}

\end{document}